\documentclass{article}

\usepackage{microtype}
\usepackage{graphicx}
\usepackage{subcaption}
\usepackage{caption}
\usepackage{multirow}
\usepackage{array}
\usepackage{threeparttable}
\usepackage{booktabs}
\usepackage{tocloft}

\usepackage{fancybox}
\usepackage{xcolor}

\usepackage{hyperref}

\usepackage[accepted]{icml2025}

\usepackage{amsmath}
\usepackage{amssymb}
\usepackage{mathtools}
\usepackage{amsthm}

\usepackage[capitalize,noabbrev]{cleveref}

\theoremstyle{plain}

\theoremstyle{definition}

\theoremstyle{remark}

\usepackage[textsize=tiny]{todonotes}

\icmltitlerunning{Mitigating Data Scarcity in Time Series Analysis: A Foundation Model with Series-Symbol Data Generation}

\begin{document}

\twocolumn[
\icmltitle{Mitigating Data Scarcity in Time Series Analysis: \\A Foundation Model with Series-Symbol Data Generation}

\begin{icmlauthorlist}
\icmlauthor{Wenxuan Wang}{xd_ai}
\icmlauthor{Kai Wu}{xd_ai}
\icmlauthor{Yujian Betterest Li}{xd_ai}
\icmlauthor{Dan Wang}{ISN}
\icmlauthor{Xiaoyu Zhang}{xd_sec}
\icmlauthor{Jing Liu}{xd_ai}
\end{icmlauthorlist}

\icmlaffiliation{xd_ai}{School of Artificial Intelligence, Xidian University, Xi'an, Shaanxi, China}
\icmlaffiliation{ISN}{The State Key Laboratory of Integrated Services Networks, Xi'an, Shaanxi, China}
\icmlaffiliation{xd_sec}{School of Network and Information Security, Xidian University, Xi'an, Shaanxi, China}

\icmlcorrespondingauthor{Kai Wu}{kwu@xidian.edu.cn}
\icmlcorrespondingauthor{Dan Wang}{danwang@xidian.edu.cn}

\icmlkeywords{Machine Learning, ICML}

\vskip 0.3in
]

\printAffiliationsAndNotice{}  

\begin{abstract}
Foundation models for time series analysis (TSA) have attracted significant attention. However, challenges such as data scarcity and data imbalance continue to hinder their development. To address this, we consider modeling complex systems through symbolic expressions that serve as semantic descriptors of time series. Building on this concept, we introduce a series-symbol (S2) dual-modulity data generation mechanism, enabling the unrestricted creation of high-quality time series data paired with corresponding symbolic representations. Leveraging the S2 dataset, we develop \texttt{SymTime}, a pre-trained foundation model for TSA. \texttt{SymTime} demonstrates competitive performance across five major TSA tasks when fine-tuned with downstream task, rivaling foundation models pre-trained on real-world datasets. This approach underscores the potential of dual-modality data generation and pretraining mechanisms in overcoming data scarcity and enhancing task performance.
\end{abstract}

\section{Introduction}
\label{sec:introduction}

researchers \cite{KDD-Survey, TSA-survey-PAMI}.
In recent years, with the rapid advancement of deep learning, foundation models for TSA have garnered widespread attention due to their superior generalization capabilities, scalability and advantages in few-shot learning \cite{GPT4TS}. 
Presently, pre-training methods such as mask time series modeling (MTM) \cite{HiMTM,TimeSiam}, contrastive learning \cite{COMET}, and generative modeling \cite{Timer} have given rise to a series of foundation models for time series, achieving significant results in TSA tasks.

\iffalse
\begin{figure}[!t]
\centerline{\includegraphics[width=0.99\linewidth]{images/SymTime_small.pdf}}
\caption{SymTime}
\label{figure:begin}
\end{figure}
\fi

Coupled with issues of data privacy \cite{DataPrivacy, COMET}, existing time series datasets are smaller compared to those in the fields of computer vision (CV) and natural language processing (NLP). Besides, current large-scale time series datasets face significant data imbalance issues, with certain types such as finance and healthcare still being relatively scarce (see Appendix \ref{sec:Analysis of Existing Dataset}). According to scaling laws \cite{Neural-Sclaing-Laws}, this can lead to performance bias in the time series foundation models, reducing their generalization capabilities on out-of-distribution data \cite{Time-Scaling-Laws}. 

To address the issue of data scarcity and data imbalance, this paper, starting from the nature and mechanisms of time series, posits that time series are representations of complex dynamical systems \cite{TimeSeries4ComplexSystems,li2023discover}. On one hand, the intricate patterns within natural systems can be captured through observed numerical data \cite{SNIP}; for instance, the time series of body temperature fluctuations throughout a day is derived from observations of the human system. On the other hand, complex systems can be expressed abstractly using mathematical symbols and formulas \cite{Symbolic}, with ordinary differential equations (ODE) and partial differential equations (PDE) being the most common methods for modeling complex systems \cite{DE4ModelingCS}. Symbols provide semantic information for modeling complex systems \cite{Exhaustive}. Therefore, as time series serve as the dynamic representation of observing complex systems, they can form a pairing relationship with the symbols used to model these systems.

To this end, we provide a series-symbol (S2) dual-modality data generation mechanism. This allows us to unrestrictedly produce high-quality time series and their paired symbolic data, constructing a large-scale S2 dataset. Then, we pretrain a time series foundation model \texttt{SymTime} with symbolic semantic information on this dataset. 
We train a Transformer-based time series encoder and a symbol encoder composed of a pre-trained large language model (LLM) \cite{GPT-2} through MTM and mask language modeling (MLM) \cite{BERT} to learn the basic representations of series and symbols respectively. Subsequently, using momentum distillation \cite{ALBEF, momentum-distillation}, we introduce contrastive losses between series and symbols through the added \texttt{[CLS]} token to further learn the cross-modal pairing representations and knowledge between series and symbols. To encapsulate our work, the contributions are as follows:

\begin{itemize}
    \item We posit that time series are representations of complex dynamical systems and symbolic expressions can be regarded as the semantic information of time series. Based on this, we provide a method for the infinite generation of high-quality dual-modality series-symbol data. 
    The size of the S2 dataset directly correlates with model performance on downstream tasks, and our S2 data can evenly cover the basic representations of all types of time series.
    \item Using synthetic S2 datasets, we construct the dual-modality time series pretrained foundation model, \texttt{SymTime}, using mask modeling and contrastive learning, thereby obtaining a pre-trained model with symbolic semantic information and cross-modal representations. Compared to foundation models pre-trained on real datasets, \texttt{SymTime} achieves competitive results in the five major TSA tasks.
    \item Additionally, we observe that: 1) \texttt{SymTime} achieves better performance with a smaller model parameter count and memory capacity than existing foundation models in forecasting; 2) \texttt{SymTime} successfully learns fundamental time series representations, enabling zero-shot imputation.
\end{itemize}

\section{Related Work}

In CV and NLP \cite{CLIP}, pre-trained foundation models (PTFMs) have been demonstrated to adapt to a variety of downstream tasks after fine-tuning on specific datasets, exhibiting excellent generalization and scalability \cite{TimeMIL}. Inspired by this, recent years have seen significant progress in PTFMs for TSA \cite{Transformer-in-TSA, KDD-Survey}, with the emergence of various pre-training methods. MOIRAI, through MTM and reconstruction, has been pre-trained on large datasets (27B), yielding a universal forecasting model with significant zero-shot advantages \cite{MOIRAI}. Timer, after generative pre-training on large datasets (1B), has performed well in forecasting \cite{Timer}. TimeGPT trained a encoder-decoder Transformer with 100B data \cite{TimeGPT}. COMET, using multi-level contrastive learning on a large ECG dataset, has obtained a medical time series PTFMs with few-shot advantages \cite{COMET}. 

As discussed in Appendix \ref{sec:Analysis of Existing Dataset}, these baseline models still face challenges related to data scarcity and data imbalance. In the next section, we introduce the proposed data generation mechanism and the corresponding dual-modality foundation model designed to address these issues. The review of other topics can be found in Appendix \ref{sec:related work}.

\begin{figure*}[!t]
\centerline{\includegraphics[width=0.95\linewidth]{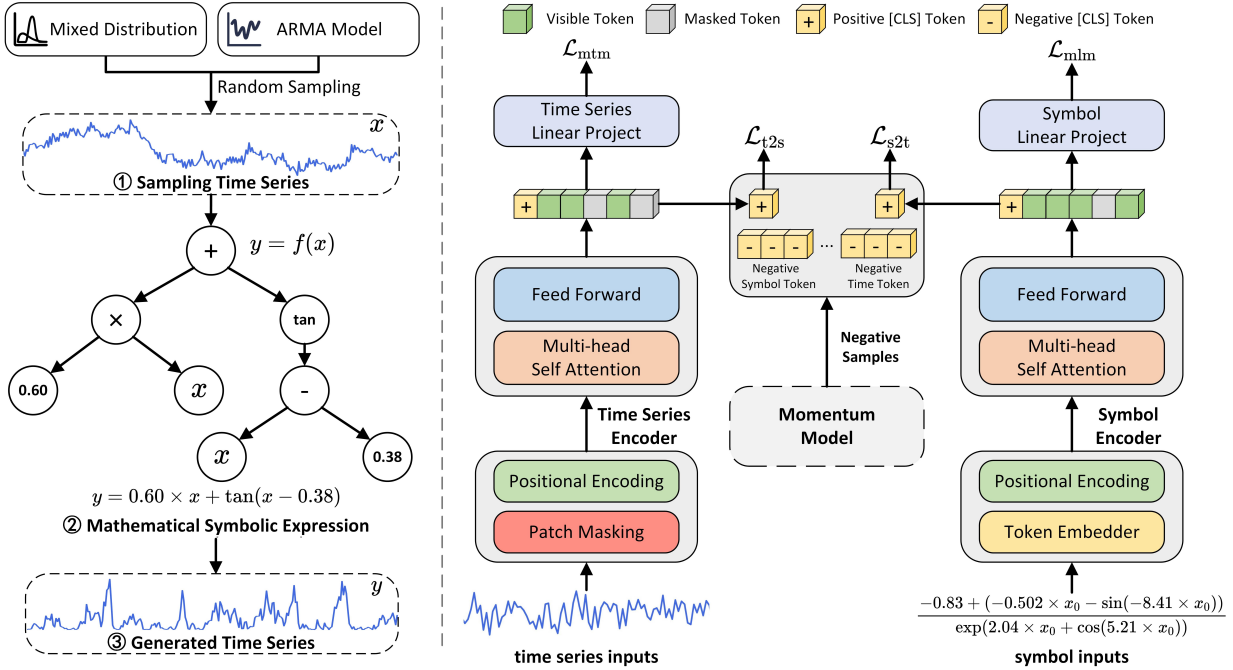}}
\vskip -0.05in
\caption{S2 dataset generation mechanism (\textbf{left}) and \texttt{SymTime} network architecture (\textbf{right}).}
\label{figure:symtime}
\end{figure*}

\section{Main Methods}
\label{sec:main methods}

\subsection{Series-Symbol (S2) Dataset Generation}
\label{sec:data generation}

The pre-training of \texttt{SymTime} relies on a large synthetic series-symbol (S2) dataset. The specific generation process is shown in Figure \ref{figure:symtime} (\textbf{left}). Firstly, we construct a multivariate input-output symbolic expression $f(\cdot)$ through random sampling \cite{DL4Symbolic}. Then, we use the randomly generated sampling series $X \in \mathbb{R} ^ {M \times L}$ to forward propagate through the symbolic expression to obtain the generated series $Y=f(X) \in \mathbb{R} ^ {N \times L}$, where $N$ and $M$ represent the dimensions of the input and output series respectively, and $L$ is the length of the series.

\textbf{Sampling of Functions.} We first determine the dimensions of the input and output series and randomly select the number of binary operators from a uniform distribution. Then, we construct a binary tree with symbolic variables and constants as leaf nodes, and binary operators as two-child nodes to form the basic framework of the symbolic expression \cite{SNIP}. Finally, we enhance the diversity of the expressions by inserting unary operators as one-child nodes and applying affine transformations randomly \cite{Symbolic}. After determining the framework of the tree, we select specific operators for all nodes from the uniform distribution $\mathcal{U}(+, -, \times)$ and $\mathcal{U}$(inv, abs, pow2, pow3, sqrt, sin, cos, tan, arctan, log, exp), and randomly initialize leaf nodes with random constants and variables. The symbolic expression can be read out via in-order traversal.

\textbf{Generating Inputs and Outputs.} To better align the generated series with time series characteristics, we sample multi-channel input series $X$ from both mixed distributions (MD) \cite{Symbolic} and randomly parameterized ARMA($p$, $q$) models \cite{ARMA}. We first select the distribution number in MD or the order $(p, q)$ in ARMA model from the uniform distribution \cite{ARIMA, DL4Symbolic}. Then, we randomly initialize the parameters of the MD (including the mean and variance of a normal distribution, as well as the range of a uniform distribution) or the ARMA model (including autoregressive and moving average components). We can obtain the sampling series $X$ through forward propagation. Finally, we standardize each channel and obtain the generated series $Y=f(X)$ through the symbolic expressions. To ensure data quality, we discard samples outside the domain of $f(\cdot)$ and excessively large generated values \cite{SNIP}.

Details and specific processes for data generation are provided in Appendix \ref{sec:appendix A}. Through the two processes mentioned, we can obtain a large number of multi-channels series along with their corresponding symbolic expressions. We concatenate the time series and divide it into patches for the time series encoder \cite{PatchTST}, and tokenize the symbolic expressions for the symbolic encoder \cite{DistilBERT}. We generated a total of 25M series-symbol pairs in S2 dataset, with the total series length of 50B.

\subsection{Model Architecture and Pre-training Objectives}
\label{sec:model architecture}

As shown in Figure \ref{figure:symtime} (\textbf{right}), \texttt{SymTime} mainly consists of three components: a time series encoder, a symbolic encoder, and momentum models \cite{MoCo_v1}, each with its own distinct pre-training objectives.

\textbf{Time Series Encoder and Mask Time Series Modeling.} We employ a 6-layer Transformer as the time series encoder. An input time series is first divided into non-overlapping patches $P$=$\{p_1, p_2, \cdots, p_n\}$ using a sliding window approach \cite{PatchTST, Time-LLM}. Then, we add random masks to these patches and embed them into the time series encoder to learn the basic representation of the series \cite{MAE}, obtaining the corresponding embedded sequence $T$=$\{t_{\text{cls}}, t_1, t_2, \cdots, t_n\} $, where $t_{\text{cls}}$ is the \texttt{[CLS]} token added by the model \cite{BERT}. Our training objective is to restore the masked patch $p_j^M$ through the final linear mapping layer. The MTM loss is as follows:
\begin{equation}
    {\mathcal{L}_{\mathrm{mtm}}}=\frac{1}{N}{{\sum_{j\in {\mathbf{M}_{\mathrm T}}}{\left \|{{p}_{j}}-\hat {p}_j\right \|}}^{2}},
\label{eq:mtm}
\end{equation}
where $\mathbf M_{\mathrm T}$ is the set of masked patches, and $\hat{p}_j$ represents the patch reconstructed by the time series encoder and linear projection \cite{SparseTSF}. This approach allows the time series encoder to learn the representation of the time series.

\textbf{Symbol Encoder and Mask Language Modeling.} We treat symbolic expressions as natural language and use a 6-layer DistilBert as the symbolic encoder to learn their representations \cite{DistilBERT}. We first randomly replace the words in the symbolic expressions with \texttt{[Mask]} \cite{BERT} and input them into the symbolic encoder for representation learning, obtaining the embedded sequence: $S = \{s_{\text{cls}}, s_1, \ldots, s_m\}$, where $s_{\text{cls}}$ is the \texttt{[CLS]} token added by the model. We have DistilBert predict the masked tokens using bidirectional information. Let $\hat{s}$ denote a masked token, and $p^{\text{mask}}(\hat{s})$ denote the model's predicted probability for the masked token. The MLM minimizes a cross-entropy $\mathbf H$ loss:
\begin{equation}
    {{\mathcal L}_{\mathrm{mlm}}}=\frac{1}{N}\sum_{j\in {{\mathbf{M}}_{\mathrm S}}}\mathbf{H}\left ({{{y}_{j}},{{p}_{j}^{mask}}(\hat{s})}\right ) ,
\end{equation}
where $\mathbf M_{\mathrm S}$ is the set of masked words in the symbolic expression, and $y_j$ is a one-hot vocabulary distribution with a probability of 1 for the ground-truth token.

\textbf{Series-Symbol Contrastive Learning.} In order to enable the time series encoder to learn the semantic information of symbolic expressions, we employ contrastive learning to associate the dual encoders in \texttt{SymTime}, allowing them to learn better unimodal representations. Specifically, \texttt{SymTime} will learn an inner product similarity function \cite{CLIP}: $\text{sim} = g_t (t_{\text{cls}}) ^ {\mathrm{T}} g_s (s_{\text{cls}})$, where $ t_{\text{cls}} $ and $ s_{\text{cls}} $ are the \texttt{[CLS]} tokens added during the embedding of time series and symbolic expressions, respectively. They represent the features of the entire input series and input symbols. $ g_t $ and $ g_s $ are linear projections that map the \texttt{[CLS]} embeddings of to a normalized low-dimensional space. Paired series-symbol representations should have higher similarity \cite{ALBEF}. Inspired by MoCo \cite{MoCo_v1}, we use a momentum model to obtain a large and consistent dictionary for contrastive learning. The normalized series and symbol features generated by the momentum model are denoted as $ g'_t(t'_{\text{cls}}) $ and $ g'_s(s'_{\text{cls}}) $, respectively. We denote $\mathrm{sim}(t, s) = g_t(t_{\text{cls}})^{\mathrm{T}} g_s'(s_{\text{cls}}')$ and $\mathrm{sim}(s, t) = g_s(s_{\text{cls}})^{\mathrm{T}} g_t'(t_{\text{cls}}')$. Thern, for each pair of series and symbol, the softmax similarity \cite{ALBEF, BEiT, MF-CLR} from series to symbol and from symbol to series can be defined as follow:
\begin{equation}
    p^{t2s}(t) = \frac{\mathrm{exp} \left ( \mathrm{sim}(t, s_m) / \tau \right )}{\sum_{m=1}^M\mathrm{exp} \left ( \mathrm{sim}(t, s_m) / \tau \right )},
\label{equation: p^{t2s}}
\end{equation}
\begin{equation}
    p^{s2t}(s) = \frac{\mathrm{exp} \left ( \mathrm{sim}(s, t_m) / \tau \right )}{\sum_{m=1}^M\mathrm{exp} \left ( \mathrm{sim}(s, t_m) / \tau \right )},
\label{equation: p^{s2t}}
\end{equation}
where $\tau$ is a learnable temperature parameter \cite{COMET, MoCo_v1}. Let $y^{t2s}(t)$ and $y^{s2t}(s)$ represent the one-hot similarity, with positive pairs having a probability of 1 and negative pairs having 0 \cite{MoCo_v1}. The contrastive loss for time series-symbol is formulated as the cross-entropy $\mathbf H$ between similarity $p$ and the true labels $y$:
\begin{equation}
    \mathcal{L}_{\mathrm{tsc}} = \frac{1}{2}\mathbb E \left [ \mathbf{H}\left ( y^{t2s}(t), p^{t2s}(t) \right ) + \mathbf{H}\left ( y^{s2t}(s), p^{s2t}(s) \right ) \right ].
\label{equation:L_tsc}
\end{equation}

By employing this method, the dual-encoder \texttt{SymTime} can align the representations of the two encoders by bringing positively correlated series-symbol pairs closer together in the representation space \cite{BEiT}.

\subsection{Momentum Distillation for Masked Data Learning}
\label{sec:momentum distillation}

Since we obtain the \texttt{[CLS]} token from masked data for contrastive learning, even with a low masking ratio, it still has an impact on the representation. Inspired by ALBEF \cite{ALBEF}, we treat the mask as noise added to the series and symbol. Therefore, we use momentum distillation \cite{momentum-distillation} to overcome the impact of masking on data representation. We train \texttt{SymTime} not only with true labels of series-symbol pairs (Equation \ref{equation:L_tsc}) \cite{CLIP} but also from the pseudo-targets generated by the momentum model \cite{MoCo_v1, KD-survey}. This allows our encoder's predictions to match the predictions of the momentum models. Let the similarity functions generated by the momentum encoders be $\text{sim}'(t, s) = g_t(t'_{\text{cls}})^{\mathrm{T}} g_s(s'_{\text{cls}}) $ and $ \text{sim}'(s, t) = g_s(s'_{\text{cls}})^{\mathrm{T}} g_t(t'_{\text{cls}}) $. We compute soft pseudo targets $ q^{t2s} (t) $ and $ q^{s2t} (s) $ by replacing $ \text{sim} $ with $ \text{sim}' $ in Equations \ref{equation: p^{t2s}} and \ref{equation: p^{s2t}}. Then, the momentum distillation loss can be defined as:
\begin{equation}
    \mathcal{L}_{\mathrm{tsc}}^{\mathrm{mod}} = \frac{1}{2} \mathbb{E} \left [ \mathbf{KL} \left ( q^{t2s}(t) \| p^{t2s}(t) \right ) + \mathbf{KL} \left ( q^{s2t}(s) \| p^{s2t}(s) \right ) \right ],
\end{equation}
where $\mathbf{KL}$ denotes the Kullback-Leibler divergence. We introduce a scaling coefficient $\alpha$ to balance $\mathcal L_{\mathrm{tsc}}$ and $\mathcal L_{\mathrm{tsc}}^{\mathrm{mod}}$. Ultimately, all the pre-training objectives of \texttt{SymTime} can be expressed as:
\begin{equation}
    \mathcal L = \mathcal L_{\mathrm{mtm}} + \mathcal L _{\mathrm{mlm}} + \alpha \mathcal L _{\mathrm{tsc}} + ( 1 - \alpha ) \mathcal L _{\mathrm{tsc}} ^ {\mathrm{mod} }.
\end{equation}

\subsection{Model Fine-tuning for Downstream Tasks}

Through MTM and series-symbol contrastive learning, we pre-train the time series encoder to imbue it with semantic information from symbols \cite{MaskTime,SNIP}. When fine-tuning on downstream tasks, we use this encoder as the backbone to extract universal representations of time series. For the input multivariate time series, we first perform instance normalization to address data offset \cite{ReVIN}. Subsequently, we will adopt different strategies according to different tasks. For classification tasks, we divide the time series into patches through sliding window \cite{PatchTST,TimeXer}. Then, we input them into the encoder to extract features and project the outputs into the classification space through linear mapping. For other reconstruction tasks (forecasting, imputation and anomaly detection) \cite{Peri-midFormer}, we first decompose the input series into two parts: trend and period. Then, the trend is output through linear mapping. We divide the period part into different patches and input them into the encoder in a channel-independent manner. After linear mapping, we reintegrate the trend and period to get the final output.

\section{Experiments}
\label{sec:experiments}

In our experiments, we first validate the effectiveness of the time series encoder and the series-symbol pre-training method on 5 mainstream TSA tasks. Details about the pre-training can be found in Appendix \ref{sec:Appendix Pre-training}. These specific downstream tasks include long-term forecasting (Section \ref{sec:long-term forecasting}), short-term forecasting (Section \ref{sec:short-term forecasting}), classification (Section \ref{sec:classification}), imputation (Section \ref{sec:classification}), and anomaly detection (Section \ref{sec:anomaly detection}). We use the same benchmarks as TimesNet \cite{TimesNet} for our experiments, with specific details in Appendix \ref{sec:Appendix downstream tasks}. Then, in Section \ref{sec:ablation experiments}, we conduct ablation studies on the pre-training objectives of \texttt{SymTime} on long and short term forecasting. Subsequently, in Section \ref{sec:discussion}, we analyze the complexity of the model and demonstrate that the time series encoder in \texttt{SymTime} learns the semantic information of the symbols. Finally, We demonstrate that through MTM and contrastive learning, SymTime has a certain ability of zero-shot imputation and can distinguish the symbolic semantics of simple series. The model architecture of the time series and symbolic encoders are shown in Table \ref{table:model architecture}.

\textbf{Baselines.} The baselines including \textbf{Transformer-based models}: PatchTST \cite{PatchTST}, iTransformer \cite{iTransformer}, Autoformer \cite{Autoformer}, ETSformer \cite{ETSformer}, FEDformer \cite{FEDformer}, Non-stationary Transformer \cite{Non-stationary-transformers}, Crossformer \cite{Crossformer}, Informer \cite{Informer}, Anomaly Transformer \cite{Anomaly-Transformer}, Peri-midFormer \cite{Peri-midFormer}; \textbf{LLM-based models}: GPT4TS \cite{GPT4TS}, Time-LLM \cite{Time-LLM}, $S^2$IP-LLM \cite{S2IP-LLM}; \textbf{CNN-based models}: TimesNet \cite{TimesNet}, TSLANet \cite{TSLANet}, Rocket \cite{Rocket}, InceptionTime (InTime) \cite{InceptionTime} and MICN \cite{MICN}; \textbf{MLP-based models}: DLinear \cite{DLinear}, LightTS \cite{LightTS}, TimeMixer \cite{TimeMixer} and FilterNet \cite{FilterNet}. We alse compare with the \textbf{pre-trained foundation models}: Moirai \cite{MOIRAI}, Timer \cite{Timer}, UniTS \cite{UniTS} and Moment \cite{Moment}. Some models can be applied to all 5 TSA tasks, while others are suitable for only one or some specific tasks.

\begin{table}[!t]
\caption{The model architecture of the time series and symbolic encoders in \texttt{SymTime}.}
\label{table:model architecture}
% \vskip 0.05in
\begin{center}
\begin{small}
\begin{tabular}{ccccccc}
\toprule
Encoder & Layers & $d_{\mathrm{model}}$ & $d_{\mathrm{ff}}$ & Heads & Params \\
\midrule
Time & 6 & 512 & 2048 & 8 & 19M \\
Symbol & 6 & 786 & 3072 & 12 & 67M \\
\bottomrule
\end{tabular}
\end{small}
\end{center}
% \vskip -0.05in
\end{table}

\begin{figure}[!t]
    \centering
    \includegraphics[width=0.95\linewidth]{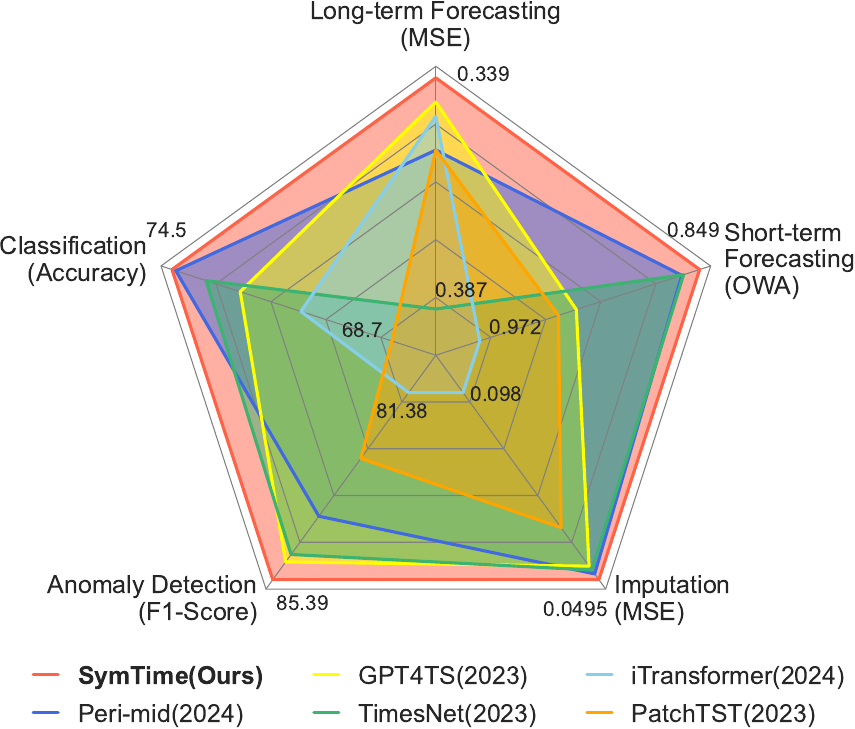}
    % \vskip -0.05in
    \caption{Model performance comparison with the state-of-the-art models in terms of five tasks.}
    \label{fig:main results of downstream tasks}
\end{figure}

\textbf{Main Results.} Figure \ref{fig:main results of downstream tasks} displays the comprehensive comparison results between \texttt{SymTime} and other foundation models, it excels across all 5 downstream tasks.

\subsection{Long-term Forecasting}
\label{sec:long-term forecasting}

\begin{table*}[!t]
\caption{Long-term forecasting task. The results are averaged from four different series length \{$96, 192, 336, 720$\}. (* means former.) See Appendix \ref{sec: long-term forecasting full results}. \textbf{\textcolor{red}{Red}}: best, \textcolor{blue}{Blue}: second best.}
\centering
%\vskip 0.05in
\begin{threeparttable}
\begin{footnotesize  }
\renewcommand{\multirowsetup}{\centering}
\setlength{\tabcolsep}{3pt}
\begin{tabular}{c|cc|cc|cc|cc|cc|cc|cc|cc|cc}
\toprule
 & \multicolumn{2}{c}{SymTime} & \multicolumn{2}{c}{Peri-mid*} & \multicolumn{2}{c}{Moriai} & \multicolumn{2}{c}{Timer} & \multicolumn{2}{c}{Time-LLM} & \multicolumn{2}{c}{TSLANet} & \multicolumn{2}{c}{S2IP-LLM} & \multicolumn{2}{c}{GPT4TS} & \multicolumn{2}{c}{TimeMixer} \\
\multirow{-2}{*}{Model} & \multicolumn{2}{c}{\textbf{(Ours)}}    & \multicolumn{2}{c}{\citeyearpar{Peri-midFormer}} & \multicolumn{2}{c}{\citeyearpar{MOIRAI}} & \multicolumn{2}{c}{\citeyearpar{Timer}}  & \multicolumn{2}{c}{\citeyearpar{Time-LLM}} & \multicolumn{2}{c}{\citeyearpar{TSLANet}} & \multicolumn{2}{c}{\citeyearpar{S2IP-LLM}}    & \multicolumn{2}{c}{\citeyearpar{GPT4TS}} & \multicolumn{2}{c}{\citeyearpar{TimeMixer}} \\
\cmidrule(lr){1-1}
\cmidrule(lr){2-3} \cmidrule(lr){4-5} \cmidrule(lr){6-7} \cmidrule(lr){8-9} \cmidrule(lr){10-11} \cmidrule(lr){12-13} \cmidrule(lr){14-15} \cmidrule(lr){16-17} \cmidrule(lr){18-19}
Metrics & MSE & MAE & MSE & MAE & MSE & MAE & MSE & MAE & MSE & MAE & MSE & MAE & MSE & MAE & MSE & MAE & MSE & MAE \\
\midrule
ETTm1 & 0.372 & {\color[HTML]{FF0000} \textbf{0.392}} & 0.409 & 0.410 & 0.398 & 0.417 & 0.388 & 0.402 & {\color[HTML]{FF0000} \textbf{0.369}} & {\color[HTML]{0000FF} 0.394} & 0.377 & 0.397 & 0.374 & 0.404 & {\color[HTML]{0000FF} 0.369} & 0.395 & 0.382 & 0.397 \\
ETTm2 & 0.283 & 0.328 & 0.290 & 0.328 & 0.296 & 0.348 & 0.405 & 0.408 & 0.275 & {\color[HTML]{FF0000} \textbf{0.324}} & 0.283 & 0.327 & {\color[HTML]{0000FF} 0.266} & {\color[HTML]{0000FF} 0.325} & {\color[HTML]{FF0000} \textbf{0.264}} & 0.328 & 0.279 & 0.325 \\
ETTh1 & {\color[HTML]{FF0000} \textbf{0.430}} & {\color[HTML]{FF0000} \textbf{0.436}} & 0.455 & 0.446 & 0.441 & 0.454 & 0.434 & 0.444 & 0.438 & 0.445 & 0.448 & 0.441 & 0.456 & 0.454 & {\color[HTML]{0000FF} 0.434} & {\color[HTML]{0000FF} 0.440} & 0.453 & 0.441 \\
ETTh2 & 0.375 & 0.405 & 0.400 & 0.416 & 0.402 & 0.411 & 0.428 & 0.441 & 0.369 & 0.407 & {\color[HTML]{FF0000} \textbf{0.355}} & {\color[HTML]{FF0000} \textbf{0.391}} & 0.362 & 0.405 & {\color[HTML]{0000FF} 0.359} & {\color[HTML]{0000FF} 0.403} & 0.388 & 0.408 \\
Weather & {\color[HTML]{0000FF} 0.247} & 0.276 & 0.262 & 0.283 & 0.265 & 0.299 & 0.329 & 0.358 & 0.247 & {\color[HTML]{FF0000} \textbf{0.269}} & 0.259 & 0.352 & {\color[HTML]{FF0000} \textbf{0.243}} & {\color[HTML]{0000FF} 0.274} & 0.265 & 0.285 & 0.253 & 0.280 \\
ECL & 0.187 & 0.276 & 0.178 & 0.267 & {\color[HTML]{FF0000} \textbf{0.167}} & {\color[HTML]{FF0000} \textbf{0.252}} & {\color[HTML]{0000FF} 0.177} & {\color[HTML]{0000FF} 0.267} & 0.180 & 0.269 & 0.199 & 0.283 & 0.191 & 0.283 & 0.206 & 0.291 & 0.185 & 0.274 \\
Traffic & 0.457 & 0.291 & 0.458 & 0.295 & 0.424 & {\color[HTML]{0000FF} 0.289} & 0.436 & {\color[HTML]{FF0000} \textbf{0.284}} & {\color[HTML]{0000FF} 0.418} & 0.306 & 0.463 & 0.310 & {\color[HTML]{FF0000} \textbf{0.417}} & 0.306 & 0.491 & 0.320 & 0.499 & 0.306 \\
Exchange & {\color[HTML]{FF0000} \textbf{0.359}} & {\color[HTML]{FF0000} \textbf{0.401}} & 0.388 & 0.417 & 0.373 & 0.417 & 0.382 & 0.425 & 0.376 & 0.414 & {\color[HTML]{0000FF} 0.368} & 0.414 & 0.472 & 0.478 & 0.370 & {\color[HTML]{0000FF} 0.411} & 0.403 & 0.423 \\
\midrule
Average & {\color[HTML]{0000FF} 0.339} & {\color[HTML]{FF0000} \textbf{0.351}} & 0.355 & 0.358 & 0.346 & 0.361 & 0.372 & 0.378 & {\color[HTML]{FF0000} \textbf{0.334}} & {\color[HTML]{0000FF} 0.353} & 0.344 & 0.364 & 0.348 & 0.366 & 0.345 & 0.359 & 0.355 & 0.357 \\
\bottomrule
\end{tabular}
\end{footnotesize  }
\end{threeparttable}
\label{table:long-term results}
% \vskip -0.15in  % 后续这里记得注释掉
\end{table*}

\begin{table*}[!t]
\caption{Short-term forecasting task on M4. The prediction lenghs are $\{6, 48 \}$ and results are weighted averaged from several datasets under different sample intervals. (* means former, TMixer is TimeMixer). See Appendix \ref{sec: short-term forecasting full results} for full results. \textbf{\textcolor{red}{Red}}: best, \textcolor{blue}{Blue}: second best.}
\centering
\vskip 0.05in
\renewcommand{\multirowsetup}{\centering}
\setlength{\tabcolsep}{2.8pt}
\begin{threeparttable}
\begin{footnotesize  }
\begin{tabular}{c|ccccccccccccc}
\toprule
 & SymTime & Peri-mid* & S2IP-LLM & Time-LLM & GPT4TS & TMixer & PatchTST & iTrans* & TimesNet & DLinear & LightTS & FED* & In* \\
\multirow{-2}{*}{Models} & \textbf{(Ours)} & \citeyearpar{Peri-midFormer} & \citeyearpar{S2IP-LLM} & \citeyearpar{Time-LLM} & \citeyearpar{GPT4TS} & \citeyearpar{TimeMixer} & \citeyearpar{PatchTST} & \citeyearpar{iTransformer} & \citeyearpar{TimesNet} & \citeyearpar{DLinear} & \citeyearpar{LightTS} & \citeyearpar{FEDformer} & \citeyearpar{Informer} \\
\midrule
SMAPE & {\color[HTML]{FF0000} \textbf{11.785}} & 11.897 & 12.514 & 12.584 & 12.367 & {\color[HTML]{0000FF} 11.885} & 12.866 & 13.233 & 11.888 & 12.500 & 11.962 & 12.605 & 15.018 \\
MASE & {\color[HTML]{FF0000} \textbf{1.584}} & 1.607 & 1.726 & 1.763 & 1.767 & {\color[HTML]{0000FF} 1.598} & 1.734 & 1.850 & 1.607 & 1.678 & 1.609 & 1.677 & 2.096 \\
OWA & {\color[HTML]{FF0000} \textbf{0.849}} & 0.859 & 0.913 & 0.915 & 0.918 & {\color[HTML]{0000FF} 0.856} & 0.928 & 0.972 & 0.858 & 0.899 & 0.862 & 0.903 & 1.102 \\
\bottomrule
\end{tabular}
\end{footnotesize  }
\end{threeparttable}
\label{table:short-term forecasting results}
% \vskip -0.10in  % 后续这里记得注释掉
\end{table*}

\begin{figure*}[!t]
\centerline{\includegraphics[width=0.95\linewidth]{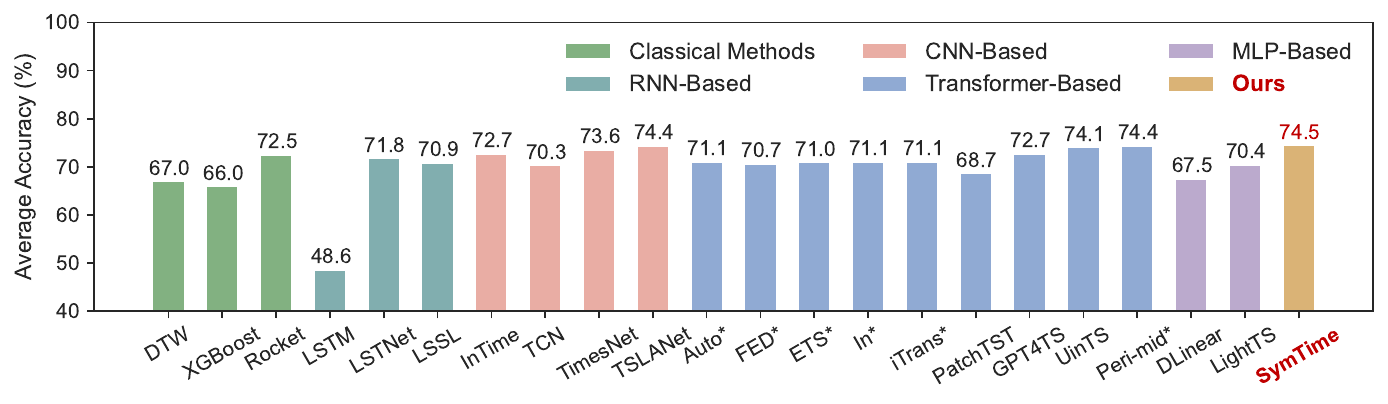}}
% \vskip -0.10in
\caption{Comparison of the average accuracy of \texttt{SymTime} and other baselines on 10 UEA datasets. See Appendix \ref{sec: classification full results} for full results.}
\label{figure:time series classification results}
\end{figure*}

\textbf{Setup.} Time series forecasting, which analyzes historical data patterns to predict future trends, is crucial for financial market analysis, inventory management, energy demand and other fields \cite{TimeMixer++, Timer-XL}. We adopt 8 real-world benchmark datasets for long-term forecasting, including ETTm1, ETTm2, ETTh1, ETTh2 \cite{Informer}, Weather \cite{weather}, ECL \cite{ECL}, Traffic \cite{traffic} and Exchange \cite{LSTNet}. The forecasting lengths are set to $\{96, 192, 336, 720 \}$. To ensure fairness in the comparison, we set the look-back window length of \texttt{SymTime} and all other models to 96, except Moirai and Timer are 672 and $S^2$IP-LLM is 512.

\textbf{Results.} Table \ref{table:long-term results} clearly demonstrates that \texttt{SymTime} achieves excellent performance in long-term forecasting tasks. Our model surpasses Peri-midFormer, GPT4TS and TimesNet, which are foundation models for the 5 major tasks, as well as Moirai and Timer, two general forecasting models. \texttt{SymTime} approaches and surpasses the customized time series forecasting model Time-LLM in terms of MSE and MAE metrics. However, Time-LLM relies on a large-scale LLM as its backbone, whereas \texttt{SymTime} can achieve comparable results through pre-training on synthetic datasets and fine-tuning with a more lightweight model.

\subsection{Short-term Forecasting}
\label{sec:short-term forecasting}

\textbf{Setup.} We adopt M4 benchmark \cite{M4team2018dataset} for short-term forecasting, which contains the yearly, quarterly and monthly collected univariate marketing data. Then, we use symmetric mean absolute error (SMAPE), mean absolute scaled error (MASE) and overall weighted average (OWA) to measure the forecasting performance, which are calculated as detailed in Appendix \ref{sec:Appendix Metrics}.

\textbf{Results.} Table \ref{table:short-term forecasting results} indicates that \texttt{SymTime} after pre-training, surpasses TimeMixer, Peri-midFormer and TimesNet on the short-term forecasting tasks in terms of SMAPE, MASE and OWA metrics, achieving state-of-the-art performance. Specifically, \texttt{SymTime} performs well on Yearly, Quarterly and Monthly datasets, demonstrating its capability to capture not only the trends of annual variations but also the cyclic characteristics of seasonal and monthly encoding.

\subsection{Classification}
\label{sec:classification}

\textbf{Setup.} Time series classification is crucial for the identification and diagnosis of patterns in complex systems and plays a significant role in various fields such as financial analysis, medical diagnosis and industrial monitoring \cite{TSC-survey}. Using the experimental setup from TimesNet \cite{TimesNet}, we test \texttt{SymTime}'s discriminative ability on 10 UEA multivariate time series classification datasets \cite{UEA}, including categories such as Industry, Face Detection, ECG, Voice and Transportation.

\textbf{Results.} As shown in Figure \ref{figure:time series classification results}, \texttt{SymTime} achieves an average accuracy of 74.5\%, surpassing all baselines, indicating that \texttt{SymTime} is competitive in classification tasks.

\subsection{Imputation}
\label{sec:imputation}

\textbf{Setup.} Sensors monitoring complex systems in the real world may experience distortions or malfunctions, leading to partial missing data in the collected time series. Therefore, time series imputation is crucial for the recovery of complete datasets. We verify \texttt{SymTime}'s imputation capabilities on 6 datasets: ETTm1, ETTm2, ETTh1, ETTh2 \cite{Informer}, Weather \cite{weather} and ECL \cite{ECL}. To test the model's imputation ability under varying degrees of missing data, we add random masks at proportions of \{$12.5\%, 25\%, 37.5\%, 50\%$\} in point level on time series of length 96. Since \texttt{SymTime} was pre-trained by randomly masking patches level for series reconstruction and masks are added randomly in point level in the imputation task. Considering the differences between these masking approaches and the potential disruption of the series's original trends and periodic features at higher mask rates, we adopt per-interpolation for the masked series from \cite{Peri-midFormer}. Analysis and ablation experiments regarding this method are presented in Appendix \ref{sec:pre-interpolation}. 
% The results demonstrate that per-interpolation can be used as a model-independent feature engineering to improve the performance of downstream tasks.

\textbf{Results.} Table \ref{table:imputation results} shows that \texttt{SymTime} outperforms Peri-midFormer, GPT4TS and TimesNet in overall performance establishing \texttt{SymTime} as the latest state-of-the-art approach. Although \texttt{SymTime}'s performance on the ETT series of datasets is not as strong as GPT4TS, it achieves more significant effects on datasets with a higher number of channels, such as ECL and Weather.

\subsection{Anomaly Detection}
\label{sec:anomaly detection}

\begin{table*}[!t]
\caption{Imputation task, where we randomly mask \{12.5\%, 25\%, 37.5\%, 50\%\} time points of length-96 time series. The reuslts averaged from 4 different mask ratios. (* means former.) See Appendix \ref{sec: imputation full results} for full results. \textbf{\textcolor{red}{Red}}: best, \textcolor{blue}{Blue}: second best.}
\centering
\vskip 0.05in
\begin{threeparttable}
\begin{footnotesize  }
\renewcommand{\multirowsetup}{\centering}
\setlength{\tabcolsep}{3.1pt}
\begin{tabular}{c|cc|cc|cc|cc|cc|cc|cc|cc|cc}
\toprule
 & \multicolumn{2}{c}{\textbf{SymTime}} & \multicolumn{2}{c}{GPT4TS} & \multicolumn{2}{c}{TimesNet} & \multicolumn{2}{c}{Peri-mid*} & \multicolumn{2}{c}{Moment} & \multicolumn{2}{c}{iTrans*} & \multicolumn{2}{c}{PatchTST} & \multicolumn{2}{c}{DLinear} & \multicolumn{2}{c}{LightTS} \\
\multirow{-2}{*}{Model} & \multicolumn{2}{c}{\textbf{(Ours)}}  & \multicolumn{2}{c}{\citeyearpar{GPT4TS}}    & \multicolumn{2}{c}{\citeyearpar{TimesNet}}   & \multicolumn{2}{c}{\citeyearpar{Peri-midFormer}}  & \multicolumn{2}{c}{\citeyearpar{Moment}} & \multicolumn{2}{c}{\citeyearpar{iTransformer}} & \multicolumn{2}{c}{\citeyearpar{PatchTST}} & \multicolumn{2}{c}{\citeyearpar{DLinear}} & \multicolumn{2}{c}{\citeyearpar{LightTS}} \\
\cmidrule(lr){1-1}
\cmidrule(lr){2-3} \cmidrule(lr){4-5} \cmidrule(lr){6-7} \cmidrule(lr){8-9} \cmidrule(lr){10-11} \cmidrule(lr){12-13} \cmidrule(lr){14-15} \cmidrule(lr){16-17} \cmidrule(lr){18-19}
Metric & MSE & MAE & MSE & MAE & MSE & MAE & MSE & MAE & MSE & MAE & MSE & MAE & MSE & MAE & MSE & MAE & MSE & MAE \\
\midrule
ETTm1 & 0.036 & 0.116 & {\color[HTML]{0000FF} 0.028} & {\color[HTML]{0000FF} 0.109} & {\color[HTML]{FF0000} \textbf{0.027}} & {\color[HTML]{FF0000} \textbf{0.107}} & 0.036 & 0.116 & 0.074 & 0.168 & 0.072 & 0.185 & 0.049 & 0.143 & 0.090 & 0.204 & 0.068 & 0.182 \\
ETTm2 & {\color[HTML]{0000FF} 0.026} & 0.088 & {\color[HTML]{FF0000} \textbf{0.022}} & {\color[HTML]{FF0000} \textbf{0.088}} & {\color[HTML]{FF0000} \textbf{0.022}} & {\color[HTML]{0000FF} 0.089} & {\color[HTML]{0000FF} 0.026} & 0.087 & 0.031 & 0.108 & 0.082 & 0.191 & 0.030 & 0.101 & 0.102 & 0.212 & 0.068 & 0.176 \\
ETTh1 & 0.095 & 0.201 & 0.093 & 0.200 & {\color[HTML]{FF0000} \textbf{0.089}} & {\color[HTML]{0000FF} 0.199} & {\color[HTML]{0000FF} 0.091} & {\color[HTML]{FF0000} \textbf{0.196}} & 0.139 & 0.234 & 0.148 & 0.269 & 0.126 & 0.231 & 0.169 & 0.283 & 0.159 & 0.278 \\
ETTh2 & 0.058 & {\color[HTML]{0000FF} 0.148} & {\color[HTML]{0000FF} 0.052} & {\color[HTML]{FF0000} \textbf{0.147}} & {\color[HTML]{FF0000} \textbf{0.050}} & {\color[HTML]{0000FF} 0.148} & 0.057 & {\color[HTML]{FF0000} \textbf{0.147}} & 0.061 & 0.159 & 0.139 & 0.254 & 0.066 & 0.164 & 0.163 & 0.273 & 0.143 & 0.258 \\
ECL & {\color[HTML]{FF0000} \textbf{0.054}} & {\color[HTML]{FF0000} \textbf{0.151}} & 0.093 & 0.212 & 0.094 & 0.211 & {\color[HTML]{0000FF} 0.063} & {\color[HTML]{0000FF} 0.169} & 0.094 & 0.211 & 0.099 & 0.224 & 0.078 & 0.192 & 0.128 & 0.256 & 0.108 & 0.238 \\
Weather & {\color[HTML]{FF0000} \textbf{0.028}} & {\color[HTML]{FF0000} \textbf{0.038}} & 0.032 & 0.058 & 0.030 & 0.056 & {\color[HTML]{0000FF} 0.029} & {\color[HTML]{0000FF} 0.041} & 0.035 & 0.075 & 0.052 & 0.114 & 0.033 & 0.057 & 0.053 & 0.116 & 0.047 & 0.106 \\
\midrule
Average & {\color[HTML]{FF0000} \textbf{0.049}} & {\color[HTML]{FF0000} \textbf{0.124}} & 0.053 & 0.136 & 0.052 & 0.135 & {\color[HTML]{0000FF} 0.050} & {\color[HTML]{0000FF} 0.126} & 0.072 & 0.159 & 0.099 & 0.206 & 0.064 & 0.148 & 0.118 & 0.224 & 0.099 & 0.206 \\
\bottomrule
\end{tabular}
\end{footnotesize  }
\end{threeparttable}
\label{table:imputation results}
\vskip -0.10in  % 后续这里记得注释掉
\end{table*}

\begin{table*}[!t]
\caption{Anomaly detection task, where we calculate the F1-score (as \%) for each dataset. (* means former.) A higher value of F1-score indicates a better performance. See Appendix \ref{sec: anomaly detection full results} for full results. \textbf{\textcolor{red}{Red}}: best, \textcolor{blue}{Blue}: second best.}
\centering
\vskip 0.05in
\begin{threeparttable}
\begin{footnotesize  }
\renewcommand{\multirowsetup}{\centering}
\setlength{\tabcolsep}{2.81pt}
\begin{tabular}{c|cccccccccccccc}
\toprule
 & SymTime & UniTS & Peri-mid* & GPT4TS & TimesNet & PatchTST & LightTS & DLinear & iTrans* & Anomaly & Stationary & Cross* & In* & Auto* \\
\multirow{-2}{*}{Model} & \textbf{(Ours)} & \citeyearpar{UniTS} & \citeyearpar{Peri-midFormer} & \citeyearpar{GPT4TS} & \citeyearpar{TimesNet} & \citeyearpar{PatchTST} & \citeyearpar{LightTS} & \citeyearpar{DLinear} & \citeyearpar{iTransformer} & \citeyearpar{Anomaly-Transformer} & \citeyearpar{Non-stationary-transformers} & \citeyearpar{Crossformer} & \citeyearpar{Informer} & \citeyearpar{Autoformer} \\
\midrule
SMD & {\color[HTML]{0000FF} 84.62} & 83.69 & 84.08 & 84.49 & 84.37 & 84.62 & 82.53 & 79.76 & 80.19 & {\color[HTML]{FF0000} \textbf{85.49}} & 82.97 & 77.22 & 77.88 & 71.17 \\
MSL & 81.77 & 81.16 & 80.68 & {\color[HTML]{0000FF} 82.03} & 81.14 & 78.70 & 78.95 & 81.87 & 72.47 & {\color[HTML]{FF0000} \textbf{83.31}} & 76.68 & 80.59 & 81.07 & 82.22 \\
SMAP & 69.87 & {\color[HTML]{FF0000} \textbf{74.00}} & 67.53 & 68.85 & 69.05 & 68.82 & 69.21 & 67.30 & 66.72 & 71.18 & 69.02 & 67.12 & 73.26 & {\color[HTML]{0000FF} 73.97} \\
SWaT & {\color[HTML]{FF0000} \textbf{93.61}} & 92.51 & 91.64 & 92.60 & 92.61 & 85.72 & {\color[HTML]{0000FF} 93.33} & 92.66 & 92.64 & 83.10 & 92.24 & 90.22 & 80.35 & 79.19 \\
PSM & 97.07 & {\color[HTML]{FF0000} \textbf{97.31}} & 96.21 & 97.09 & 97.06 & 96.08 & 97.15 & 96.64 & 94.88 & 79.40 & {\color[HTML]{0000FF} 97.23} & 92.52 & 90.43 & 88.24 \\
\midrule
Avg F1 & {\color[HTML]{0000FF} 85.39} & {\color[HTML]{FF0000} \textbf{85.73}} & 84.03 & 85.01 & 84.85 & 82.79 & 84.23 & 83.64 & 81.38 & 80.50 & 83.63 & 81.53 & 80.60 & 78.96 \\
\bottomrule
\end{tabular}
\end{footnotesize  }
\end{threeparttable}
\label{table:anomaly detection results}
\vskip -0.05in  % 后续这里记得注释掉
\end{table*}

\begin{figure*}[!t]
\begin{subfigure}{0.6\textwidth}
    \includegraphics[width=\linewidth]{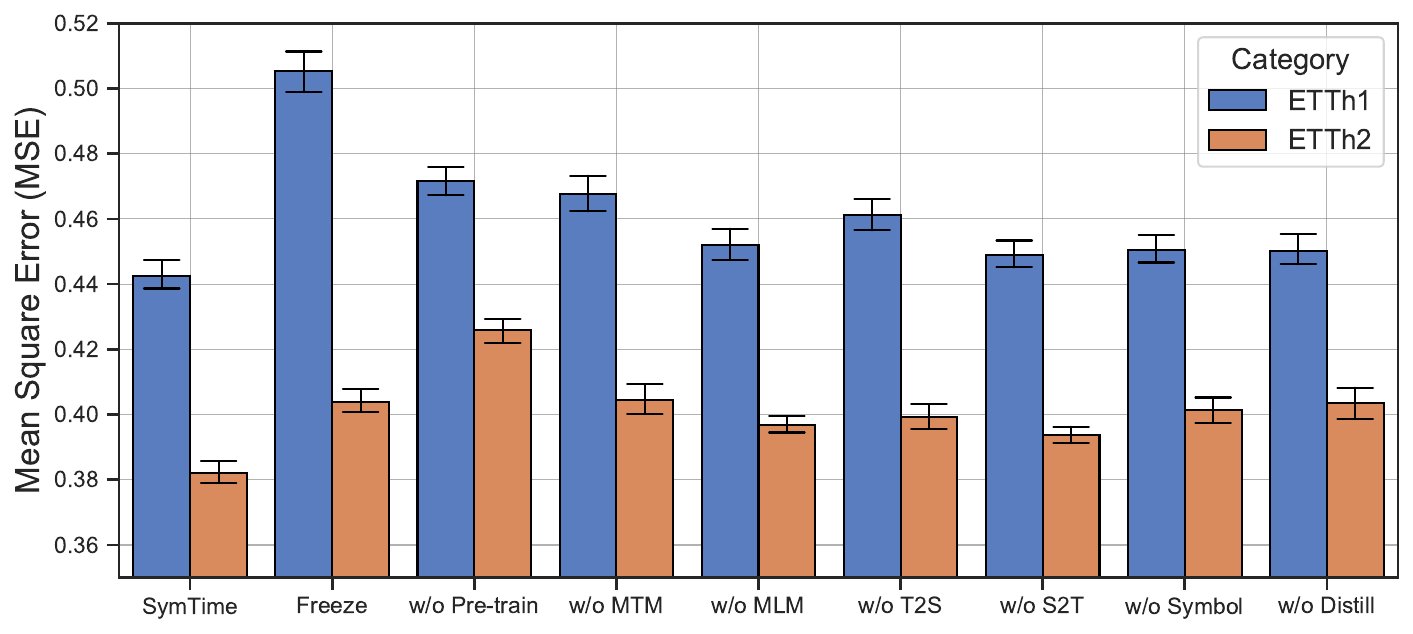}
    \caption{Ablation on pre-training objectives.}
\end{subfigure}
\hfill
\begin{subfigure}{0.4\textwidth}
    \includegraphics[width=\linewidth]{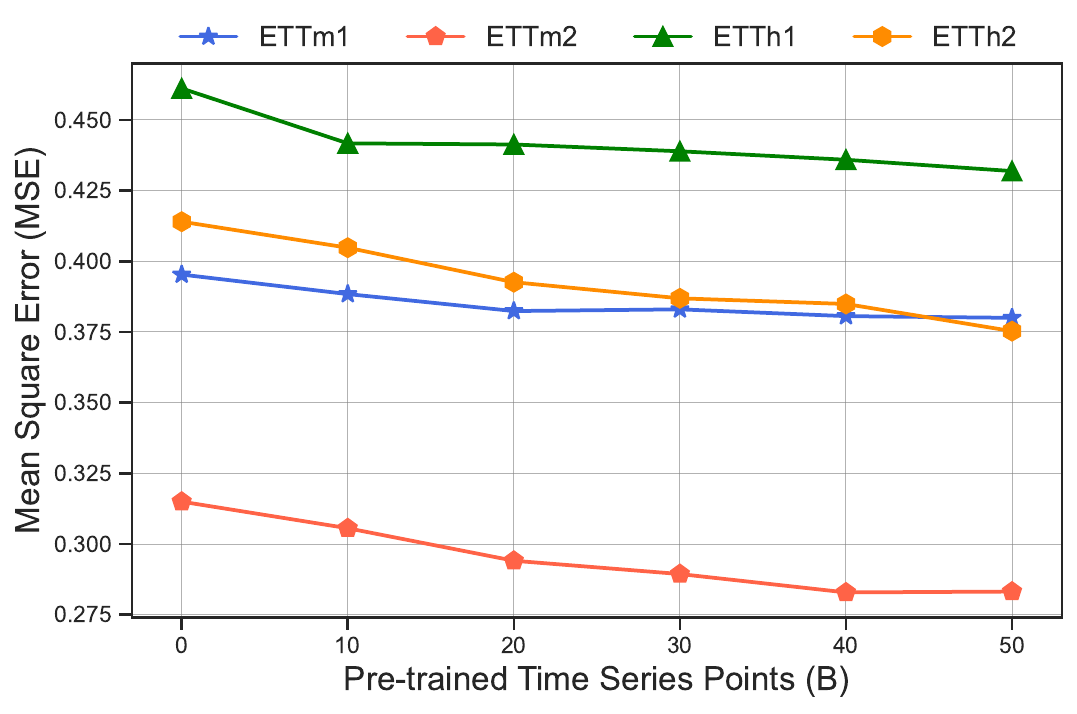}
    \caption{Ablation on the size of pre-training dataset.}
\end{subfigure}
%\vskip -0.05in
\caption{Ablation experiments on long-term forecasting task.}
\label{figure: ablation on long}
%\vskip -0.03in
\end{figure*}

\textbf{Setup.} Time series anomaly detection is crucial for rapidly identifying anomalies in critical areas, aiding in risk prevention and decision optimization. Due to the difficulty in annotating time series anomalies, we focus primarily on unsupervised anomaly detection. We conduct experiments on 5 widely used anomaly detection datasets: SMD and SMAP \cite{SMD}, MSL \cite{MSL}, SWaT \cite{SWaT}, PSM \cite{PSM}, encompassing service monitoring, space \& earth exploration, and water treatment applications. We adopt the same data preprocessing method as the Anomaly Transformer \cite{Anomaly-Transformer}, dividing the data into non-overlapping segments of length 100 for reconstruction. Specifically, normal data is used for model training and we employ a simple reconstruction loss to help the model learn the distribution of normal data \cite{Peri-midFormer}. In subsequent testing phases, reconstructed outputs exceeding a specified threshold are considered anomalies.

\textbf{Results.} Table \ref{table:anomaly detection results} indicates that \texttt{SymTime} surpasses previous state-of-the-art methods such as TimesNet and GPT4TS and achieves commendable performance on the SMD and SWaT datasets. However, due to UniTS employing real data and downstream task-relevant pre-training, there is a slight performance difference between \texttt{SymTime} and UniTS.

\subsection{Ablation Experiments}
\label{sec:ablation experiments}

\textbf{Setup.} We conduct ablation studies on \texttt{SymTime}'s pre-training objectives and the size of the pre-training dataset using the ETT long-term forecasting dataset \cite{Informer}. First, we establish 8 different control groups based on whether pre-training is performed, freezing the model and various pre-training losses: (1) Freeze, (2) w/o Pre-train, (3) w/o MTM, (4) w/o MLM, (5) w/o T2S, (6) w/o S2T, (7) w/o Symbol and (8) w/o Distill. Specific explanations for the above control groups are provided in Appendix \ref{sec:Appendix Ablation Experiments Details}. We use the average MSE of the prediction lengths \{$96, 192, 336, 720$\} on the ETTh1 and ETTh2 datasets as the evaluation metric, with the average results shown in Figure \ref{figure: ablation on long} (a). Then, we set the sizes of the pre-training datasets to \{$0, 10\mathrm{B}, 20\mathrm{B}, 30\mathrm{B}, 40\mathrm{B}, 50\mathrm{B}$\}, where 0 indicates no pre-training. We subsequently observe the changes in the model's MSE with the size of the pre-training dataset, with specific results shown in Figure \ref{figure: ablation on long} (b). We apply the same experimental configuration for ablation studies on short-term forecasting tasks, with detailed results and analysis provided in Appendix \ref{sec:ablation on short-term forecasting}.

\textbf{Results.} Figure \ref{figure: ablation on long} (a) indicates that pre-training with the standard configuration significantly enhances \texttt{SymTime}'s performance in long-term forecasting. Moreover, removing the symbolic component and relying solely on MTM losses to learn representations of time series can moderately degrade model performance. This suggests that the semantic information provided by the symbol encoder and contrastive learning improves the time series encoder's performance in long-term forecasting \cite{AutoTimes}. Additionally, eliminating any single pre-training objective can also impact model performance to some extent. Figure \ref{figure: ablation on long} (b) shows that as the size of the pre-training dataset increases, the model's performance on downstream tasks also improves, highlighting the importance of a large-scale and comprehensively representative dataset for model pre-training. However, on ETTm1 and ETTm2 datasets, a slight saturation trend is observed once the data volume reaches 40B.

\begin{figure}[!t]
    \centering
    \includegraphics[width=\linewidth]{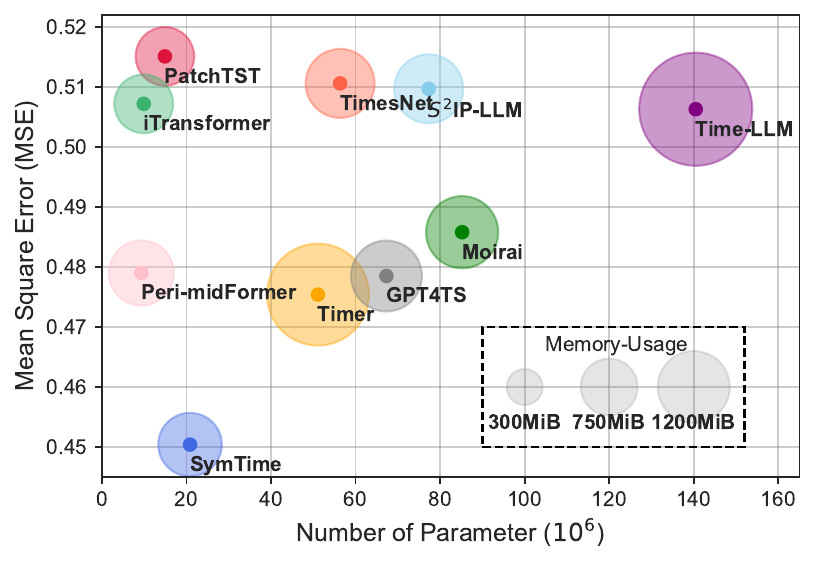}
    \vskip -0.1in
    \caption{Complexity analysis on long time series prediction tasks (ETTh1 dataset, forecasting length is 720). Note that since the original backbone of Time-LLM \cite{Time-LLM} has too many parameters, we replaced it with GPT2 \cite{GPT-2}.}
    \label{figure: complexity analysis}
    \vskip -0.03in
\end{figure}

\subsection{Discussion}
\label{sec:discussion}

%\label{sec:complexity analysis}
\textbf{Complexity Analysis.}
We analyze the complexity of the model on the long-term forecasting ETTh1 dataset, with results shown in Figure \ref{figure: complexity analysis}. We consider the parameter count, the GPU memory required for forward and backward propagation when the batch size is 1, the MSE as an evaluation metric. \texttt{SymTime} achieves better performance with a smaller model parameter count and memory capacity than existing foundation models in forecasting. Although \texttt{SymTime}'s performance is slightly lower than Time-LLM in the final experimental results of all datasets, its complexity is also significantly lower than Time-LLM. The computational load and complexity of our model are reasonable and acceptable in the vast majority of application scenarios.

\textbf{Statistical Characterization of S2 dataset.} We quantify the range of representations that the S2 dataset can cover through statistical metrics (including stationarity \cite{ADF}, forecastability \cite{forecastable}, seasonality and entropy \cite{permutation}), proving that our unrestricted data generation mechanism can evenly cover the basic representations of all types of time series, thereby solving the data scarcity problem. See Appendix \ref{sec:statistical} for full results.

\textbf{Zero-shot Imputation.} After large-scale masked time series modeling (Equation \ref{eq:mtm}), \texttt{SymTime} successfully learned the basic representation of time series and has the ability of zero-shot imputation for S2 out-of-domain data and real-world time series data. See Appendix \ref{sec: model pre-training and representation learning} for full results.

\textbf{Series-Symbol Representation Learning.} Through contrastive learning (Equation \ref{equation:L_tsc}), our time series encoder is able to distinguish series of simple unary symbolic expressions, proving that our encoder has learned the semantic information of the symbols \cite{ChatTime}. Specific results and analysis are provided in Appendix \ref{sec: model pre-training and representation learning}.

\section{Conclusion}
\label{sec:conclusion}

To address the challenges of data scarcity and distribution imbalance in time series analysis, we introduce a dual-modality data generation mechanism that enables the unrestricted creation of high-quality time series data, along with corresponding symbolic representations. Leveraging this large-scale series-symbol synthetic dataset, we propose \texttt{SymTime}, a pre-trained foundation model that integrates both time series representations and symbolic semantic information. Our pre-trained model demonstrates exceptional performance across five major TSA tasks, highlighting the effectiveness of both our data generation strategy and pre-training methodology. Looking ahead, we aim to scale up our approach by training larger models on synthetic datasets, further boosting performance on downstream tasks.

\section*{Impact Statement}

The potential value of this work lies in its ability to mitigate fundamental challenges in TSA, such as the lack of sufficient labeled data and the issue of imbalanced datasets. By generating rich, diverse, and high-quality synthetic data, our approach not only addresses these issues but also opens new avenues for improving model generalization across a wide range of applications. Furthermore, the dual-modality framework, which combines time series data with symbolic semantics, introduces a novel way of enriching the representation power of models, allowing them to better understand complex temporal dynamics and their underlying patterns.

We foresee that pre-training models on synthetic datasets, especially those that combine structured symbolic information with time series data, will become a key development trend in the TSA field. This could pave the way for more robust and scalable solutions in a variety of domains, including finance, healthcare, and climate modeling, where time series data is abundant, but labeled data is often scarce or hard to obtain.

\iffalse
\section*{Accessibility}
Authors are kindly asked to make their submissions as accessible as possible for everyone including people with disabilities and sensory or neurological differences.
Tips of how to achieve this and what to pay attention to will be provided on the conference website \url{http://icml.cc/}.

\section*{Software and Data}

If a paper is accepted, we strongly encourage the publication of software and data with the
camera-ready version of the paper whenever appropriate. This can be
done by including a URL in the camera-ready copy. However, \textbf{do not}
include URLs that reveal your institution or identity in your
submission for review. Instead, provide an anonymous URL or upload
the material as ``Supplementary Material'' into the OpenReview reviewing
system. Note that reviewers are not required to look at this material
when writing their review.

\section*{Acknowledgements}

\textbf{Do not} include acknowledgements in the initial version of
the paper submitted for blind review.

If a paper is accepted, the final camera-ready version can (and
usually should) include acknowledgements.  Such acknowledgements
should be placed at the end of the section, in an unnumbered section
that does not count towards the paper page limit. Typically, this will 
include thanks to reviewers who gave useful comments, to colleagues 
who contributed to the ideas, and to funding agencies and corporate 
sponsors that provided financial support.
\fi

\bibliography{main}
\bibliographystyle{icml2025}

\newpage
\appendix
\onecolumn

\section{Series-Symbol (S2) Pre-training Data Details}
\label{sec:appendix A}

In this section, we primarily detail the S2 dataset generation process. Section \ref{sec:sampling of functions} describes the generation of symbolic expressions. Subsequently, \ref{sec:generating inputs and outputs} explains the process of generating sampling series from mixed distributions and random ARMA series. Then, \ref{sec:series-symbol display} presents the series-symbol data we generated. Finally, in Section \ref{sec:series-symbol datasets}, we provide a detailed introduction to the usage of the S2 dataset for \texttt{SymTime} pre-training. Some of the mathematical symbols used and their explanations are shown in Table \ref{table:symbols}.

\begin{table*}[ht]
\caption{Some symbols used in data generation and their explanations.}
\label{table:symbols}
\vskip 0.15in
\begin{center}
\begin{small}
\begin{sc}
\begin{tabular}{cccc}
\toprule
Symbols      & Explanation                       & Symbols      & Explanation                     \\
\midrule
$X$            & sampling series                   & $Y$            & generated series                \\
$f(\cdot)$   & symbolic expression               & $e_t$        & white noise sequence            \\
$M$            & the input channels number         & $N$            & the output channels number      \\
$\mathcal U$ & uniform distribution              & $\mathcal N$ & normal distribution             \\
$p$            & the order of the AR process       & $q$            & the order of the MA process     \\
$\phi_p$     & the parameters of the AR process & $\theta_q$   & the parameter of the MA process \\
\bottomrule
\end{tabular}
\end{sc}
\end{small}
\end{center}
\vskip 0.10in
\end{table*}

\subsection{Sampling of Functions}
\label{sec:sampling of functions}

Since the structure of mathematical expressions is inherently tree-like \cite{SNIP}, where constants and variables can be considered leaf nodes, and binary operators can be seen as root nodes with two child nodes, while unary operators can be regarded as root nodes with a single child node \cite{Symbolic, DL4Symbolic}. Therefore, we construct a binary tree using input variables and binary operators as the basic framework for symbolic expressions \cite{SymbolicBrittleness}. Subsequently, we randomly insert unary operators within the binary tree and introduce constants through affine transformations to increase the diversity and complexity of the expressions\cite{Exhaustive, SNIP, Symbolic}. The specific process is illustrated below. The three key steps are shown in Figure \ref{figure:tree}.

\begin{figure*}[ht]
\centering
\begin{subfigure}{0.32\textwidth}
    \includegraphics[width=\linewidth]{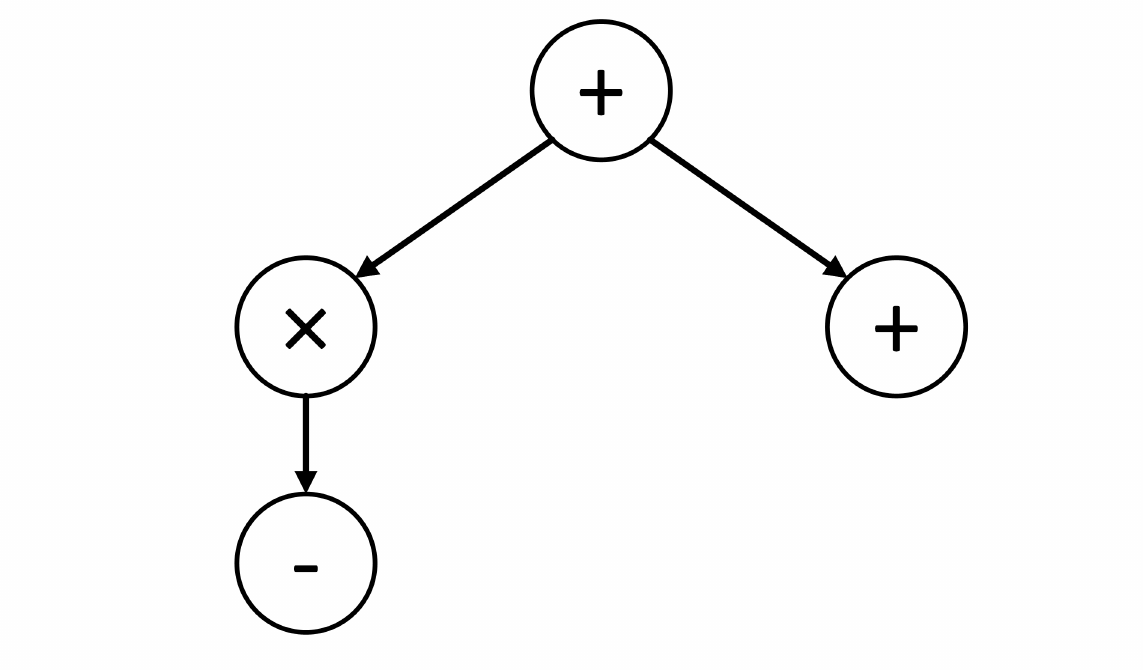}
    \caption{tree construction}
\end{subfigure}
\hfill
\begin{subfigure}{0.32\textwidth}
    \includegraphics[width=\linewidth]{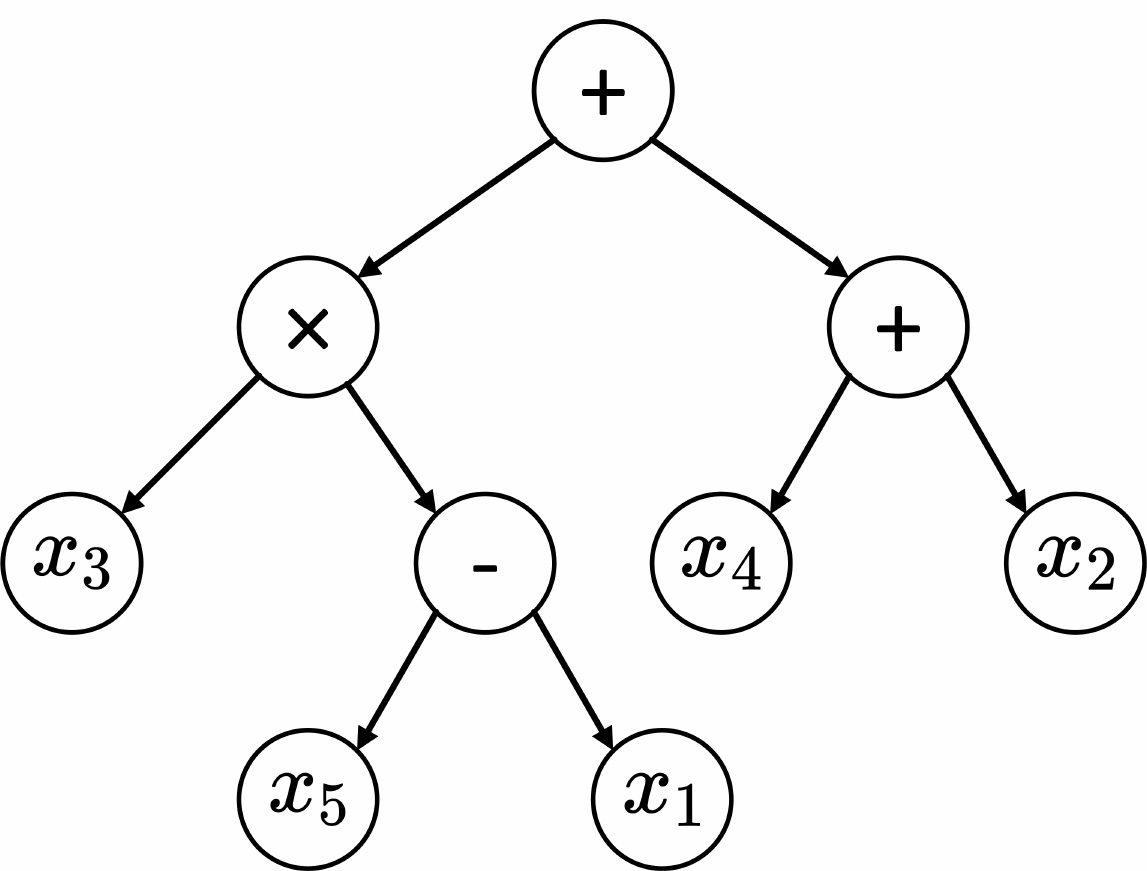}
    \caption{variable assignment to leaf nodes}
\end{subfigure}
\hfill
\begin{subfigure}{0.32\textwidth}
    \includegraphics[width=\linewidth]{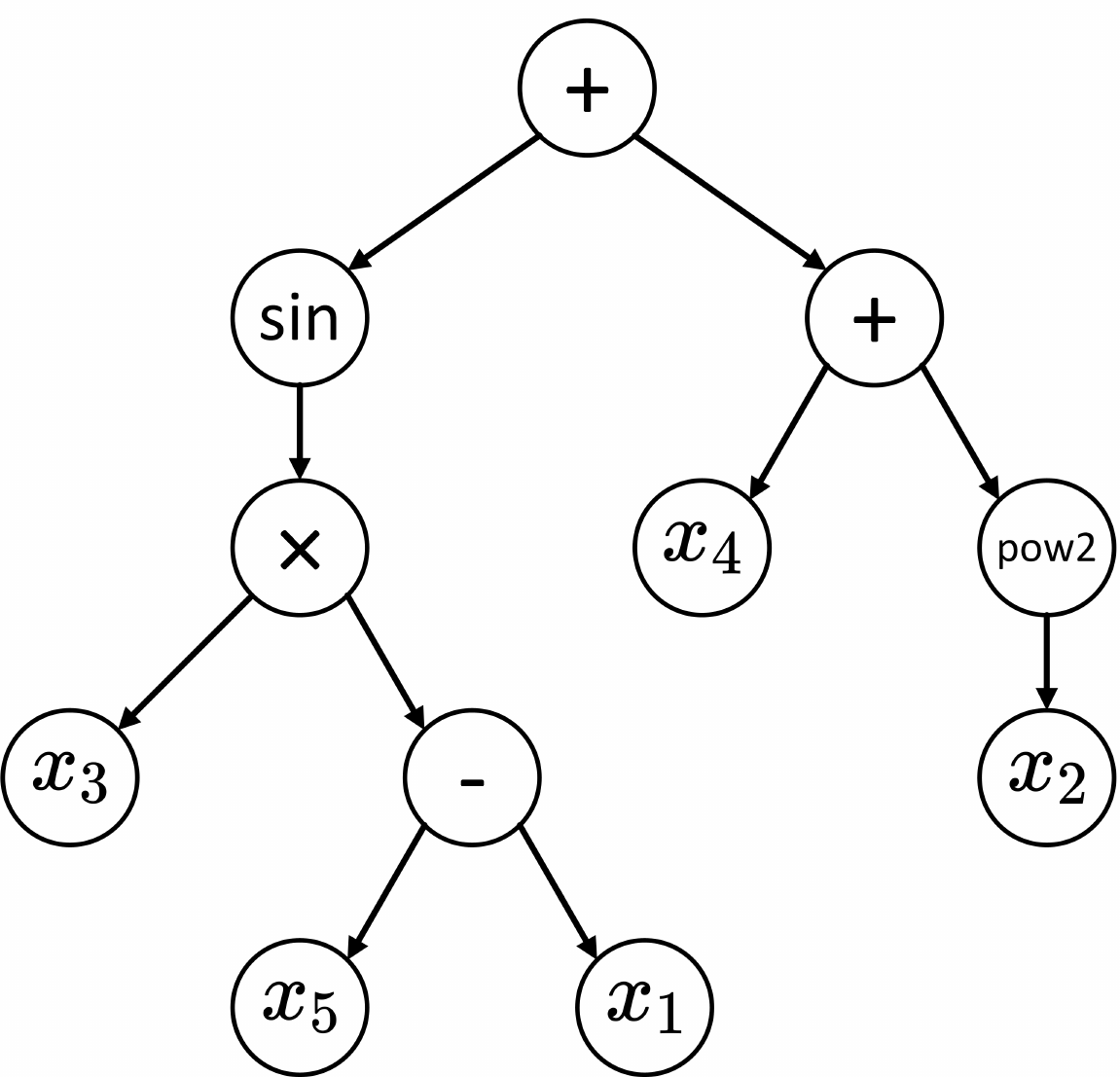}
    \caption{unary operator insertion}
\end{subfigure}
\caption{The process of building a binary tree when sampling symbolic expressions.}
\label{figure:tree}
\end{figure*}

\textbf{Input and Output Dimension Selection.} To learn the features of numerical series and symbolic expressions, previous works generated the dimensions $M$ for sampling series and $N$ for generated series from uniform distributions $\mathcal{U}(1, M_\mathrm{max})$ and $\mathcal{U}(1, N_\mathrm{max})$ \cite{SNIP, Symbolic, neurosymbolic}. However, this paper traverses $[1, M_\mathrm{max}]$ and $[1, N_\mathrm{max}]$ directly to cover representations of multivariate time series, considering $M_\mathrm{max}=6$ and $N_\mathrm{max}=12$ due to the complexity and uncertainty of data generation. An input series dimension of $M$ implies the construction of expressions with $M$ variable nodes $x_1, x_2, \cdots, x_M$. An output series dimension of $N$ indicates sampling $N$ expressions $y_i = f_i(x_1, x_2, \cdots, x_M), i=1, 2, \cdots, N$, yielding $N$ channel-related generated series \cite{DL4Symbolic}.

\textbf{Binary Operator Quantity Selection.} After determining the input and output dimensions, we sample the number of binary operators $b$ from $\mathcal{U}(b_\mathrm{min}, b_\mathrm{max})$ \cite{Symbolic}. These binary operators will serve as the root nodes of binary trees, forming the basic skeleton of the tree. Subsequently, for each binary operator node, we randomly draw the specific operation for the corresponding position from $\mathcal{U}(+, -, \times)$. This step ensures the diversity and complexity of the generated expressions \cite{SNIP, GA4SR1, GA4SR2}.

\textbf{Tree Construction and Variable Assignment to Leaf Nodes.} Based on the number of binary operators obtained through random sampling, we construct a binary tree to simulate the structure of mathematical functions \cite{SNIP}. The binary operators will act as the root nodes, forming the basic skeleton of the binary tree \cite{Symbolic, DL4Symbolic}, which is shown in Figure \ref{figure:tree} (a). After obtaining the basic skeleton of the binary tree, input variables $x_1,x_2, \cdots ,x_M$ are inserted as leaf nodes into the binary tree, ensuring that each leaf node corresponds to a variable. This process is shown in Figure \ref{figure:tree} (b).

\textbf{Unary Operator Insertion.} After inserting the leaf nodes to form a complete binary tree, we select the number of unary operators $u$ from $\mathcal{U}(u_\mathrm{min}, u_\mathrm{max})$ and insert unary operators at random positions in the binary tree. The available unary operators include $\{ \mathrm{inv,abs,pow2,pow3,sqrt,sin,cos,tan,arctan,log,exp} \}$ \cite{SNIP, Symbolic}. This process is shown in Figure \ref{figure:tree} (c).

\textbf{Affine Transformation.} To further diversify the generated symbolic expressions, we perform random affine transformations on each random variable $x_d$ and unary operator $u_d$ in the binary tree. Specifically, we replace $x_d$ and $u_d$ with $ax_d+b$ and $au_d+b$, respectively, where $a$ and $b$ are random constants \cite{SNIP, Symbolic}. For example, we perform an affine transformation on the unary operation function $\mathrm{tan}(\cdot)$ to obtain $a \times \mathrm{tan}(\cdot) + b$, where $a$ and $b$ are constants.

\subsection{Generating Inputs and Outputs Series}
\label{sec:generating inputs and outputs}

After obtaining the symbolic expression $f_i(\cdot)$, we generate a sampling series $X = [x_1, x_2, \cdots, x_M] \in \mathbb R ^{M \times L}$ and obtain the generated series $Y = [y_1, y_2, \cdots , y_N] \in \mathbb R ^{N \times L}$ through forward propagation of the symbolic expression, where $y_i=f_i(X)=f_i(x_1, x_2, \cdots ,x_M), i=1, 2, \cdots, N$. In order to ensure the quality and diversity of the data, previous works generated sampling series from mixed distributions \cite{SNIP, Symbolic, DL4Symbolic}. To make the generated series data more representative of time series, the data in this paper is not only sampled from mixed distributions but also generated from autoregressive moving average (ARMA) models \cite{ARIMA, ARMA} with random parameters. The ARMA($p$, $q$) model consists of moving average (MA) and autoregressive (AR) processes \cite{ARIMA_old}, which can be expressed as:
\begin{equation}
    {{Y}_{t}}={{\phi }_{1}}{{Y}_{t-1}}+{{\phi }_{2}}{{Y}_{t-2}}+\cdots +{{\phi }_{p}}{{Y}_{t-p}}+{{e}_{t}}-{{\theta }_{1}}{{e}_{t-1}}-{{\theta }_{2}}{{e}_{t-2}}-\cdots -{{\theta }_{q}}{{e}_{t-q}},
\end{equation}
where $p$ and $q$ represent the orders of the AR and MA models, respectively, $\phi_p$ and $\theta_q$ are the parameters of the AR and MA processes \cite{ARMA}, and $e_t \sim \mathcal N(0,1)$ denotes the observed white noise sequence. Since ARMA  possess both the temporal correlation of the AR process and the randomness of the MA process, series obtained from mixed distributions and ARMA sampling better reflect the characteristics of time series.

To ensure the quality of the generated series $Y$, if the sampling series value $x_i$ falls outside the domain of the expression $f(\cdot)$ or if the generated target value $y_i$ is excessively large, exceeding $10^4$, then the sample is discarded and resampled \cite{SNIP, Symbolic, Conformal}. This measure ensures the proper generation of data. Furthermore, for each random seed, we traverse all input and output channels to generate symbolic expressions, and each expression is sampled only once. The specific process for generating the sampling series $X$ is as follows.

\textbf{Mixture Distribution Number and ARMA($p$, $q$) Model Order Selection.} Our generated series originate from either a mixed distribution \cite{SNIP, Symbolic, Conformal} or a randomly parameterized ARMA($p$, $q$) model \cite{ARIMA}. Thus, the initial step involves randomly deciding with probability $P$ whether to employ mixed distribution sampling or the ARMA($p$, $q$) model, with this paper setting  $P \le 0.5$ for the mixed distribution and $P > 0.5$ for ARMA($p$, $q$). When opting for the mixed distribution, we select the number of distributions $k$ from the uniform distribution $\mathcal U(1, k_{\mathrm{max}})$ and determine the weights for each distribution $\{w_j \sim \mathcal U(0, 1)\}_{j=1}^{k}$, normalizing them so that $\sum_{j}w_{j} = 1$ \cite{SNIP}. For the ARMA($p$, $q$) model, we independently choose the orders of the AR and MA components, $p$ and $q$, from the uniform distributions $\mathcal U(1, p_{\mathrm{max}})$ and $\mathcal U(1, q_{\mathrm{max}})$, respectively.

\textbf{Generation of Distribution and Parameters.} When utilizing a mixed distribution, we select the mean $\mu_j$ and variance $\sigma_j$ for the $j$-th mixed distribution from $\mathcal N(0, 1)$ and $\mathcal U(0, 1)$, respectively. Ultimately, we randomly determine a Gaussian distribution $\mathcal N(\mu_j, \sigma_j^2)$ or a uniform distribution $\mathcal U(0, \mu_j)$ \cite{SNIP}. When employing the ARMA(p, q) model, we randomly generate the parameters $\theta_1, \theta_2, \ldots, \theta_q$ for the MA process from $\mathcal U(-1, 1)$. Similarly, we generate the parameters $\phi_1, \phi_2, \ldots, \phi_p$ for the AR process from the uniform distribution $\mathcal U(-1, 1)$ \cite{ARIMA}. However, to ensure the stationarity of the sampling series $X$, we need to ensure that the characteristic equation of the AR process has a stationary solution \cite{ARMA}. Specifically, we impose the following constraints on the parameters of the AR process:
\begin{equation}
    \left .{\begin{matrix}{{\phi }_{1}}+{{\phi }_{2}}+\cdots +{{\phi }_{p}}<1\\\left |{{{\phi }_{p}}}\right |<1\end{matrix}}\right \}.
\end{equation}
\textbf{Input Point Generation and Normalization.} After determining the parameters for the mixed distribution and the ARMA($p$, $q$) model, we generate the corresponding sampling series $X$ and normalize it on each dimension. Finally, we obtain the generated series $Y=f(X)$ through the symbolic expression. The length of the sampling series is $256$.

\subsection{Series-Symbol Data Display}
\label{sec:series-symbol display}

\begin{figure*}[!t]
\centering
\begin{subfigure}{0.49\textwidth}
    \includegraphics[width=\linewidth]{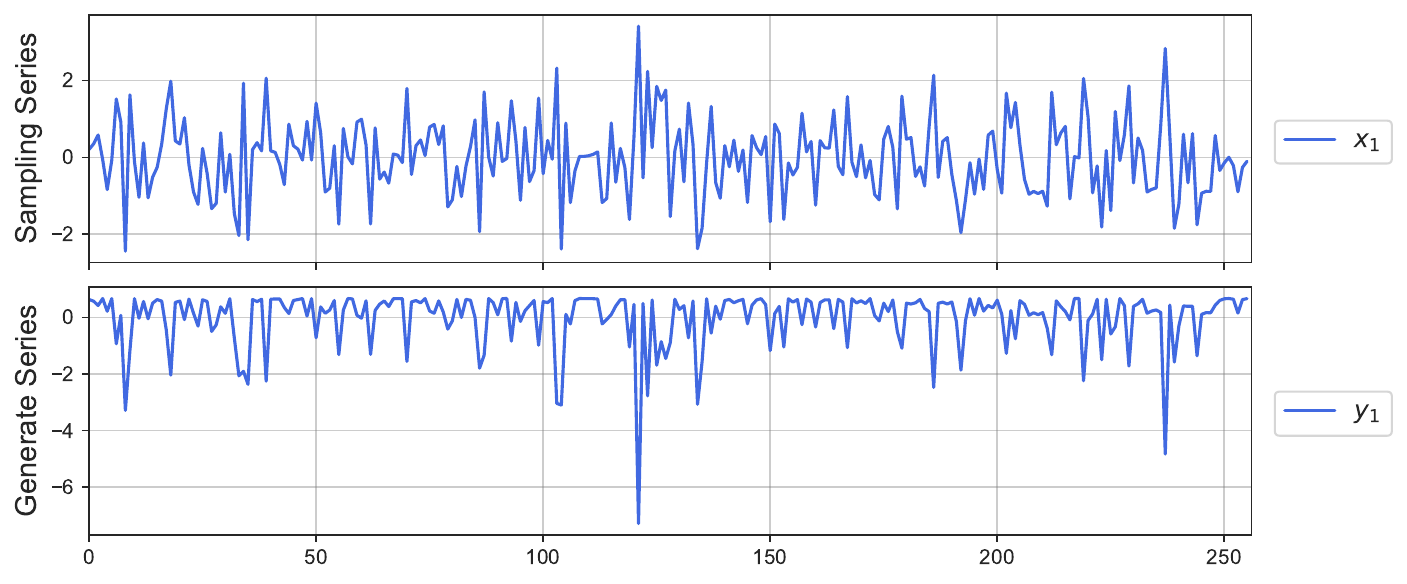}
    \caption{1 input channel 1 output channel data example 1}
\end{subfigure}
\hfill
\begin{subfigure}{0.49\textwidth}
    \includegraphics[width=\linewidth]{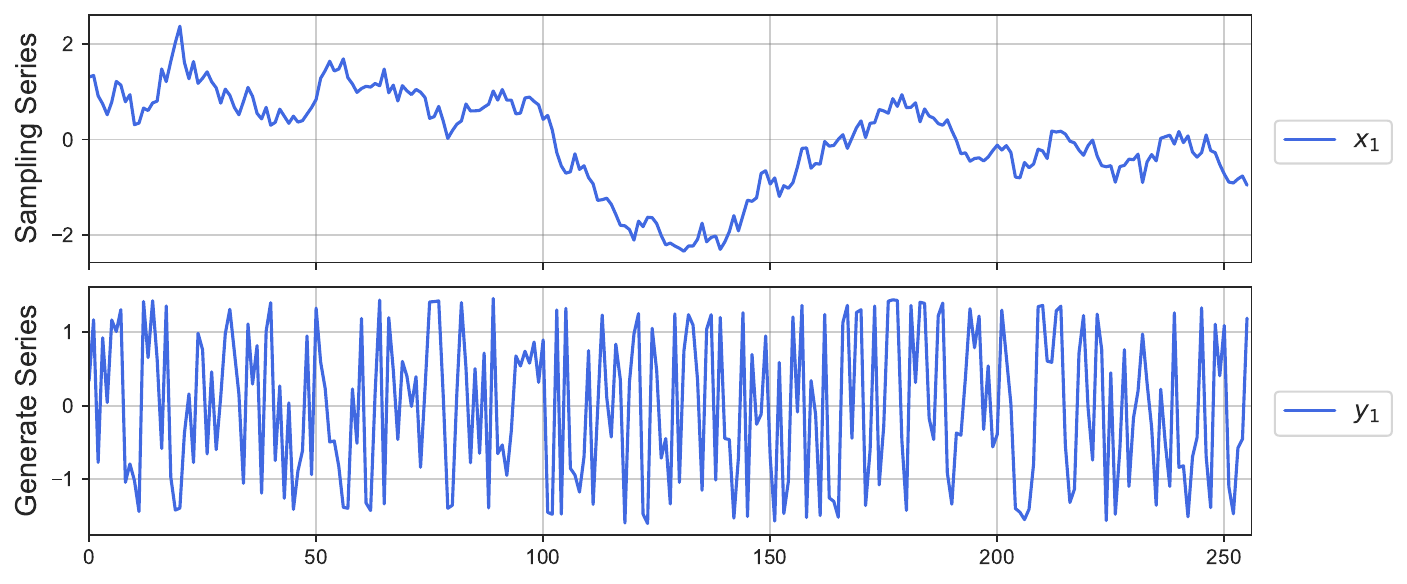}
    \caption{1 input channel 1 output channel data example 2}
\end{subfigure}
\medskip
\begin{subfigure}{0.49\textwidth}
    \includegraphics[width=\linewidth]{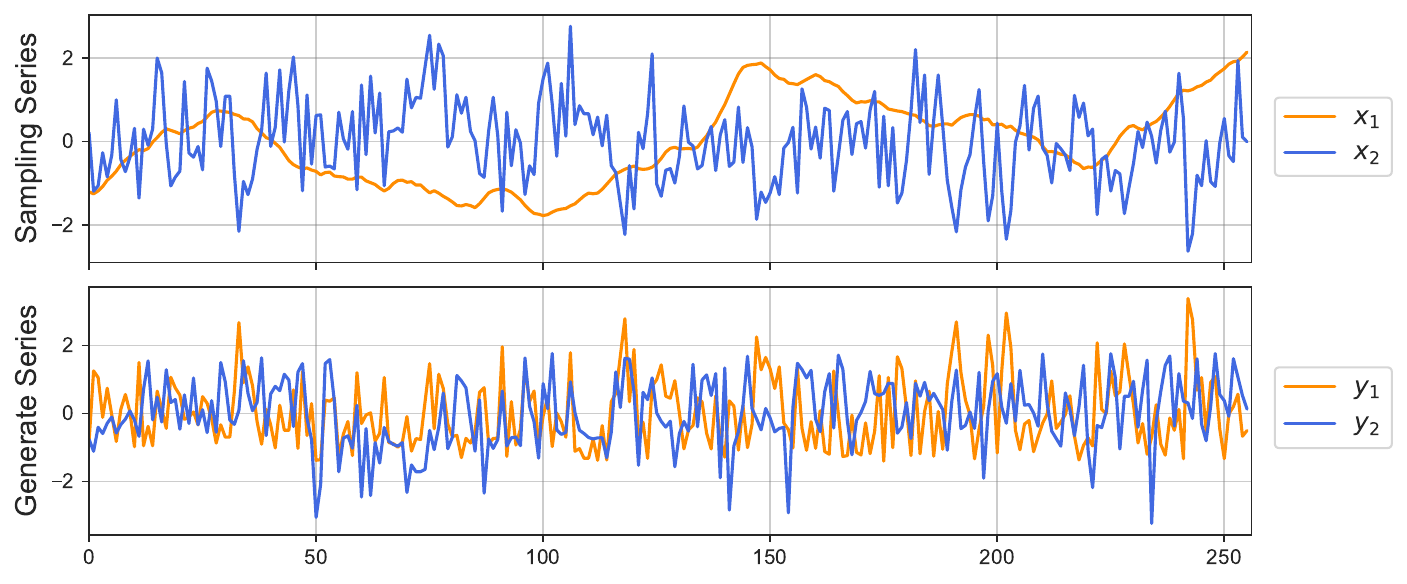}
    \caption{2 input channels 2 output channels data example 1}
\end{subfigure}
\hfill
\begin{subfigure}{0.49\textwidth}
    \includegraphics[width=\linewidth]{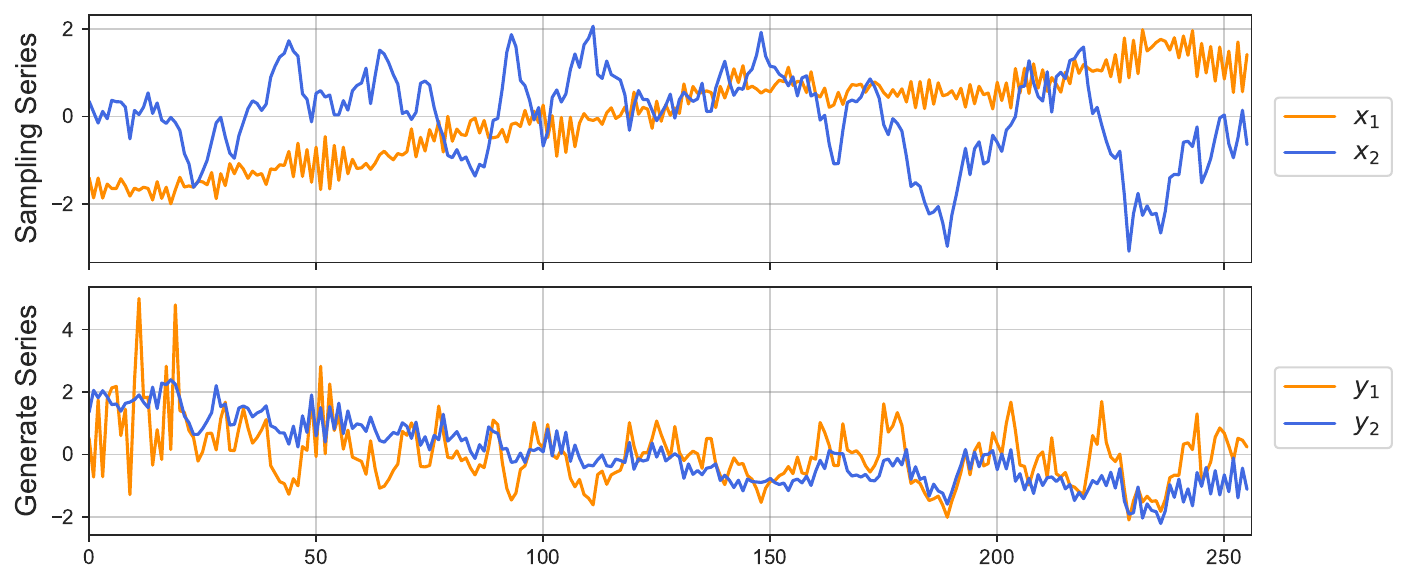}
    \caption{2 input channels 2 output channels data example 2}
\end{subfigure}
\medskip
\begin{subfigure}{0.49\textwidth}
    \includegraphics[width=\linewidth]{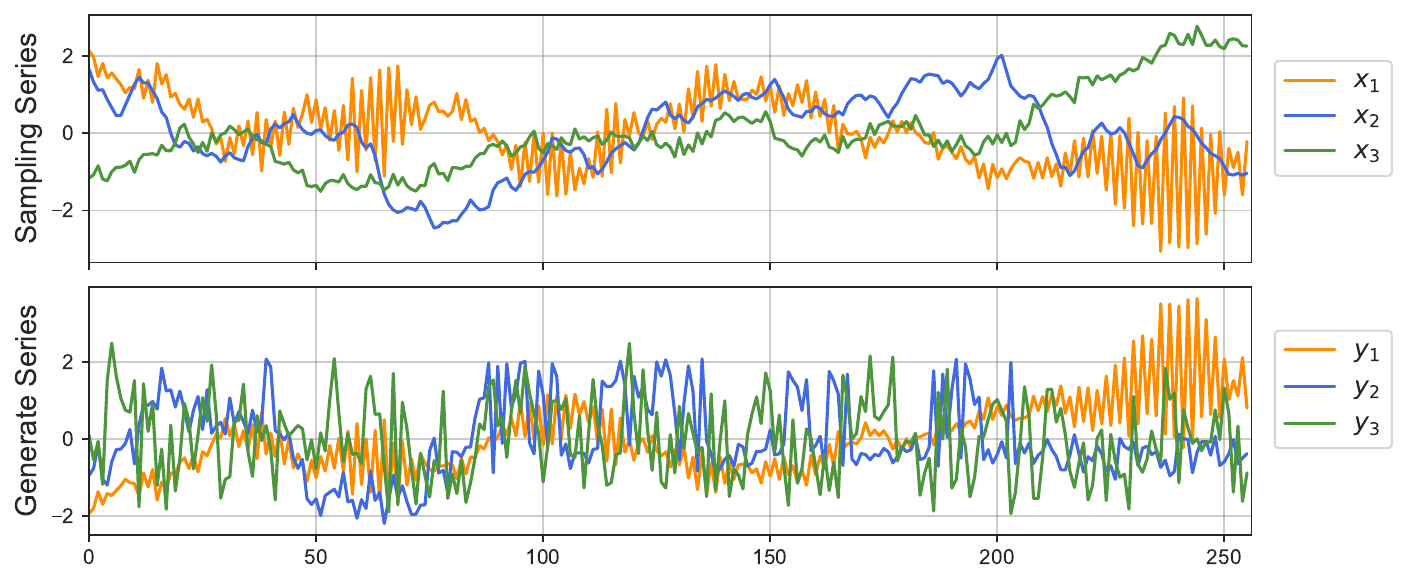}
    \caption{3 input channels 3 output channels data example 1}
\end{subfigure}
\hfill
\begin{subfigure}{0.49\textwidth}
    \includegraphics[width=\linewidth]{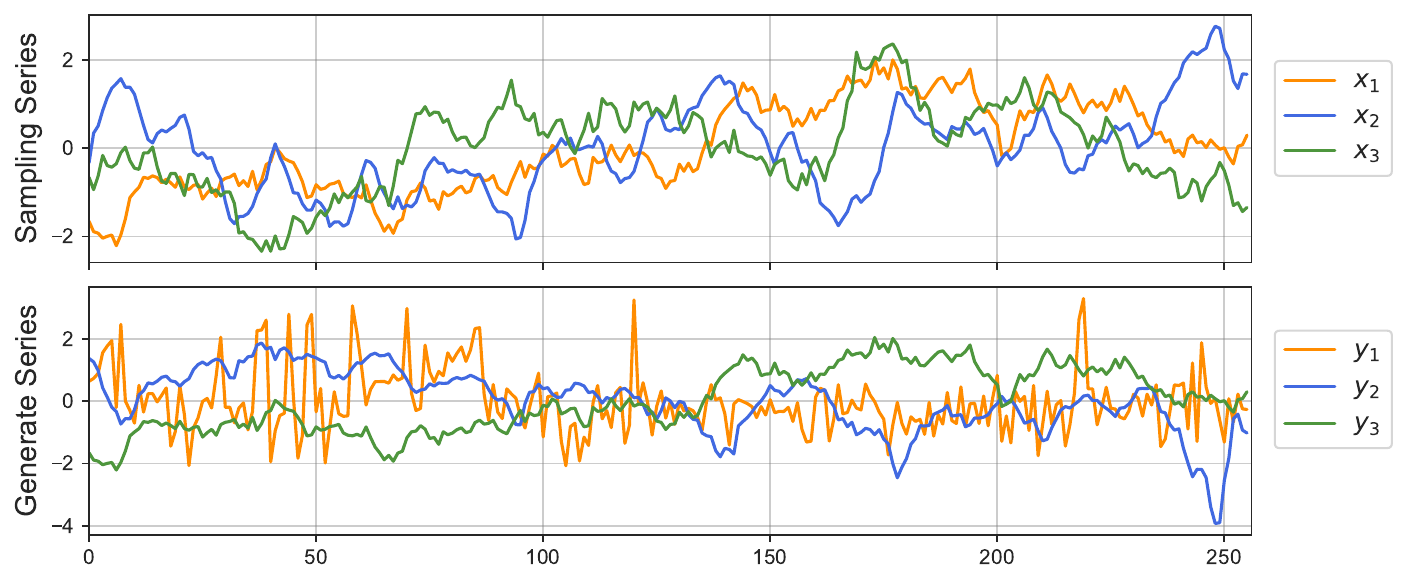}
    \caption{3 input channels 3 output channels data example 2}
\end{subfigure}
\medskip
\begin{subfigure}{0.49\textwidth}
    \includegraphics[width=\linewidth]{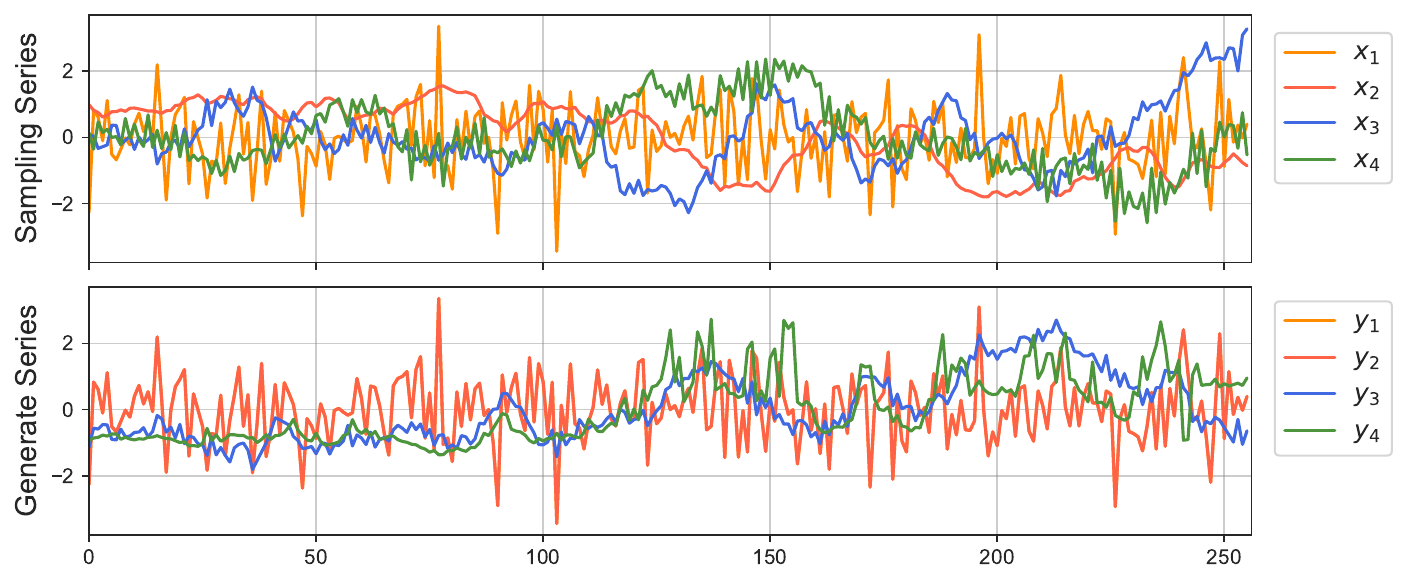}
    \caption{4 input channels 4 output channels data example 1}
\end{subfigure}
\hfill
\begin{subfigure}{0.49\textwidth}
    \includegraphics[width=\linewidth]{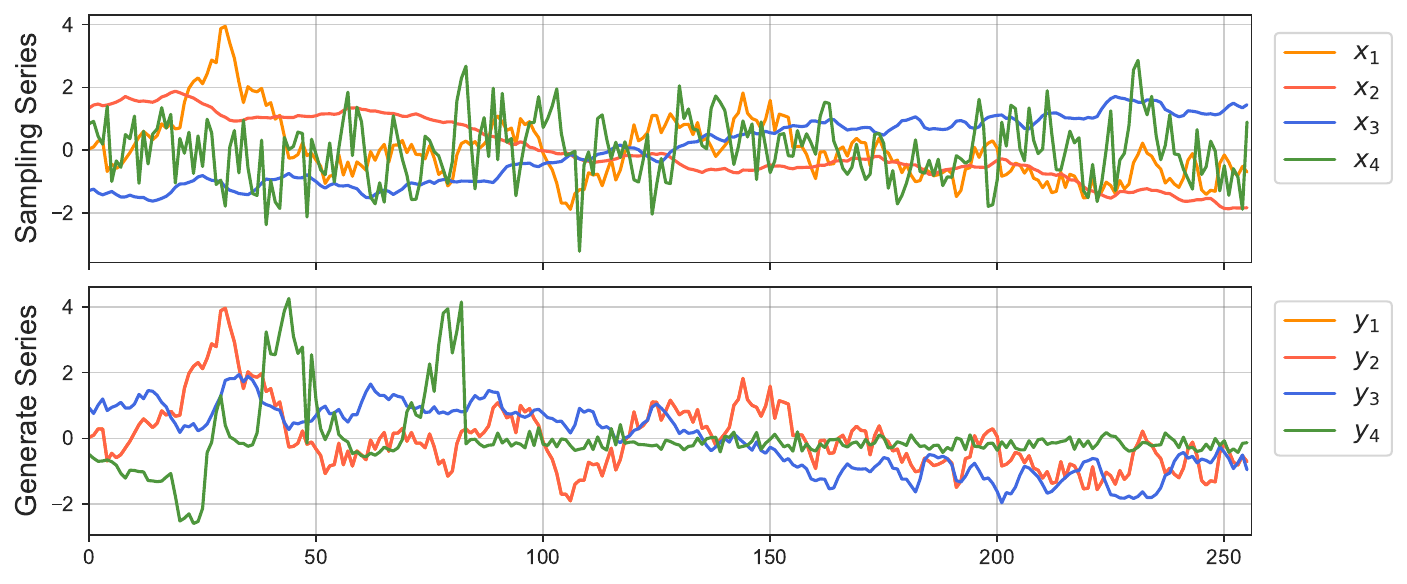}
    \caption{4 input channels 4 output channels data example 2}
\end{subfigure}
\medskip
\caption{Visualization of series from 1 input channel 1 output channel to 4 input channels 4 output channels.}
\label{figure:data plot}
\end{figure*}

\ovalbox{\begin{minipage}{\linewidth}
\ \ \ \ The symbolic expressions with text format are shown as follow: \\ 
  \textbf{Symbolic expression of Figure \ref{figure:data plot} (a)} \\
  \texttt{$y_1$ = (-0.795 add ((-0.675 mul ((0.999 add (-6.7 mul $x_1$)))**2) add ((-0.798 mul inv((-5.99 add (-0.751 mul $x_1$)))) sub (9.68 mul sqrt((-7.37 add (0.756 mul $x_1$)))))))} 
  
  \textbf{Symbolic expression of Figure \ref{figure:data plot} (c)} \\
  \texttt{$y_1$ = (-3.39 add (((0.56 mul (inv((-98.9 add (58.2 mul $x_2$))) mul ((-19.7000 mul $x_1$) sub (31.9000 mul $x_2$)))) sub (40.4000 mul $x_1$)) add (0.71 mul (((7.13 mul $x_2$) sub (-1.68 mul ($x_1$ mul sqrt((-92.8000 add (0.054 mul ($x_2$ mul ((0.327 mul $x_2$) sub (2.3 mul $x_2$))))))))) mul $x_1$))))} \\
  \texttt{$y_2$ = (1.0 add ((68.9 mul $x_2$) sub (((80.9 mul ($x_1$ mul ($x_1$ mul ((6.1000 mul $x_2$) sub ((34.2 mul sqrt((64.4 add (29.2000 mul $x_1$)))) add (-5.24 mul $x_1$)))))) add (6.78 mul $x_2$)) sub (((0.5730 mul $x_1$) sub ((2.34 mul $x_2$) sub (-6.72 mul $x_1$))) add (0.966 mul sqrt((76.8000 add (-7.79 mul $x_1$))))))))}  
  
  \textbf{Symbolic expression of Figure \ref{figure:data plot} (e)} \\
  \texttt{$y_1$ = (0.795 add ((0.42 mul $x_3$) sub ((4.39 mul $x_1$) add (((0.1430 mul $x_2$) sub ((-5.28 mul $x_3$) add (((-0.028 mul ((((1.27 mul $x_3$) sub (((0.331 mul $x_2$) sub ((2.99 mul $x_3$) add (-0.932 mul (((0.606 mul $x_1$) sub (0.967 mul $x_3$)) mul $x_3$)))) sub (-0.609 mul $x_3$))) add (-1.25 mul $x_1$)) mul $x_1$)) sub (77.3000 mul $x_1$)) sub (1.93 mul $x_3$)))) sub (16.7 mul $x_3$)))))} \\
  \texttt{$y_2$ = (-9.2900 add ((0.398 mul ((((-49.7 mul $x_1$) sub ((5.93 mul sin((6.54 add (-0.045 mul $x_1$)))) add ((62.3000 mul inv(((0.138 mul $x_2$) add (29.0 mul $x_1$)))) add (8.75 mul $x_2$)))) add ((-0.9500 mul $x_3$) add (-8.1 mul $x_1$))) mul $x_3$)) add ((-9.74 mul $x_3$) add ((((-0.9 mul $x_3$) sub (4.45 mul sqrt((-0.373 add (-0.151 mul $x_3$))))) add (-54.6 mul $x_3$)) sub (-0.758 mul ((85.3000 add (8.74 mul $x_3$)))**2)))))} \\ \texttt{$y_3$ = (-0.975 add ((-54.4000 mul sqrt((-0.722 add (-9.33 mul $x_2$)))) sub (1.45 mul ((66.4 add (-9.65 mul $x_1$)))**2)))}

  \textbf{Symbolic expression of Figure \ref{figure:data plot} (g)} \\
  \texttt{$y_1$ = (-7.17 add (0.537 mul $x_1$))} \\
  \texttt{$y_2$ = (-0.843 add (48.8000 mul $x_1$))} \\
  \texttt{$y_3$ = (57.3000 add (((-0.449 mul $x_2$) add (-1.32 mul $x_3$)) add ((-0.9400 mul $x_4$) add (0.51 mul $x_1$))))} \\
  \texttt{$y_4$ = (-0.2040 add (((-6.6000 mul inv((0.88 add (58.1 mul $x_4$)))) sub ((-23.0 mul $x_4$) add ((-91.0 mul $x_3$) sub (-93.6000 mul $x_2$)))) sub ((-6.6000 mul $x_4$) sub (0.9580 mul (($x_3$ mul $x_3$) mul ((-0.45 mul $x_2$) sub ((((-9.09 mul $x_4$) sub ((8.93 mul sqrt(((-26.6 mul $x_4$) add (-0.907 mul $x_1$)))) add (-6.2 mul $x_4$))) sub (-0.078 mul $x_4$)) sub (-16.5 mul $x_2$))))))))} \\
\end{minipage}
}

In Figure \ref{figure:data plot}, we show the visualization of the generated series from 1 input channel and 1 output channel to 4 input channels and 4 output channels. We show two sets of cases for each input and output channel. The symbolic expressions $f(\cdot)$ for the generated series in (a), (c), (e) and (g) in Figure \ref{figure:data plot} are shown above.

\subsection{Composition and Usage of the Series-Symbol Dataset}
\label{sec:series-symbol datasets}

\begin{figure*}[ht]
\centerline{\includegraphics[width=0.75\linewidth]{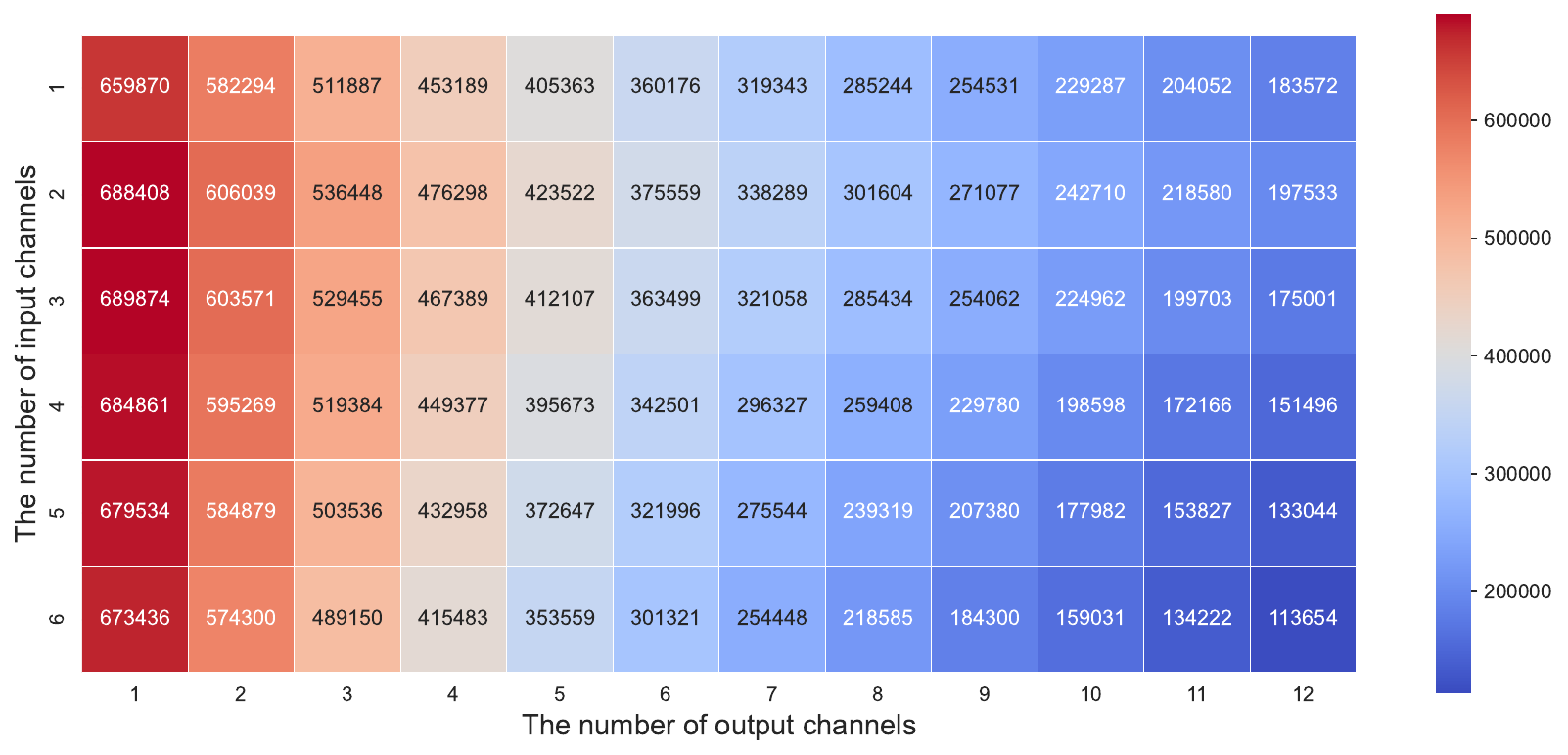}}
\caption{The number of samples in each part of the S2 dataset.}
\label{figure:heatmap}
\end{figure*}

We set the maximum number of input channels and the maximum number of output channels to 6 and 12 respectively to generate symbolic expressions and series. Each symbolic expression is sampled only once. We generated a total of 25M pairs of series and symbols. The cumulative series length is 50B. The data number of each input channel and output channel in the dataset is shown in Figure \ref{figure:heatmap}.

When pre-training \texttt{SymTime} with S2 dataset, we start by combining the sampled and generated series and then segmenting them into patches using a sliding window \cite{PatchTST, Time-LLM}. The sliding window's kernel size and step size are both set to 16. Due to the requirement for mask time series modeling (MTM) \cite{SimMTM, HiMTM}, there is no overlap between adjacent patches. Given the varying number of input and output channels in the data, the series from the maximum input and output channels can be segmented into up to 288 patches ($18\times 256 / 16$) \cite{video-transformer}. For series with fewer than 288 patches, we pad them with zeros to align the length. Next, for symbolic expressions in natural language form \cite{BERT, SNIP, S2IP-LLM, Autoformer, ChatTime}, we set a maximum length of 512 characters and perform tokenization. Ultimately, the time series patches and natural language tokens are fed into the time series encoder and the LLM of the Transformer architecture, respectively.

\section{Analysis of Series-Symbol (S2) Dataset and Model Pre-training}
\label{sec:appendix B}

In Appendix \ref{sec:appendix A}, we provide a detailed introduction to the generation process, composition and usage of the S2 dataset \cite{SNIP, Symbolic, DL4Symbolic}. In this section, we first conduct a random sampling analysis of the statistical characteristics of the S2 dataset, including stationarity \cite{ADF} and predictability \cite{forecastable, Timer}. Then, we present the composition of four existing large-scale time series pre-training datasets. It is evident that all current real-world time series datasets face significant data imbalance issues, leading to performance biases in the foundation models pre-trained on them. However, the S2 data generation method provided in this paper can unrestrictedly supply high-quality time series data, thus covering almost all time series representations. 

\subsection{Statistics Analysis}
\label{sec:statistics analysis}

\begin{table*}[ht]
\caption{Results of the stationarity and forecastability tests for the S2 dataset.}
\label{table:stats}
\vskip 0.10in
\begin{center}
\begin{small}
\begin{sc}
\begin{tabular}{cccccccccc}
\toprule
inputs & outputs & ADF    & p value & forecast & inputs & outputs & ADF    & p value & Forecastability \\
\midrule
1     & 1      & -12.77 & 0.0538  & 0.3155   & 1              & 6               & -11.48 & 0.0619  & 0.3375   \\
2     & 2      & -11.89 & 0.0568  & 0.3199   & 2              & 6               & -11.46 & 0.0733  & 0.3218   \\
3     & 3      & -12.40  & 0.0544  & 0.3328   & 3              & 6               & -11.43 & 0.0625  & 0.3244   \\
4     & 4      & -11.66 & 0.0617  & 0.3491   & 4              & 6               & -11.53 & 0.0640   & 0.3428   \\
5     & 5      & -11.38 & 0.0628  & 0.3140    & 5              & 6               & -12.32 & 0.0597  & 0.3284   \\
6     & 6      & -12.43 & 0.0625  & 0.3262   & 6              & 8               & -11.52 & 0.0555  & 0.3246   \\
6     & 10     & -11.65 & 0.0619  & 0.3287   & 6              & 12              & -11.66 & 0.0520  & 0.3310   \\
\bottomrule
\end{tabular}
\end{sc}
\end{small}
\end{center}
% \vskip 0.10in
\end{table*}

\textbf{Stationarity.} Stationarity is one of the fundamental properties of time series \cite{ARIMA, ARMA}. This attribute ensures that the statistical characteristics of time series data remain consistent across different time points, which is crucial for building effective predictive models and making reliable statistical inferences. To this end, we employ the Augmented Dickey-Fuller (ADF) \cite{ADF} test to examine the stationarity of the data, thereby determining whether the generated S2 dataset is suitable for deep neural networks (DNNs) to learn representations of time series.

\textbf{Forecastability.} The forecastability of a time series refers to the ability and accuracy to forecast future values based on historical data and statistical models \cite{ARIMA_old, Timer}. For certain specific time series and complex systems, such as stock markets, it is often challenging to predict their subsequent developments. Therefore, it is necessary to test whether the S2 dataset is non-chaotic and learnable. Forecastability is calculated by subtracting the entropy of the series' Fourier decomposition as adopted from \cite{forecastable} and \cite{Timer}, where a higher forecastability value indicates better predictability. Please note that since the method provided by \cite{forecastable} is only applicable to multivariate time series, we merge the input channels and output channels together for calculation.

\textbf{Test Methods and Results.} For the multiple input-output channels presented in the Table \ref{table:stats}, we randomly selected 1,000 samples to calculate their average ADF statistics, p-values, and Forecastability metrics. The results indicate that the average p-value from the ADF test across all samples is greater than 0.05, suggesting that the majority of the generated series in the S2 dataset are non-stationary time series, posing a challenge in modeling and learning \cite{ADF}. However, the Forecastability metric, which is greater than 0.3 for all tested samples, indicates that the generated series $Y$ is not produced by a chaotic system and is, overall, predictable.

\subsection{Analysis of Existing Large-scale Datasets for Time Series Pre-training}
\label{sec:Analysis of Existing Dataset}

\begin{table}[!t]
\caption{Time-300B time series dataset from Time-MoE \cite{Time-MoE}.}
\label{table:time-300}
\vskip 0.05in
% \begin{center}
% \begin{small}
% \begin{sc}
\begin{tabular}{ccccccccccc}
\toprule
\multicolumn{1}{l}{} & \textbf{Energy} & \textbf{Finance} & \textbf{Health} & \textbf{Nature}   & \textbf{Sales}   & \textbf{Synthetic} & \textbf{Transport} & \textbf{Web}    & \textbf{Other}   & \textbf{Total}   \\ 
\midrule
\# Obs.              & 15.98B & 413.70K & 471.04K    & 279.72B & 26.38M  & 9.22B     & 2.13B     & 1.80B  & 20.32M  & 309.09B \\
\%                   & 5.17\% & 5.17\%  & 0.0001\%   & 90.50\%  & 0.008\% & 2.98\%    & 0.69\%    & 0.58\% & 0.006\% & 100\%  \\
\bottomrule
\end{tabular}
% \end{sc}
% \end{small}
% \end{center}
\end{table}

\begin{table}[!t]
\caption{UTSD time series dataset from Timer \cite{Timer}, where Envir. means Environment, Trans. means Transport, Fin. means Finance, Mise. means Multiple Sources.}
\vskip 0.05in
\begin{center}
\begin{tabular}{cccccccccccc}
\toprule
\multicolumn{1}{l}{} & \textbf{Energy}  & \textbf{Envir.} & \textbf{Health}  & \textbf{IoT}     & \textbf{Nature}  & \textbf{Trans.} & \textbf{Web}     & \textbf{Cloud} & \textbf{Sales}   & \textbf{Fin.} & \textbf{Mise.}   \\
\midrule
\# Obs.              & 16.86B  & 70.45M      & 233.M & 165M  & 201B & 4.9B      & 157M & 2.15B    & 198M & 0.33M   & 56.52M  \\
\%                   & 7.461\% & 0.031\%     & 0.103\% & 0.073\% & 89\% & 2.17\%    & 0.07\%  & 0.95\%   & 0.088\% & 0.00\%  & 0.025\% \\
\bottomrule
\end{tabular}
\end{center}
\label{table:UTSD}
\end{table}

\begin{table}[!t]
\caption{LOTSA time series dataset from Moirai \cite{MOIRAI}.}
\label{table:LOTSA}
\vskip 0.05in
\begin{tabular}{ccccccccccc}
\toprule
\multicolumn{1}{l}{} & \textbf{Energy}  & \textbf{Transport} & \textbf{Climate} & \textbf{CloudOps} & \textbf{Web}     & \textbf{Sales}   & \textbf{Nature} & \textbf{Finance} & \textbf{Health} & \textbf{Total}  \\
\midrule
\# Obs.              & 16.36B  & 4.90B     & 4.19B   & 1.52B    & 428M & 198M & 28.55M & 24.92M   & 1.59M      & 27.65B \\
\%                   & 59.17\% & 17.73\%   & 15.15\% & 5.49\%   & 1.55\%  & 0.72\%  & 0.09\% & 0.10\%   & 0.01\%     & 100\%  \\
\bottomrule
\end{tabular}
\end{table}

\begin{table}[!t]
\caption{Time series datasets from neural scaling laws \cite{Time-Scaling-Laws}}
\vskip 0.05in
\begin{center}
\begin{tabular}{ccccccccc}
\toprule
\multicolumn{1}{l}{} & \textbf{Transport} & \textbf{Climate} & \textbf{Energy}  & \textbf{CloudOps} & \textbf{Health} & \textbf{Sales}  & \textbf{Web}    & \textbf{Total}  \\
\midrule
\# Obs.              & 4.82B     & 4.73B   & 2.34B   & 2.15B    & 240M   & 140M   & 600M   & 14.46B \\
\%                   & 33.31\%   & 32.71\% & 16.15\% & 14.86\%  & 1.61\% & 0.96\% & 0.40\% & 100\%  \\
\bottomrule
\end{tabular}
\end{center}
\label{table:scaling laws}
\end{table}

\textbf{Large-scale datasets are crucial for building foundation models.} Almost all deep learning models today are data-driven, relying on training data \cite{Informer, InceptionTime, MAE, Defenders}. Therefore, when constructing a pre-trained foundation model for time series, a large-scale and comprehensively representative pre-training dataset is indispensable \cite{MOIRAI, Timer, BEiT, CLIP, SimCLR}. The scaling laws of neural networks indicate that the learning effectiveness of deep neural networks is primarily influenced by three factors: the number of model parameters, the size of the training dataset, and the amount of computational resources \cite{Neural-Sclaing-Laws, Time-Scaling-Laws, scaling-in-llm, Scaling-Laws-for-Generative}. Expanding the scale of the pre-training dataset can effectively improve the model's generalization capability and performance, and the performance gains from increasing data volume are independent of the model architecture and training methods \cite{Time-Scaling-Laws, Data-driven-weather, ECG-LLM, ALERT}. Consequently, an increasing number of models are adopting the approach of training larger-scale models on large-scale pre-training datasets to achieve better performance \cite{graph-forecasting}. This paper surveys the pre-training datasets used by the three current mainstream pre-trained foundation models—Time-MoE \cite{Time-MoE}, Moirai \cite{MOIRAI}, and Timer \cite{Timer}—as well as the datasets utilized in the study of time series scaling laws \cite{Time-Scaling-Laws}, which are shown in Tables \ref{table:time-300}, \ref{table:UTSD}, \ref{table:LOTSA} and \ref{table:scaling laws}.

\textbf{Imbalanced domain distribution issues in large-scale time series datasets.} The distribution of data across various domains indicates that the four large-scale time series pre-training datasets all face issues with imbalanced domain data distribution. For instance, domains such as Nature, Energy and Transport have the most datasets \cite{Informer}, while others like Sales, IoT, Web, Finance and Multiple Sources suffer from extremely low data volumes due to difficulties in data collection or data privacy concerns. According to the scaling laws of neural networks, the imbalance in the pre-training dataset distribution can lead to significant performance biases in in-domain and out-of-domain forecasting tasks for the trained foundation models \cite{Time-Scaling-Laws, graph-forecasting}, meaning there is a considerable performance gap between domains with less data and those with more data. To address this, this paper proposes an unrestricted method for generating high-quality time series data to alleviate the scarcity and imbalanced distribution of data in time series analysis domains.

\subsection{S2 Dataset Statistical Characterization Coverage Experiments}
\label{sec:statistical}

\begin{figure*}[!t]
\centering
\begin{subfigure}{0.33\textwidth}
    \includegraphics[width=\linewidth]{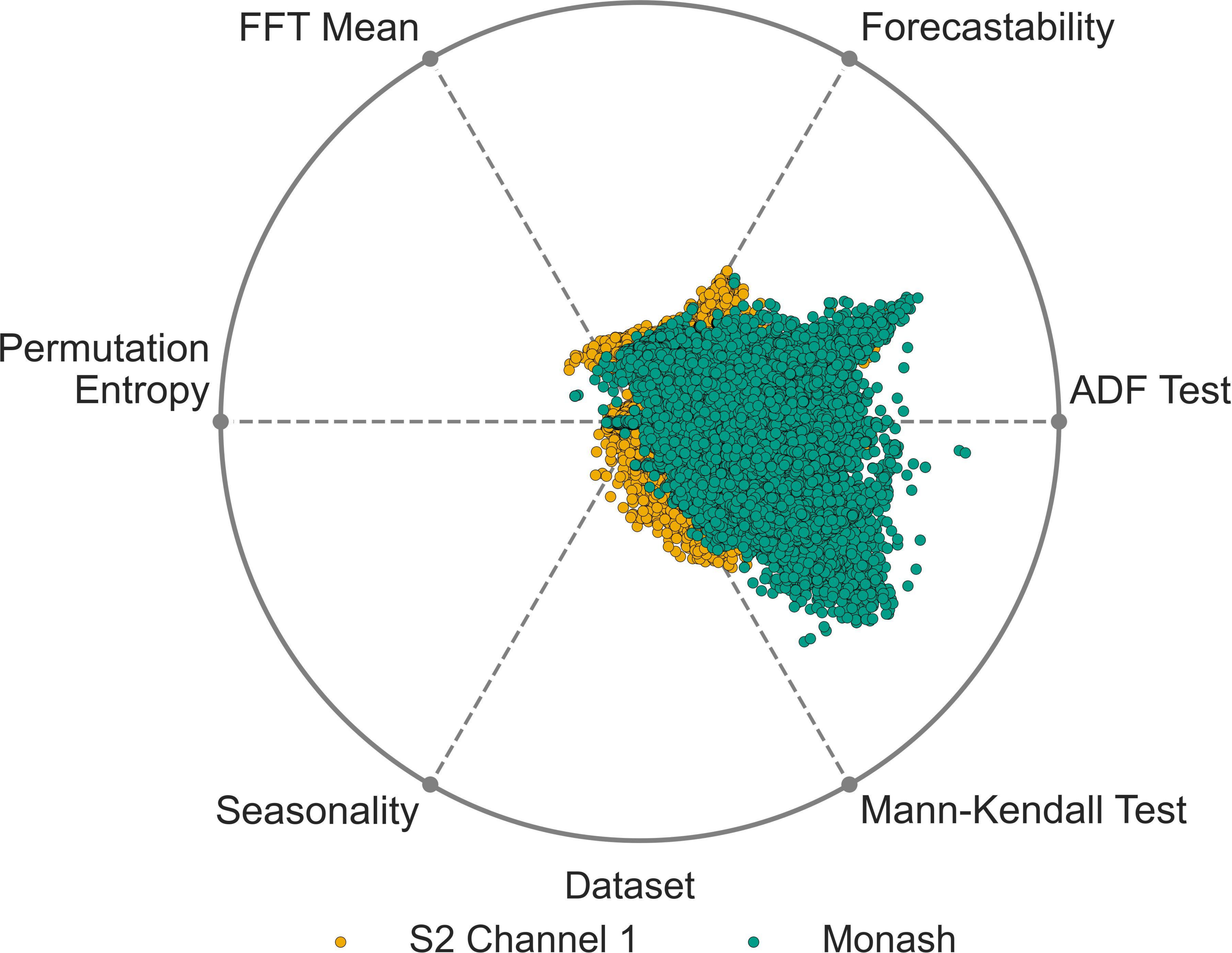}
    \caption{$100K$ single input channel samples}
\end{subfigure}
\hfill
\begin{subfigure}{0.33\textwidth}
    \includegraphics[width=\linewidth]{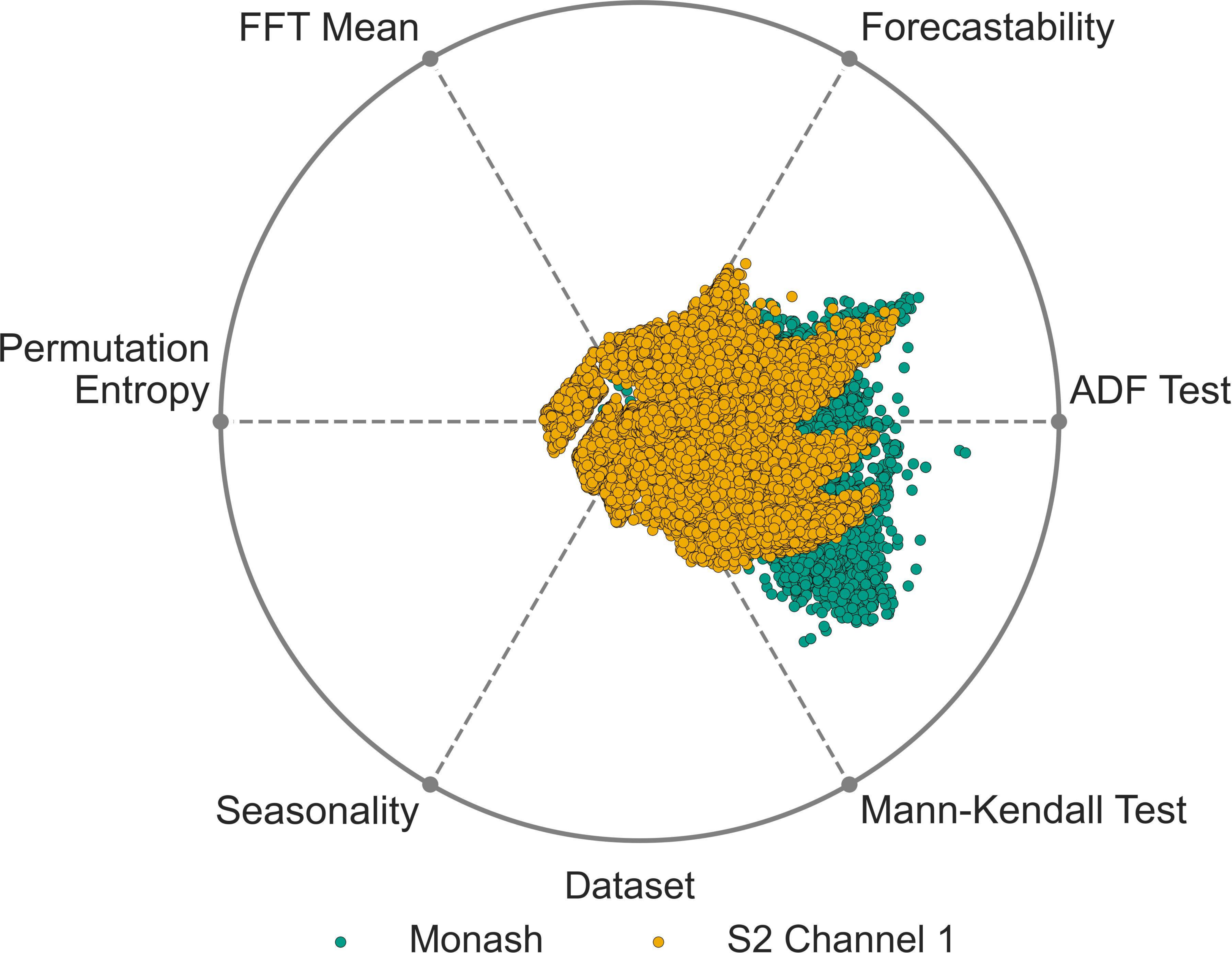}
    \caption{$200K$ single input channel samples}
\end{subfigure}
\hfill
\begin{subfigure}{0.33\textwidth}
    \includegraphics[width=\linewidth]{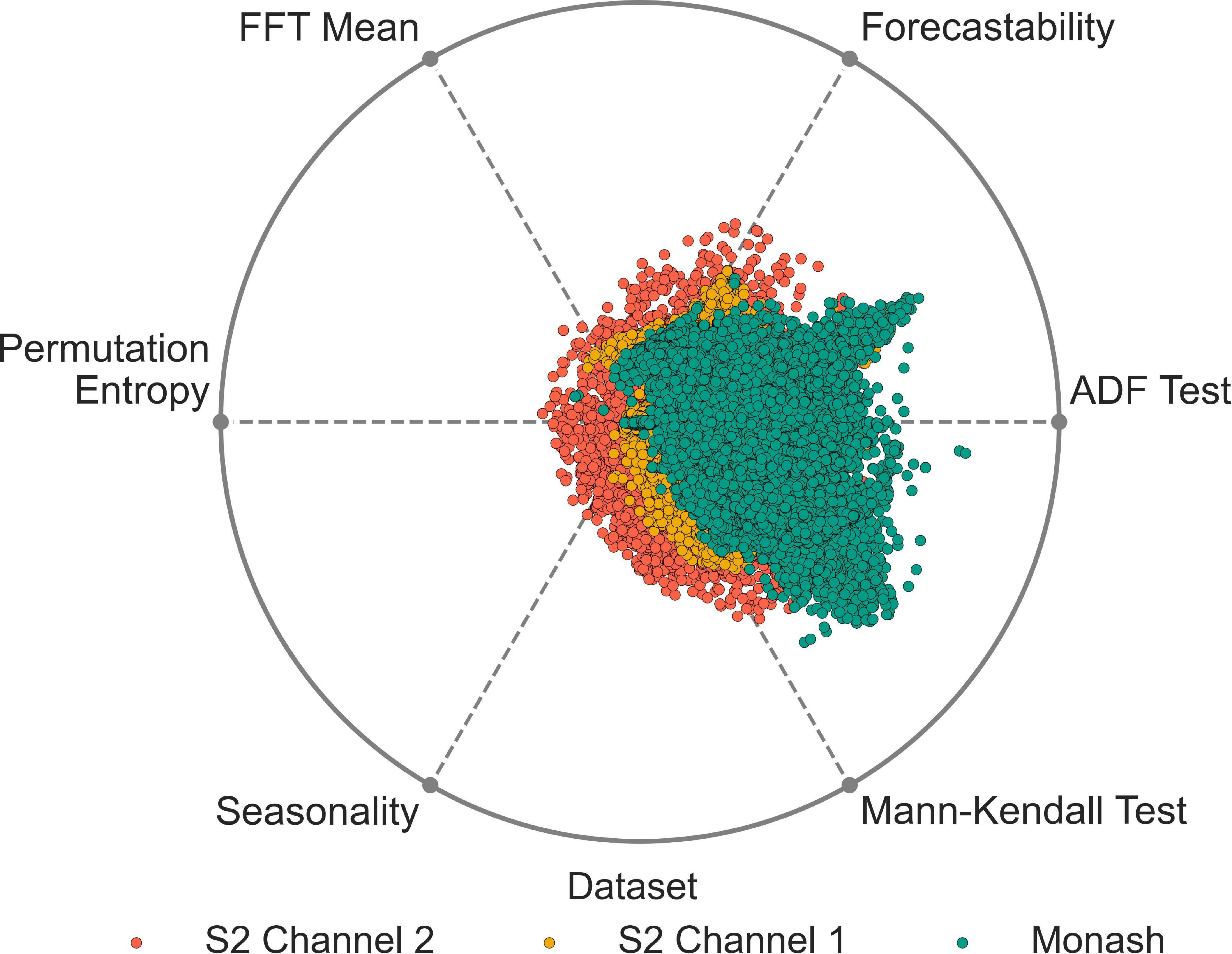}
    \caption{$100K$ samples for single and dual channel}
\end{subfigure}
\caption{The Radviz visualization of S2 and Monash datasets.}
\label{figure:radviz}
\end{figure*}

\textbf{Metric.} To further examine the diversity of the artificially synthesized data in the S2 dataset, we conduct a sampling assessment from six dimensions: stationarity, predictability, frequency domain characteristics, complexity, seasonality intensity, and trend characteristics. For each dimension, we select corresponding statistical indicators for dataset evaluation and quantification, as detailed below: \textbf{(1) Augmented Dickey-Fuller (ADF) Test:} Consistent with section \ref{sec:statistics analysis}, we employ the ADF test to assess the stationarity of time series, using its test statistic as an indicator of time series stationarity \cite{ADF, Timer}. \textbf{(2) Forecastability:} Based on \cite{forecastable} method, we determine whether a time series is chaotic or can be accurately predicted through machine learning models by using Fourier decomposition and entropy \cite{Timer}. Note that since the method provided by \cite{forecastable} is only applicable to multivariate time series, we invert the sampled single-channel time series to form a dual-channel series to calculate the indicator. \textbf{(3) FFT Mean:} We utilize the average of the Fourier transform power spectrum to evaluate the frequency domain characteristics of time series. This indicator can be used to measure the overall intensity of time series and assess the energy distribution. \textbf{(4) Permutation Entropy:} This indicator assesses the dynamic complexity of a time series by analyzing its permutation patterns \cite{permutation}. We set the embedding dimension $m=3$ and time delay $\tau = 1$, and calculate its specific value using Shannon Entropy in Equation \ref{eq:entropy}. See \cite{permutation} for more detailed calculation. \textbf{(5) Seasonality:} We decompose the time series into trend, seasonal and residual components using the Seasonal-Trend Decomposition using LOESS (STL) algorithm \cite{seasonality}. Then, we calculate the intensity of the seasonal component in the time series according to Equation \ref{eq:seasonality}. \textbf{(6) Mann-Kendall Test:} This is a non-parametric statistical method used to detect monotonic trends in time series \cite{MK-test}. The basic principle is to compare the size relationship between each data point and other data points in the time series. Therefore, this method does not rely on a specific distribution of data and is not affected by outliers. We use the statistical test results of this method as the evaluation indicator, where -1 indicates a downward trend, 1 indicates an upward trend, and 0 indicates no obvious trend.
\begin{equation}
    \mathrm{Permutation} = -\sum_{j = 1} ^ {K} P_j \times \mathrm{ln}P_j,
\label{eq:entropy}
\end{equation}
\begin{equation}
\left\{\begin{matrix}
Y_t = T_t + S_t + R_t
 \\
\mathrm{Seasonality} = \mathrm{max} \left \{ 0, 1 - \frac{\mathrm{Var}(R_t)}{\mathrm{Var}(S_t + R_t) } \right \}
\end{matrix}\right. ,
\label{eq:seasonality}
\end{equation}
where, $P_i$ represents the frequency of the $i$-th permutation model in the permutation entropy, and $K=m!$ is the total number of permutation patterns \cite{permutation}. $Y_t$ represents the original time series, $T_t$, $S_t$ and $R_t$ are the trend, seasonal and residual components decomposed by the STL algorithm \cite{seasonality} respectively. $\mathrm{Var}(\cdot)$ means calculating the variance of a series.

\textbf{Setup.} We use the six evaluation metric mentioned above as 6 dimensions \cite{Timer, seasonality, forecastable}, and employ Radviz plots for high-dimensional space visualization \cite{radviz}. Each time series segment is considered as a sample to calculate its statistical indicators, which are projected as coordinates in the high-dimensional space onto the Radviz plot. We compare the statistical characterization of the S2 dataset and the Monash real-world time series dataset to analyze the quality of our synthetic dataset \cite{Monash}. Monash covers time series representing various domains such as weather \cite{weather}, traffic \cite{traffic}, electricity \cite{ECL}, tourism \cite{MS-TIP}, medicine \cite{ECG-LLM}, and energy \cite{Informer}. We uniformly sample $200K$ samples with a length of 256 from each sub-dataset of Monash and calculate their statistical metric. Since the S2 dataset consists of series-symbol sample pairs with different input and output channels, we randomly sample $100K$ samples with a length of 256 from the single-input-channel samples of the S2 dataset and calculate their statistical representations. The high-dimensional visualization results in the Radviz plot are shown in Figure \ref{figure:radviz} (a). Subsequently, we randomly sample $200K$ samples from the single-input-channel samples in the S2 dataset, and the Radviz visualization result is shown in Figure \ref{figure:radviz} (b). Finally, we randomly sample 100K samples from both the single-input-channel and dual-input-channel samples in the S2 dataset, and the Radviz visualization result is shown in Figure \ref{figure:radviz} (c).

\textbf{Results.} The results in Figure \ref{figure:radviz} (a) show that our synthetic S2 dataset and the real-world time series dataset have a large overlap in statistical characterization \cite{SNIP, Timer, DL4Symbolic, Symbolic}. Therefore, S2 dataset is very close to the real dataset in terms of stationarity, predictability, frequency domain characteristics, complexity, seasonality intensity, and trend characteristics. Pre-training models with the S2 dataset allows learning the basic representations of time series \cite{Moment, MOIRAI}. Moreover, since our S2 dataset generation method can provide high-quality series-symbol data unrestrictedly, it essentially covers the entire representation space of time series. As shown in Figure \ref{figure:radviz} (b), when the sampling data increases from $100K$ to $200K$, the coverage of representation also expands, and some data representations even surpass the Monash dataset \cite{Monash}. However, all time series samples in the single-input channel are obtained through sampling from only one input series, i.e., all generated data $y$ is obtained through an expression with only one input variable $f(x_1)$, thus limiting its diversity to some extent. To address this, in Figure \ref{figure:radviz} (c), we sample $100K$ samples each from both single-channel and dual-channel time series and the results show that the diversity of dual-channel time series samples is significantly higher than that of single-input channel samples. By increasing the number of input channels or input variables, we can sample more diverse and complex symbolic expressions $f(x_1, x_2, \cdots, x_n)$, thereby obtaining more diverse time series \cite{SNIP, Symbolic}. This experiment demonstrates that our S2 dataset and the Monash real-world time series dataset have a large overlapping area in statistical representation, indicating the high quality of our synthetic dataset \cite{ForecastPFN, TimePFN}. Considering that our method can generate time series data without limitation, and the coverage of time series statistical characterization expands continuously with the increase of data volume, our synthetic dataset can uniformly cover the basic representations of all types of time series \cite{TimesNet, GPT4TS, graph-forecasting, ALERT, Discovering}.

\section{Model Pre-training and Representation Learning}
\label{sec: model pre-training and representation learning}

\begin{figure*}[!t]
\centering
\begin{subfigure}{0.49\textwidth}
    \includegraphics[width=\linewidth]{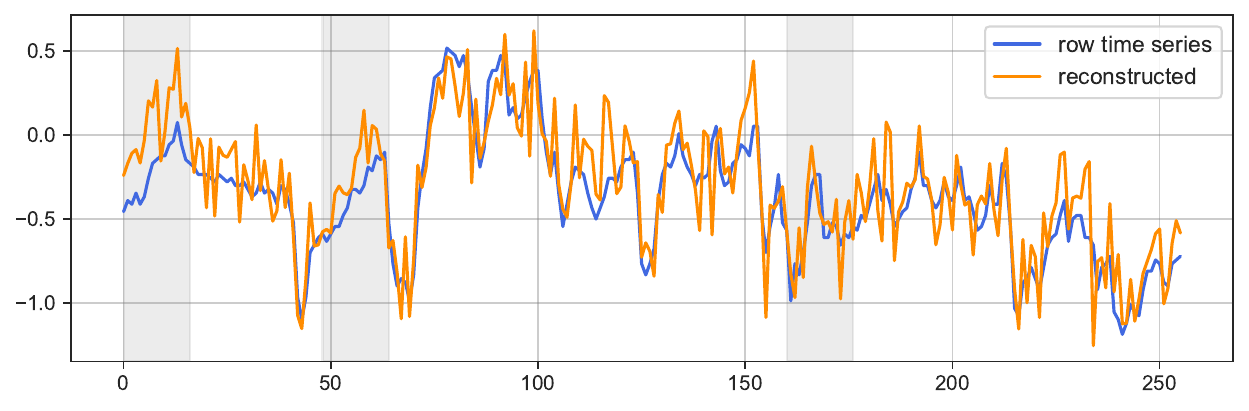}
\end{subfigure}
\hfill
\begin{subfigure}{0.49\textwidth}
    \includegraphics[width=\linewidth]{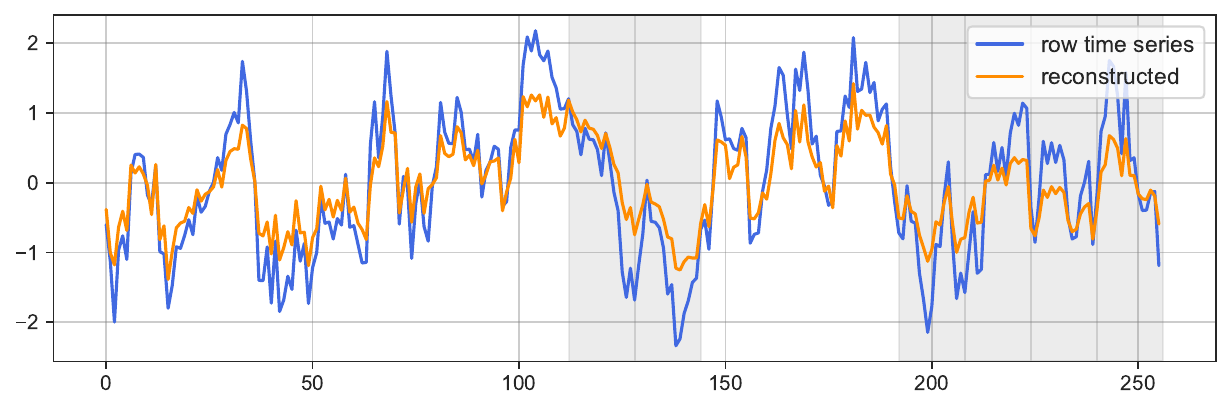}
\end{subfigure}
\medskip
\begin{subfigure}{0.49\textwidth}
    \includegraphics[width=\linewidth]{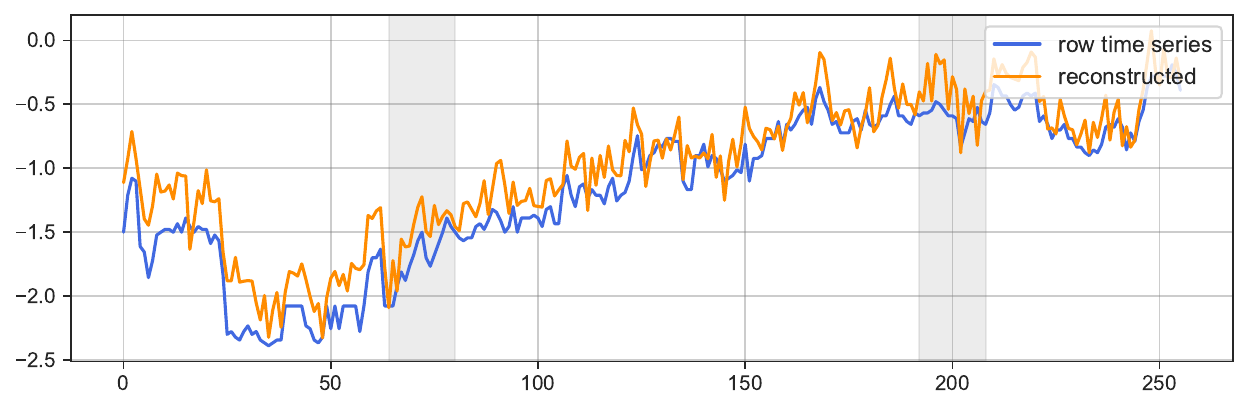}
\end{subfigure}
\hfill
\begin{subfigure}{0.49\textwidth}
    \includegraphics[width=\linewidth]{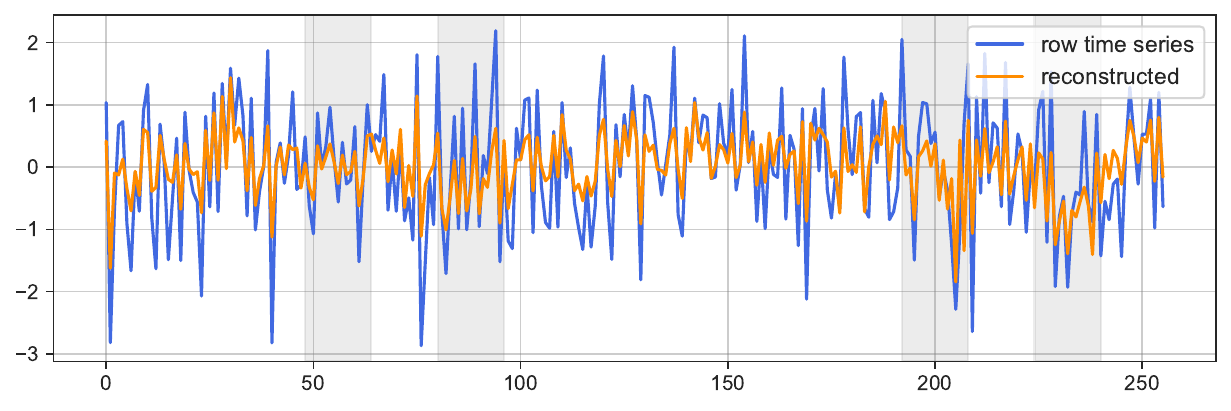}
\end{subfigure}
\medskip
% \begin{subfigure}{0.49\textwidth}
%     \includegraphics[width=\linewidth]{images/data/7.pdf}
% \end{subfigure}
% \hfill
% \begin{subfigure}{0.49\textwidth}
%     \includegraphics[width=\linewidth]{images/data/8.pdf}
% \end{subfigure}
% \medskip
% \begin{subfigure}{0.49\textwidth}
%     \includegraphics[width=\linewidth]{images/data/9.pdf}
% \end{subfigure}
% \hfill
% \begin{subfigure}{0.49\textwidth}
%     \includegraphics[width=\linewidth]{images/data/10.pdf}
% \end{subfigure}
% \medskip
\begin{subfigure}{0.49\textwidth}
    \includegraphics[width=\linewidth]{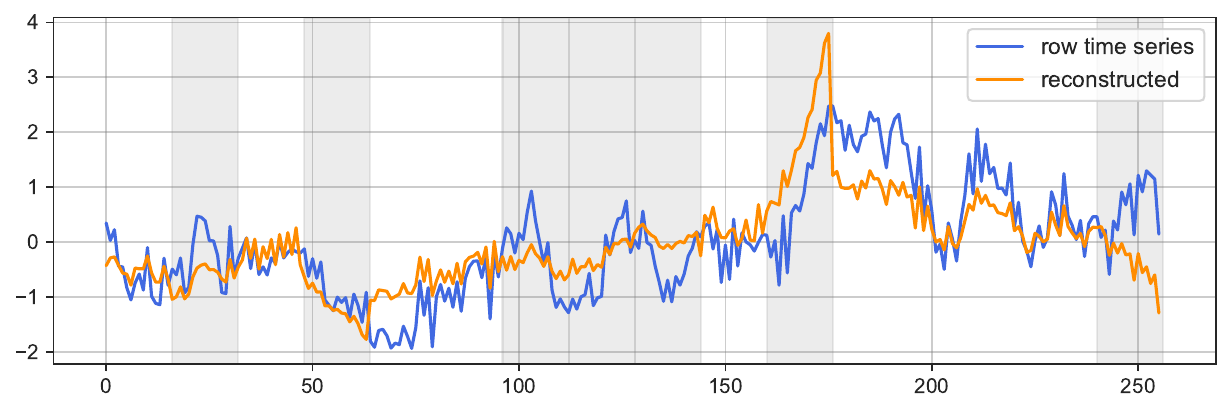}
\end{subfigure}
\hfill
\begin{subfigure}{0.49\textwidth}
    \includegraphics[width=\linewidth]{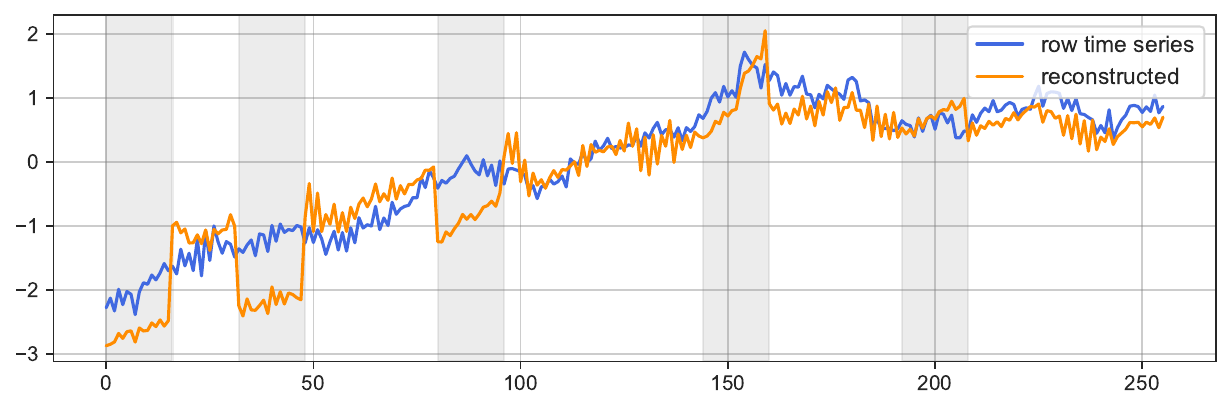}
\end{subfigure}
\medskip
\caption{Zero-shot time series imputation in S2 out-of-domain data.}
\label{figure: zero-shot in out-of-domain data}
\end{figure*}

\begin{figure*}[!t]
\centering
\begin{subfigure}{0.49\textwidth}
    \includegraphics[width=\linewidth]{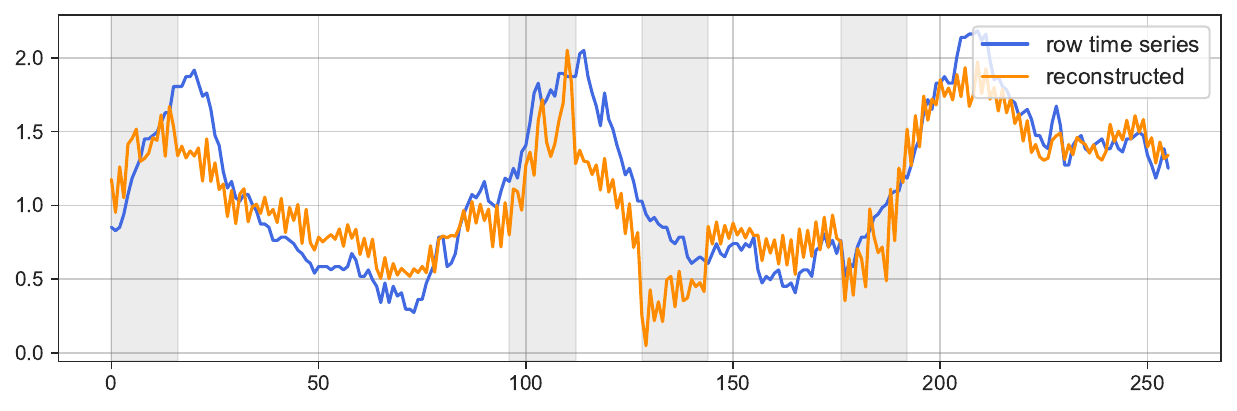}
    \caption{ETTm1}
\end{subfigure}
\hfill
\begin{subfigure}{0.49\textwidth}
    \includegraphics[width=\linewidth]{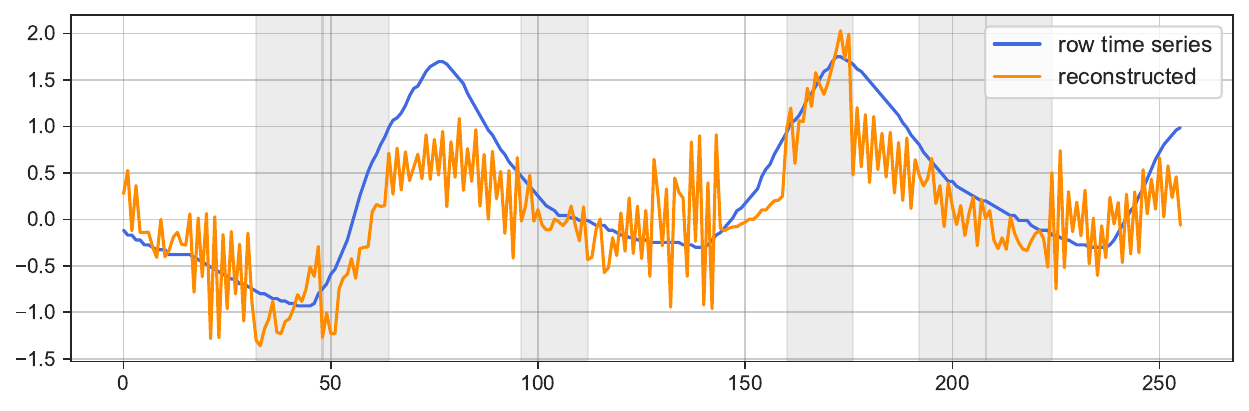}
    \caption{ETTm2}
\end{subfigure}
\medskip
\begin{subfigure}{0.49\textwidth}
    \includegraphics[width=\linewidth]{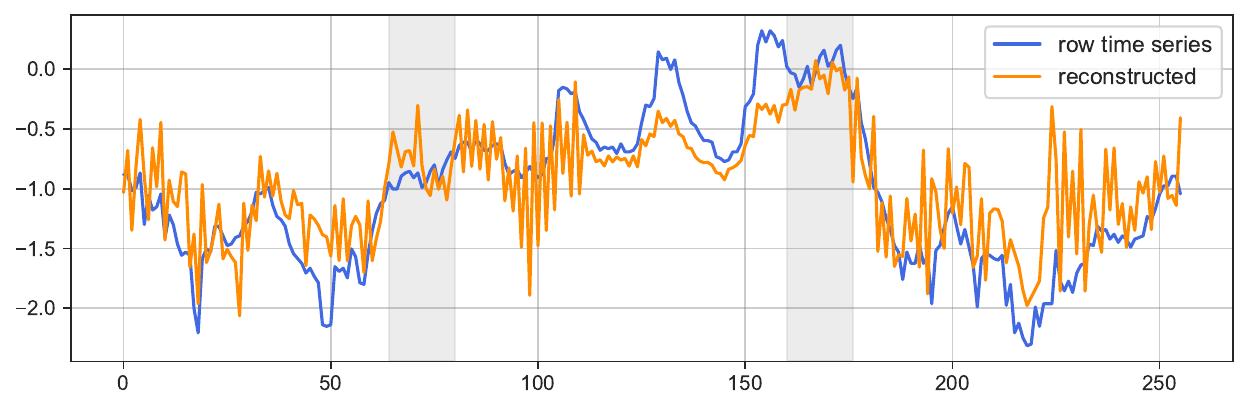}
    \caption{ETTh1}
\end{subfigure}
\hfill
\begin{subfigure}{0.49\textwidth}
    \includegraphics[width=\linewidth]{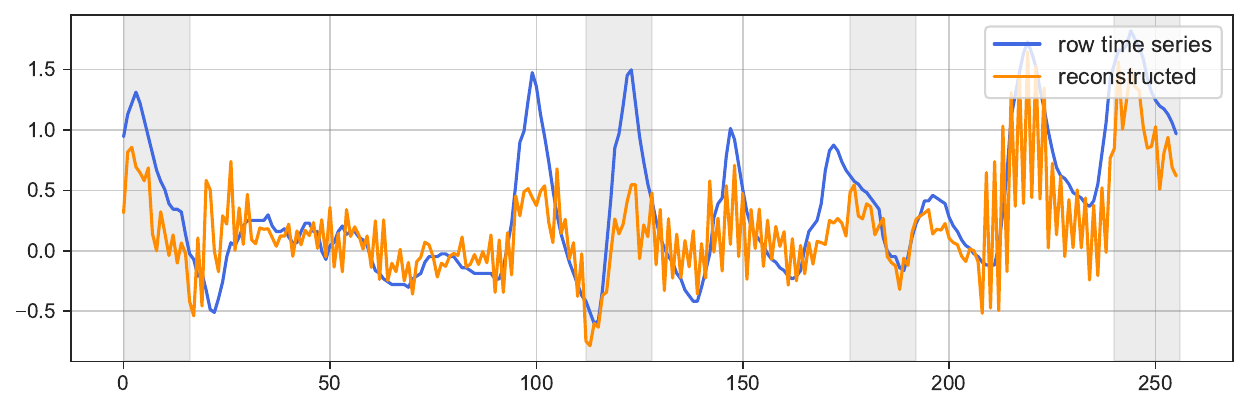}
    \caption{ETTh2}
\end{subfigure}
\medskip
\begin{subfigure}{0.49\textwidth}
    \includegraphics[width=\linewidth]{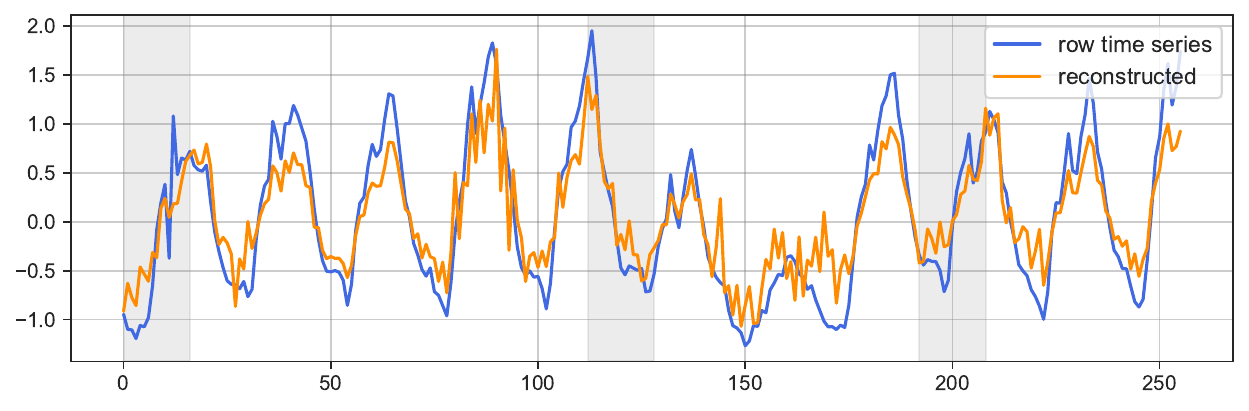}
    \caption{Electricity}
\end{subfigure}
\hfill
\begin{subfigure}{0.49\textwidth}
    \includegraphics[width=\linewidth]{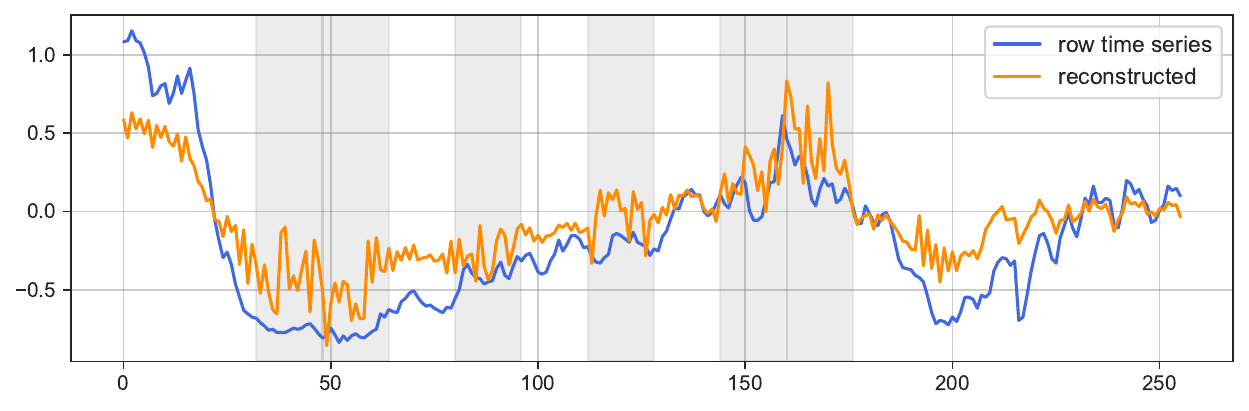}
    \caption{Weather}
\end{subfigure}
\medskip
\caption{Zero-shot time series imputation in real world time series dataset in ETTm1, ETTm2, ETTh1, ETTh2 \cite{Informer}, Electricity \cite{ECL} and Weather \cite{weather}.}
\label{figure: zero-shot in real world data}
\end{figure*}

In this section, we analyze and present the model's pre-training status and the learned representations \cite{CLIP, BEiT}. We first examine the masked time series modeling (MTM) \cite{PatchTST, SimMTM} of the time series encoder in \texttt{SymTime} and test its zero-shot imputation capability on both newly generated data and real time series data. Then, we analyze and visualize the cross-modal representations learned by the time series encoder and the symbol encoder through series-to-symbol and symbol-to-series contrastive loss \cite{ALBEF, MoCo_v1}. By comparing the representation learning effects before and after model pre-training \cite{graph-forecasting}, we demonstrate that the time series encoder in \texttt{SymTime} can not only learn the basic representations of time series data through masked time series modeling but also acquire semantic information of symbolic expressions through contrastive losses.

\subsection{Masked Time Series Modeling and Zero-shot Imputation}

\textbf{Setup.} Since we incorporate MTM loss in the pre-training process of \texttt{SymTime}, in this section, we assess the specific learning effects of the time series encoder in \texttt{SymTime} through masked modeling \cite{PatchTST, SimMTM, TimeSiam, TimeMixer}. We test the model's performance using both pre-trained synthetic data not in the S2 dataset and real datasets from time series imputation tasks \cite{SNIP, Symbolic, DL4Symbolic}. As \texttt{SymTime} adds masks in units of patches of length 16 during pre-training, we also add masks in the form of 16-length patches. The reconstruction effect of the masked parts by the time series encoder is shown in Figure \ref{figure: zero-shot in out-of-domain data} and \ref{figure: zero-shot in real world data}.

\textbf{S2 Dataset Out-of-Domain Data.} In Figure \ref{figure: zero-shot in out-of-domain data}, we generate new data using the method from the S2 dataset and add masks to test the reconstruction ability of the time series encoder \cite{SNIP, Symbolic, DL4Symbolic}. The gray sections represent the masked segments, while blue and orange represent the original and reconstructed series, respectively. We input time series outside the gray parts in patches and have the model reconstruct the gray sections based on the remaining information. Since we only calculate the MTM loss on the masked parts \cite{SimMTM, TimeSiam}, the visible reconstruction does not overlap with the original input series \cite{Timer, TimeFM, Moment}. From the Figure \ref{figure: zero-shot in out-of-domain data}, it can be observed that the time series encoder in \texttt{SymTime} performs well in fitting the fluctuations and trends of time series, demonstrating that our encoder successfully learned the fundamental representations of time series during pre-training \cite{MOIRAI, SAMformer, Discovering}.

\textbf{Real-world Time Series Data.} In Figure \ref{figure: zero-shot in real world data}, we conduct representation learning tests on 6 real datasets: ETTm1, ETTm2, ETTh1, ETTh2 \cite{Informer}, Electricity \cite{ECL}, and Weather \cite{weather}. Since no real data are used for model pre-training, these datasets are also considered as out of domain data. We similarly add masks in patch units (gray sections). It can be observed that the time series encoder in \texttt{SymTime} also performs well in zero-shot reconstruction on real-world data \cite{SimMTM, MaskTime}.

\subsection{Series-Symbol Representation Learning of Time Series Encoder}

\begin{figure*}[!t]
\centering
\begin{subfigure}{0.49\textwidth}
    \includegraphics[width=\linewidth]{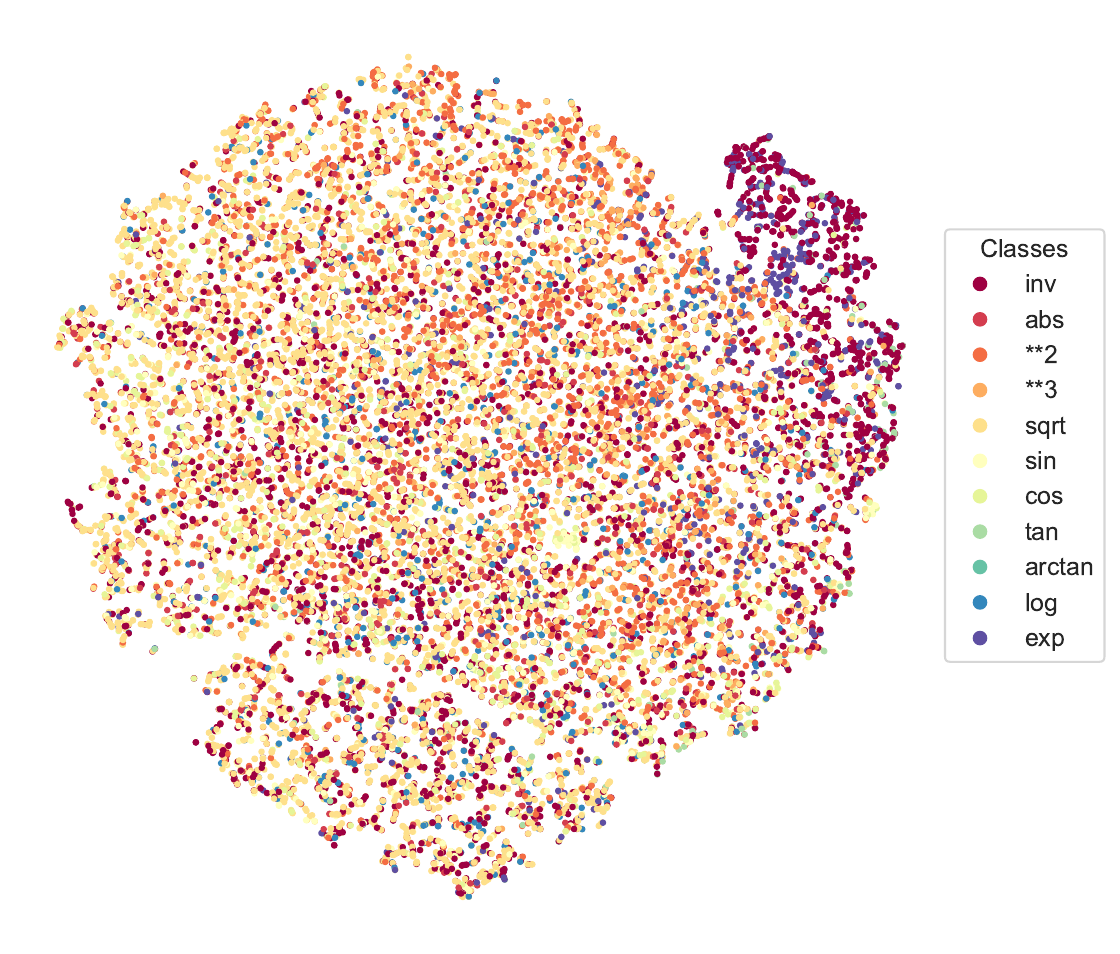}
    \caption{without pre-training}
\end{subfigure}
\hfill
\begin{subfigure}{0.49\textwidth}
    \includegraphics[width=\linewidth]{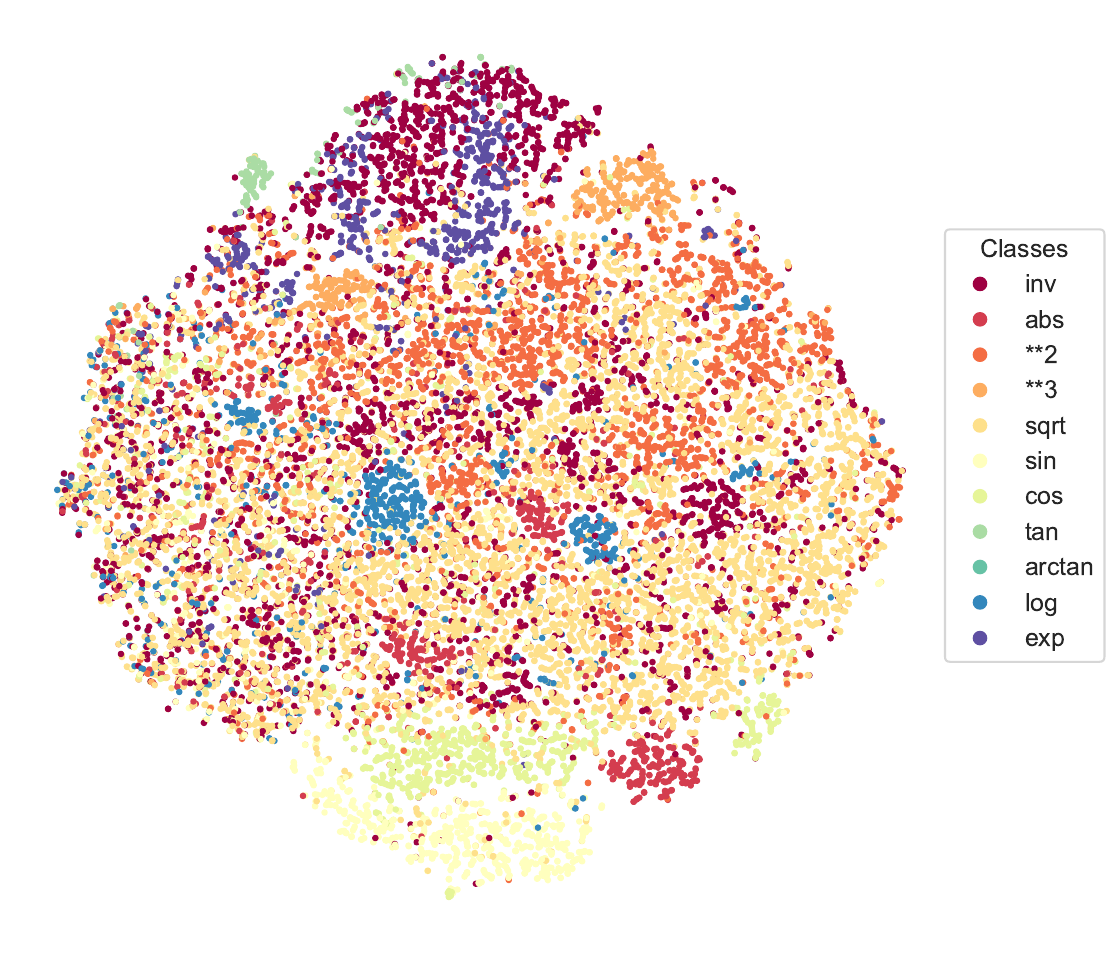}
    \caption{with pre-training}
\end{subfigure}
\medskip
\caption{The t-SNE visualization of time series encoder in \texttt{SymTime} representation space.}
\label{figure: representation of time series encoder}
\end{figure*}

\textbf{Setup.} To evaluate the cross-modal representations \cite{ALBEF} and semantic information \cite{CLIP, S2IP-LLM, Time-LLM} by the time series encoder in \texttt{SymTime}, we select $20\mathrm K$ single-input single-output channel symbolic expressions containing only one type of unary operator and their corresponding sampled series from the S2 dataset \cite{SNIP}. These symbolic expressions include only one specific math operator from the set $\{ \mathrm{inv, abs, pow2, pow3, sqrt, sin, cos, tan, arctan, log, exp} \}$, using this operator as the category label for the entire series data \cite{Symbolic, DL4Symbolic}. Then, we do patching on the sampled time series and input them into the \texttt{SymTime} time series encoder, both pre-trained and without pre-trained \cite{PatchTST}. Finally, we use the series representations reconstructed by the time series encoder and apply t-SNE \cite{t-SNE} to reduce them to a two-dimensional space to observe the distribution of representations for series generated by different types of symbols. The specific results are shown in Figure \ref{figure: representation of time series encoder}.

\textbf{Results.} From Figure \ref{figure: representation of time series encoder} (a), it can be observed that the time series encoder without pre-training cannot effectively distinguish different series-symbol categories. Only a few symbols, such as inv and exp, which may have more distinct data representations, form certain clusters even without pre-training. Due to we introduce both series-to-symbol and symbol-to-series contrastive losses in the pre-training of \texttt{SymTime}, contrastive learning can make positive samples as close as possible in the representation space and negative samples as distant as possible \cite{CLIP, COMET, MoCo_v1, SimCLR}. Through this form of learning, as shown in Figure \ref{figure: representation of time series encoder} (b), the pre-trained time series encoder can clearly form simple clusters for series of the same type of unary symbol operator. Similar symbols like sin and cos are relatively close in the two-dimensional representation space after t-SNE dimensionality reduction \cite{t-SNE}. The three polynomial operations, pow2, pow3, and sqrt, which are the most numerous, also show a certain pattern in the representation space. The log operation is surrounded by polynomial operations and also forms a distinct cluster. This indicates that \texttt{SymTime} has successfully brought samples of the same category closer together in the latent space through contrastive learning, and thus our time series encoder has learned the semantic information of symbols \cite{SNIP}. Finally, ablation experiments reveal that learning with both contrastive losses simultaneously enhances the performance of downstream time series tasks.

\begin{figure*}[!t]
\centering
\begin{subfigure}{0.49\textwidth}
    \includegraphics[width=\linewidth]{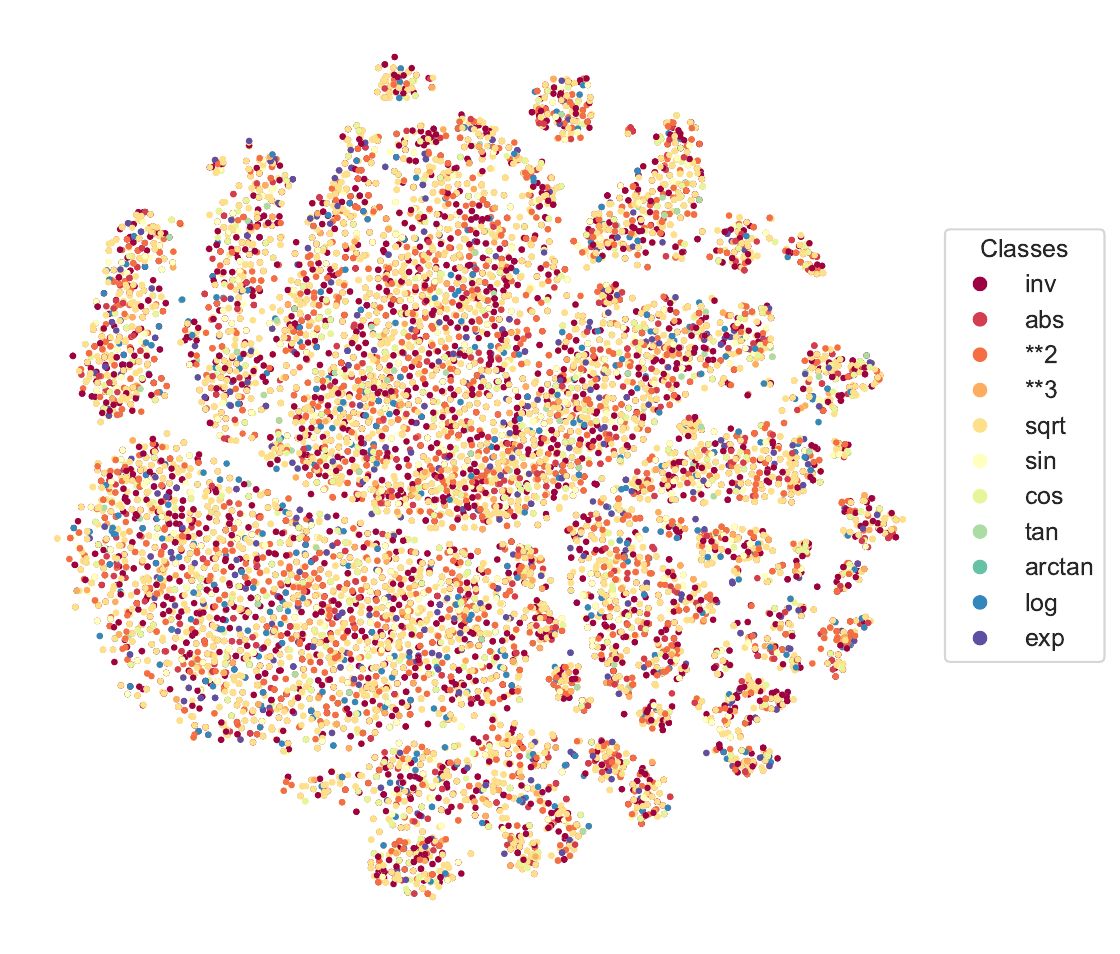}
    \caption{without pre-training}
\end{subfigure}
\hfill
\begin{subfigure}{0.49\textwidth}
    \includegraphics[width=\linewidth]{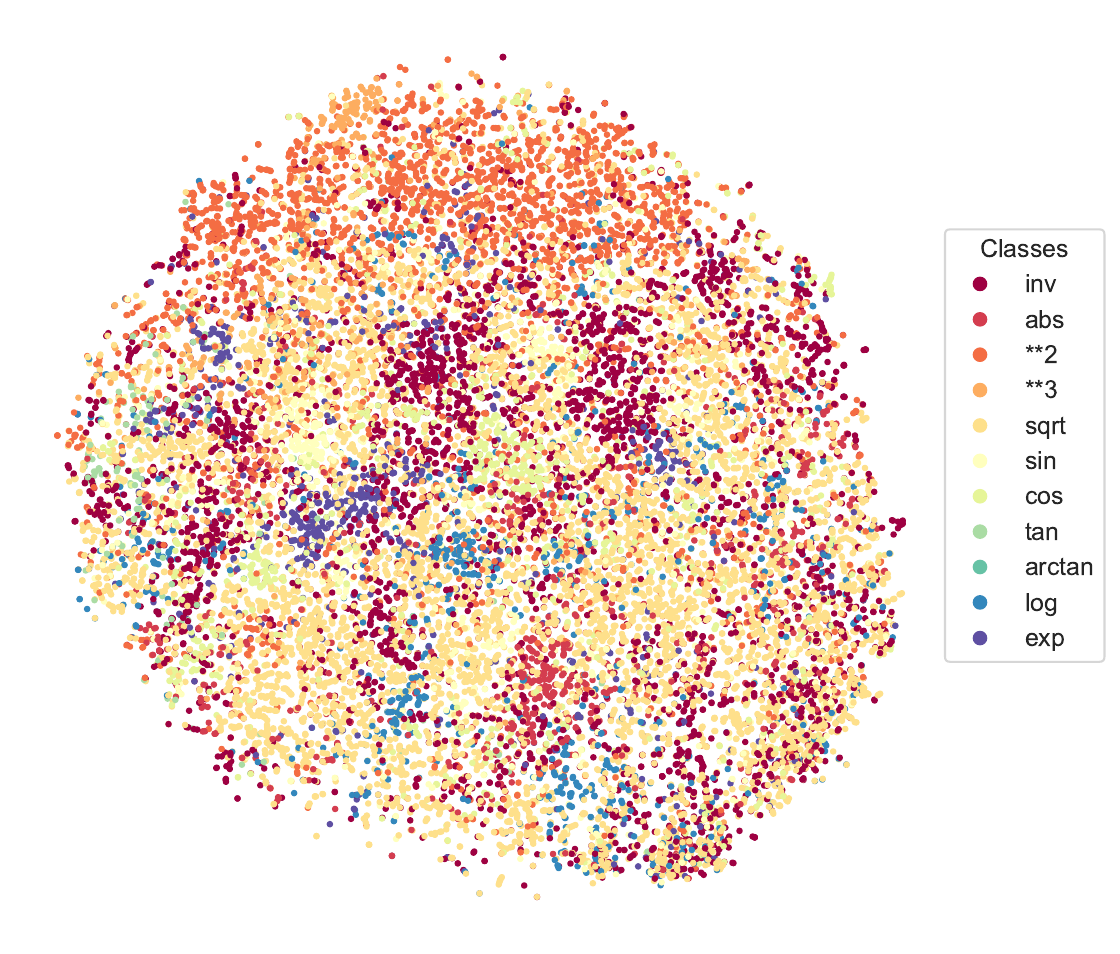}
    \caption{with pre-training}
\end{subfigure}
\medskip
\caption{The t-SNE visualization of symbol encoder representation space.}
\label{figure: representation of symbol encoder}
\end{figure*}

\subsection{Series-Symbol Representation Learning of Symbol Encoder}

\textbf{Setup.} To capture the features of symbolic expressions and provide representation data for the contrastive learning of the time series encoder, we introduce a pre-trained distilled BERT as the symbol encoder \cite{BERT, DistilBERT}. This approach pre-trains the time series encoder and equips it with semantic information from symbolic expressions. To examine the cross-modal representation learned by the symbol encoder, we employ the same method as above, using $20 \mathrm K$ symbolic expressions with a single mathematical operator for our experiment. We first tokenize this series of symbolic expressions and then input them into both the pre-trained and non-pre-trained symbol encoders to obtain the \texttt{[CLS]} token from the last layer output. Finally, we use t-SNE to reduce the obtained \texttt{[CLS]} tokens and project them onto a two-dimensional representation space. The results are shown in Figure \ref{figure: representation of symbol encoder}.

\textbf{Results.} Since the generated symbolic expressions contain specific unary operators, random constants, and binary operators such as $\{ +, -, \times \}$ \cite{SNIP}, it can be observed from the symbol encoder without pre-training in Figure \ref{figure: representation of symbol encoder} (a) that the embeddings of various symbols exhibit certain distinctive shapes but are not sensitive to the influence of specific unary operators. All types of symbols are chaotically mixed together. However, after masked language modeling (MLM) generative pre-training and contrastive learning \cite{GPT-2, BERT, BEiT}, the symbol encoder, composed of a LLM, becomes more sensitive to specific symbolic data, and similar types of symbolic expressions converge to form distinct cluster structures in Figure \ref{figure: representation of symbol encoder} (b).

\section{Implementation Details}
\label{sec:implementation details}

In this section, we first provide a detailed introduction to the datasets and evaluation metrics used for the five TSA tasks. Subsequently, we elaborate on the training details of our experiments, including how we pre-trained \texttt{SymTime} on the S2 dataset and how we fine-tuned it on downstream task datasets. All experiments and deep neural networks training are implemented in PyTorch on NVIDIA A6000 48GB GPU.

\subsection{Downstream Tasks Datasets Details}
\label{sec:Appendix downstream tasks}

We conduct experiments using the TimesNet benchmark \cite{TimesNet}, with a detailed description of the dataset provided in Table \ref{table:dataset}. Specifically, we utilize 8 datasets including ETTh1, ETTh2, ETTm1, ETTm2 \cite{Informer}, Electricity \cite{ECL}, Traffic \cite{traffic}, Weather \cite{weather}, and Exchange \cite{LSTNet} to conduct long-term time series forecasting experiments. Our model, \texttt{SymTime}, employ input series of lookback lengths 96 and 512, with forecast horizons of 96, 192, 336, and 720. For short-term forecasting experiments, we employ the M4 benchmark dataset, predicting data of various frequencies \cite{M4team2018dataset}. In the time series imputation task, we test on 6 datasets—ETTh1, ETTh2, ETTm1, ETTm2 \cite{Informer}, Electricity \cite{ECL}, and Weather \cite{weather}—with mask rates of 12.5\%, 25\%, 37.5\%, and 50\%. For time series classification, we utilize ten UEA multivariate time series classification benchmark datasets \cite{UEA}. For anomaly detection in time series, we experiment with five datasets: SMD \cite{SMD}, MSL \cite{MSL}, SMAP \cite{MSL}, SWaT \cite{SWaT}, and PSM \cite{PSM}.

\begin{table}[!t]
  \caption{Dataset descriptions. The dataset size is organized in (Train, Validation, Test).}
  \label{table:dataset}
  \vskip 0.10in
  \centering
  \begin{threeparttable}
  \begin{small}
  \renewcommand{\multirowsetup}{\centering}
  \setlength{\extrarowheight}{2.5pt}
  \setlength{\tabcolsep}{5pt}
  \begin{tabular}{c|l|c|c|c|c}
    \toprule
    Tasks & Dataset & Dim & Series Length & Dataset Size & \scalebox{0.8}{Information (Frequency)} \\
    \toprule
     & ETTm1, ETTm2 & 7 & \scalebox{0.8}{\{96, 192, 336, 720\}} & (34465, 11521, 11521) & \scalebox{0.8}{Electricity (15 mins)}\\
    \cmidrule{2-6}
     & ETTh1, ETTh2 & 7 & \scalebox{0.8}{\{96, 192, 336, 720\}} & (8545, 2881, 2881) & \scalebox{0.8}{Electricity (15 mins)} \\
    \cmidrule{2-6}
    Forecasting & Electricity & 321 & \scalebox{0.8}{\{96, 192, 336, 720\}} & (18317, 2633, 5261) & \scalebox{0.8}{Electricity (Hourly)} \\
    \cmidrule{2-6}
    (Long-term) & Traffic & 862 & \scalebox{0.8}{\{96, 192, 336, 720\}} & (12185, 1757, 3509) & \scalebox{0.8}{Transportation (Hourly)} \\
    \cmidrule{2-6}
     & Weather & 21 & \scalebox{0.8}{\{96, 192, 336, 720\}} & (36792, 5271, 10540) & \scalebox{0.8}{Weather (10 mins)} \\
    \cmidrule{2-6}
     & Exchange & 8 & \scalebox{0.8}{\{96, 192, 336, 720\}} & (5120, 665, 1422) & \scalebox{0.8}{Exchange rate (Daily)}\\
    \midrule
     & M4-Yearly & 1 & 6 & (23000, 0, 23000) & \scalebox{0.8}{Demographic} \\
    \cmidrule{2-5}
     & M4-Quarterly & 1 & 8 & (24000, 0, 24000) & \scalebox{0.8}{Finance} \\
    \cmidrule{2-5}
    Forecasting & M4-Monthly & 1 & 18 & (48000, 0, 48000) & \scalebox{0.8}{Industry} \\
    \cmidrule{2-5}
    (short-term) & M4-Weakly & 1 & 13 & (359, 0, 359) & \scalebox{0.8}{Macro} \\
    \cmidrule{2-5}
     & M4-Daily & 1 & 14 & (4227, 0, 4227) & \scalebox{0.8}{Micro} \\
    \cmidrule{2-5}
     & M4-Hourly & 1 &48 & (414, 0, 414) & \scalebox{0.8}{Other} \\
    \midrule
    \multirow{5}{*}{Imputation} & ETTm1, ETTm2 & 7 & 96 & (34465, 11521, 11521) & \scalebox{0.8}{Electricity (15 mins)} \\
    \cmidrule{2-6}
     & ETTh1, ETTh2 & 7 & 96 & (8545, 2881, 2881) & \scalebox{0.8}{Electricity (15 mins)}\\
    \cmidrule{2-6}
     & Electricity & 321 & 96 & (18317, 2633, 5261) & \scalebox{0.8}{Electricity (15 mins)}\\
    \cmidrule{2-6}
    & Weather & 21 & 96 & (36792, 5271, 10540) & \scalebox{0.8}{Weather (10 mins)} \\
    \midrule
     & \scalebox{0.8}{EthanolConcentration} & 3 & 1751 & (261, 0, 263) & \scalebox{0.8}{Alcohol Industry}\\
    \cmidrule{2-6}
    & \scalebox{0.8}{FaceDetection} & 144 & 62 & (5890, 0, 3524) & \scalebox{0.8}{Face (250Hz)}\\
    \cmidrule{2-6}
    & \scalebox{0.8}{Handwriting} & 3 & 152 & (150, 0, 850) & \scalebox{0.8}{Handwriting}\\
    \cmidrule{2-6}
    & \scalebox{0.8}{Heartbeat} & 61 & 405 & (204, 0, 205)& \scalebox{0.8}{Heart Beat}\\
    \cmidrule{2-6}
    Classification & \scalebox{0.8}{JapaneseVowels} & 12 & 29 & (270, 0, 370) & \scalebox{0.8}{Voice}\\
    \cmidrule{2-6}
    (UEA) & \scalebox{0.8}{PEMS-SF} & 963 & 144 & (267, 0, 173) & \scalebox{0.8}{Transportation (Daily)}\\
    \cmidrule{2-6}
    & \scalebox{0.8}{SelfRegulationSCP1} & 6 & 896 & (268, 0, 293) & \scalebox{0.8}{Health (256Hz)}\\
    \cmidrule{2-6}
    & \scalebox{0.8}{SelfRegulationSCP2} & 7 & 1152 & (200, 0, 180) & \scalebox{0.8}{Health (256Hz)}\\
    \cmidrule{2-6}
    & \scalebox{0.8}{SpokenArabicDigits} & 13 & 93 & (6599, 0, 2199) & \scalebox{0.8}{Voice (11025Hz)}\\
    \cmidrule{2-6}
    & \scalebox{0.8}{UWaveGestureLibrary} & 3 & 315 & (120, 0, 320) & \scalebox{0.8}{Gesture}\\
    \midrule
     & SMD & 38 & 100 & (566724, 141681, 708420) & \scalebox{0.8}{Server Machine} \\
    \cmidrule{2-6}
    Anomaly & MSL & 55 & 100 & (44653, 11664, 73729) & \scalebox{0.8}{Spacecraft} \\
    \cmidrule{2-6}
    Detection & SMAP & 25 & 100 & (108146, 27037, 427617) & \scalebox{0.8}{Spacecraft} \\
    \cmidrule{2-6}
     & SWaT & 51 & 100 & (396000, 99000, 449919) & \scalebox{0.8}{Infrastructure} \\
    \cmidrule{2-6}
    & PSM & 25 & 100 & (105984, 26497, 87841)& \scalebox{0.8}{Server Machine} \\
    \bottomrule
    \end{tabular}
    \end{small}
  \end{threeparttable}
  \vspace{-5pt}
\end{table}

\subsection{Metrics}
\label{sec:Appendix Metrics}

We assess the five TSA tasks using various metrics. For long-term forecasting and imputation tasks, we employ mean squared error (MSE) and mean absolute error (MAE). For short-term forecasting, we utilize symmetric mean absolute percentage error (SMAPE), mean absolute scaled Error (MASE), and overall weighted average (OWA), with OWA being a metric unique to the M4 competition. For time series classification tasks, we use classification accuracy as the metric. For anomaly detection tasks, we adopt precision, recall, and F1-score as our evaluation metrics. The calculations for these metrics are as follows.
\begin{equation}
    \mathrm{MSE} = \sum_{i=1}^{n} \left ( y_i - \hat{y}_i  \right ) ^ 2,
\end{equation}
\begin{equation}
    \mathrm{MAE} = \sum_{i=1}^{n} \left | y_i - \hat{y}_i  \right |,
\end{equation}
\begin{equation}
    \mathrm{SMAPE} = \frac{200}{T} \sum_{i = 1} ^ {T} \frac{\left | \mathbf{X}_i - \hat{\mathbf{Y}_i} \right |}{\left |\mathbf{X}_i \right | + \left | \hat{\mathbf{Y}}_i  \right |},
\end{equation}
\begin{equation}
    \mathrm{MAPE} = \frac{100}{T}\sum_{i = 1} ^ {T} \frac{\left | \mathbf{X}_i - \hat{\mathbf{Y}_i} \right |}{\left |\mathbf{X}_i \right |},
\end{equation}
\begin{equation}
    \mathrm{MASE} = \frac{1}{T}\sum_{i = 1} ^ {T} \frac{\left | \mathbf{X}_i - \hat{\mathbf{Y}_i} \right |}{\frac{1}{T-q} \sum_{j = q+1}^{T} \left | \mathbf{X}_j  - \mathbf{X}_{j - q} \right |},    
\end{equation}
\begin{equation}
    \mathrm{OWA} = \frac{1}{2} \left [ \frac{\mathrm{SMAPE} }{\mathrm{SMAPE}_\text{Naïve2}} + \frac{\mathrm{MASE}}{\mathrm{MASE}_\text{Naïve2} } \right ],
\end{equation}
where, $y_i$ is the ground true value, $\hat{y}_i$ is the model prediction, $q$ is the peridoicity of the time series data. $\mathbf{X}, \hat{\mathbf{Y}} \in \mathbb{R} ^ {T \times C}$ are the ground truth and prediction results of the future with $T$ time points and $C$ dimensions. $\mathbf{X}_i$ means the $i$-th future time point.

\subsection{Pre-training}
\label{sec:Appendix Pre-training}

\textbf{Model Hyper-parameter.} The parameter configurations for the time series encoder and symbol encoder in \texttt{SymTime} are shown in Table \ref{table:model architecture}. During model pre-training, we primarily set three hyperparameters: (1) the masking ratio of time series patches, (2) the masking ratio for natural language symbols, and (3) the proportion factor $\alpha$ used to balance pseudo-targets in momentum distillation. Based on the masked time series modeling pre-training experimental configuration of PatchTST \cite{PatchTST} and SimMTM \cite{SimMTM}, we set the masking ratio for time series to 40\%. Following the experimental configuration of BERT in masked language modeling \cite{BERT, DistilBERT}, we set the masking ratio for symbolic data to 15\%. Based on the experimental configuration of momentum distillation in ALBEF \cite{ALBEF, momentum-distillation, MoCo_v1}, we set $\alpha$ to 0.6.

\textbf{Training Configurations.} During the pre-training of \texttt{SymTime}, we employ AdamW \cite{Adam, AdamW} as the optimizer with the defult hyperparameter configuration for ($\beta_1$, $\beta_2$) as (0.9, 0.999). Then, we utilize the OneCycle policy to dynamically adjust the learning rate. We set the warmup epochs to 10, during which the learning rate gradually grows up to an initial value of $5 \times 10^{-5}$, and then adjust it dynamically using a cosine annealing schedule, with the minimum learning rate set at $1 \times 10^{-7}$. We conduct pre-training using data parallelism on a hardware setup consisting of 8 NVIDIA RTX A6000 GPUs with 48GB of memory each. We set the batch size to 128 and trained for a total of 85 epochs. Unlike SNIP \cite{SNIP}, we do not generate data on-the-fly during training for pre-training. Instead, we prepare the data in advance and then load it into the device for pre-training. Due to the large size of our generat S2 dataset, we load data into the GPU in batches during each epoch for pre-training.

\subsection{Fine-tuning}
\label{sec:Appendix Fine-tuning}

For the five major tasks in TSA, we conduct downstream task fine-tuning experiments using the configurations in Table \ref{table: fine-tuning configs}. For all downstream task fine-tuning experiments, we employ the Adam optimizer \cite{Adam, AdamW} with hyperparameters $(\beta_1, \beta_2)$ set to $(0.9, 0.999)$. The LR in the table represents the initial learning rate and we utilize the dynamic learning rate adjustment strategy from TimesNet \cite{TimesNet}.

\begin{table}[ht]
\caption{Experiment configuration of \texttt{SymTime} fine-tuning.}
\centering
\vskip 0.10in
\begin{threeparttable}
\begin{small}
\setlength{\extrarowheight}{2pt}
\setlength{\tabcolsep}{6pt}
\begin{tabular}{c|c|c|c|c|c|c|c}
\toprule
\multirow{2}{*}{\textbf{Tasks / Configurations}} & \multicolumn{3}{c}{\textbf{Model Parameter}}& \multicolumn{4}{c}{\textbf{Training Configurations}}\\
\cmidrule(lr){2-4} \cmidrule(lr){5-8}  
& $d_{\mathrm{model}}$& $d_{\mathrm{ff}}$ & Layers  & LR& Loss& Batch Size & Epochs \\
\midrule
Long-term Forecasting& \multirow{5}{*}{512} & \multirow{5}{*}{2048} & 3, 6 & $10^{-4} - 5 \times 10 ^ {-4}$ & MSE& 4-64 & 20 \\
Short-term Forecasting& & & 2, 3 & $10^{-4} - 2 \times 10 ^ {-4}$ & SMAPE & 8-32 & 16 \\
Classification & & & 1-6 & $10^{-4} - 5 \times 10 ^ {-3}$ & Cross Entropy & 4-64 & 64 \\
Imputation & & & 2, 3, 6 & $10^{-4} - 5 \times 10 ^ {-4}$ & MSE & 4-64 & 32\\
Anomaly Detection& && 3, 6 & $10^{-4} - 5 \times 10 ^ {-4}$ & MSE& 4-64& 12 \\
\bottomrule
\end{tabular}
\end{small}
\end{threeparttable}
\label{table: fine-tuning configs}
\end{table}

\subsection{Ablation Experiments Details}
\label{sec:Appendix Ablation Experiments Details}

\textbf{Ablation study on pre-training strategies and objectives.} To further verify the effectiveness of our series-symbol pre-training strategy and objectives, we establish 8 distinct ablation experiment groups and a control group. The specific configurations of these 8 ablation experiment groups are as follows.
\begin{enumerate}
    \item \textbf{Freeze}: All parameters in the pre-trained time series encoder are frozen, with only the linear projection layer for outputting prediction results fine-tuned.
    \item \textbf{w/o Pretrain}: No series-symbol pre-training is conducted; the time series encoder with initialized parameters is used for downstream task experiments.
    \item \textbf{w/o MTM}: The masked time series modeling (MTM) is removed from the pre-training objectives.
    \item \textbf{w/o MLM}: The masked language modeling (MLM) is removed from the pre-training objectives.
    \item \textbf{w/o T2S}: The contrastive loss from time series to symbols is removed from the pre-training objectives.
    \item \textbf{w/o S2T}: The contrastive loss from symbols to time series is removed from the pre-training objectives.
    \item \textbf{w/o Symbol}: Only time series data from the S2 dataset are used to pre-train the time series encoder via MTM, disregarding the correspondence with symbols.
    \item \textbf{w/o Distill}: The contrastive loss in pre-training does not use the pseudo objective of momentum distillation.
\end{enumerate}

\subsection{Ablation Experiments on Short-term Forecasting}
\label{sec:ablation on short-term forecasting}

\begin{figure*}
\begin{subfigure}{0.6\textwidth}
    \includegraphics[width=\linewidth]{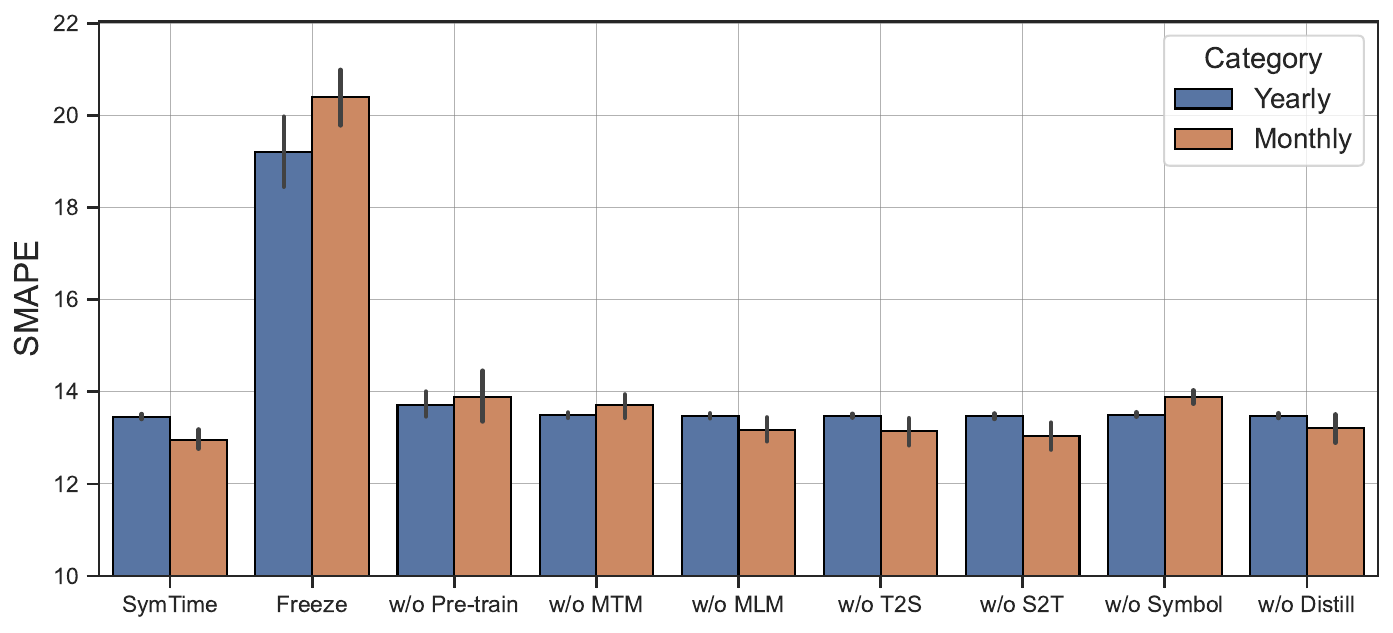}
    \caption{Ablation on pre-training objectives.}
\end{subfigure}
\hfill
\begin{subfigure}{0.4\textwidth}
    \includegraphics[width=\linewidth]{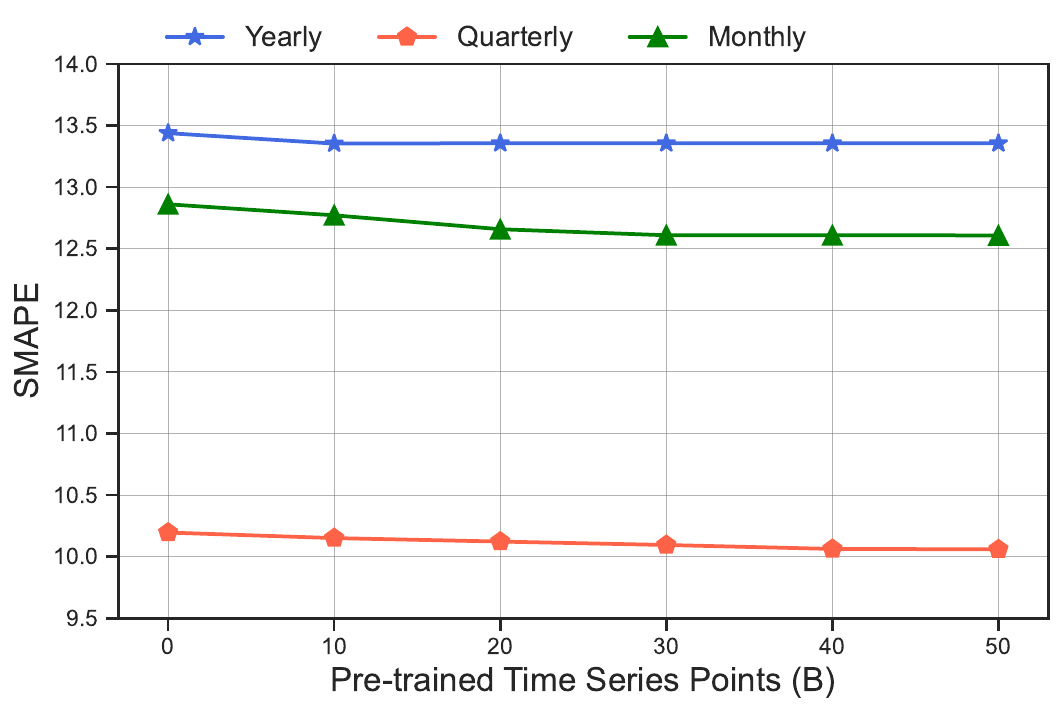}
    \caption{Ablation on the size of pre-training dataset.}
\end{subfigure}
\caption{Ablation experiments on short-term forecasting task.}
\label{figure: ablation on short}
\end{figure*}

\textbf{Setup.} We adopt the same experimental setup as in Section \ref{sec:ablation experiments} to conduct ablation studies on short-term time series forecasting tasks. We first select the Yearly and Monthly sub-datasets from the M4 benchmark dataset \cite{M4team2018dataset} to perform ablation experiments on SymTime's pre-training objectives. We choose SMAPE as the evaluation metric and the results are shown in Figure \ref{figure: ablation on short} (a). Subsequently, we select the Yearly, Quarterly and Monthly sub-datasets to verify the impact of the pre-training dataset size on downstream task performance. The results are shown in Figure \ref{figure: ablation on short} (b), where 0 indicates no pre-training.

\textbf{Results.} Figure \ref{figure: ablation on short} (a) indicates that SymTime's performance drops sharply when the backbone encoder is frozen and no pre-training is conducted. When some pre-training objectives are removed, the model's performance in short-term time series forecasting also declines, but the sensitivity of performance degradation is not as pronounced as in long-term forecasting experiments. Figure \ref{figure: ablation on short} (a) shows that as the size of the pre-training dataset increases, SymTime's performance on the Quarterly dataset improves significantly.

\subsection{The Impact and Ablation of Pre-interpolation on Time Series Imputation Task}
\label{sec:pre-interpolation}

\begin{table}[!t]
\caption{Ablation Experiments of pre-interpolation in inputation task on ECL dataset. The per-interpolation results for Peri-midFormer, TimesNet, PatchTST, DLinear and Pyraformer are copied from \cite{Peri-midFormer}.}
\centering
\vskip 0.10in
\begin{threeparttable}
\begin{small}
\setlength{\extrarowheight}{1.5pt}
\setlength{\tabcolsep}{6pt}
\begin{tabular}{c|c|cccc|cccc}
\toprule
\multirow{2}{*}{Methods} & \multirow{2}{*}{Metric} & \multicolumn{4}{c}{w/o per-interpolation} & \multicolumn{4}{c}{with per-interpolation} \\
\cmidrule(lr){3-6} \cmidrule(lr){7-10}  
 & & 0.125 & 0.25 & 0.375 & 0.5 & 0.125 & 0.25 & 0.375 & 0.5 \\
\midrule
\multirow{2}{*}{Per-interpolation} & MSE & - & - & - & - & 0.086 & 0.110 & 0.149 & 0.206 \\
 & MAE & - & - & - & - & 0.188 & 0.213 & 0.251 & 0.301 \\
\midrule

\multirow{2}{*}{\textbf{SymTime (Ours)}} & MSE & 0.050 & 0.064 & 0.074 & 0.092 & 0.037 & 0.047 & 0.060 & 0.075 \\
 & MAE & 0.145 & 0.169 & 0.181 & 0.206 & 0.122 & 0.139 & 0.160 & 0.181 \\
\midrule

\multirow{2}{*}{Peri-midFormer \cite{Peri-midFormer}} & MSE & 0.073 & 0.092 & 0.107 & 0.122 & 0.047 & 0.053 & 0.067 & 0.085 \\
 & MAE & 0.187 & 0.214 & 0.231 & 0.248 & 0.140 & 0.162 & 0.179 & 0.195 \\
\midrule
 
\multirow{2}{*}{TimesNet \cite{TimesNet}} & MSE & 0.088 & 0.092 & 0.096 & 0.102 & 0.081 & 0.083 & 0.086 & 0.091 \\
 & MAE & 0.203 & 0.208 & 0.214 & 0.221 & 0.196 & 0.198 & 0.201 & 0.207 \\
\midrule
 
\multirow{2}{*}{PatchTST \cite{PatchTST}} & MSE & 0.061 & 0.072 & 0.082 & 0.097 & 0.050 & 0.059 & 0.070 & 0.087 \\
 & MAE & 0.170 & 0.185 & 0.198 & 0.216 & 0.148 & 0.164 & 0.181 & 0.202 \\
\midrule
 
\multirow{2}{*}{DLinear \cite{DLinear}} & MSE & 0.084 & 0.113 & 0.141 & 0.173 & 0.050 & 0.062 & 0.789 & 0.105 \\
 & MAE & 0.206 & 0.243 & 0.273 & 0.303 & 0.144 & 0.164 & 0.189 & 0.225 \\
\midrule

\multirow{2}{*}{Pyraformer \cite{Pyraformer}} & MSE & 0.297 & 0.294 & 0.296 & 0.299 & 0.165 & 0.165 & 0.171 & 0.173 \\
 & MAE & 0.383 & 0.380 & 0.381 & 0.383 & 0.290 & 0.291 & 0.293 & 0.295 \\
\bottomrule
\end{tabular}
\end{small}
\end{threeparttable}
\label{table:pre-inter}
\end{table}

\begin{figure*}[!t]
\centering
\begin{subfigure}{0.33\textwidth}
    \includegraphics[width=\linewidth]{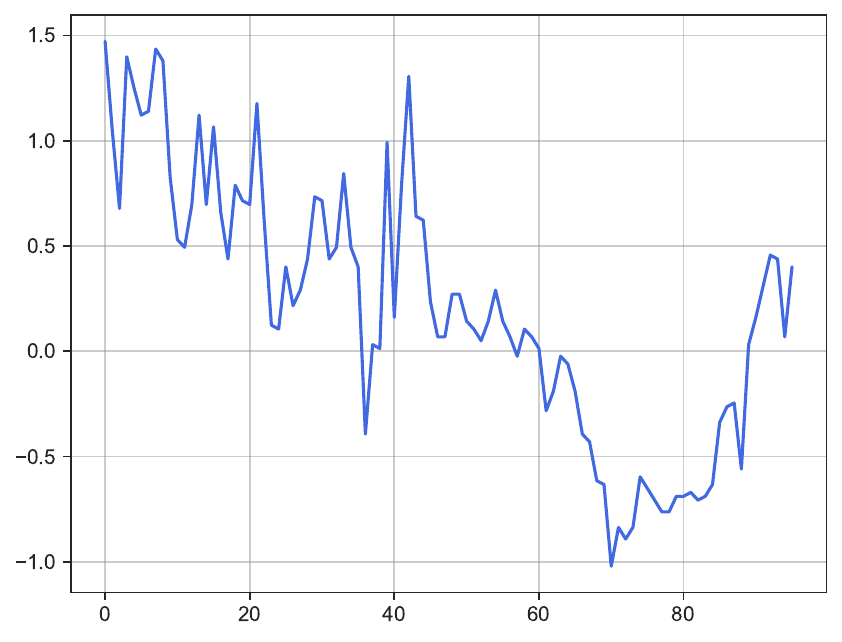}
    \caption{Original data}
\end{subfigure}
\hfill
\begin{subfigure}{0.33\textwidth}
    \includegraphics[width=\linewidth]{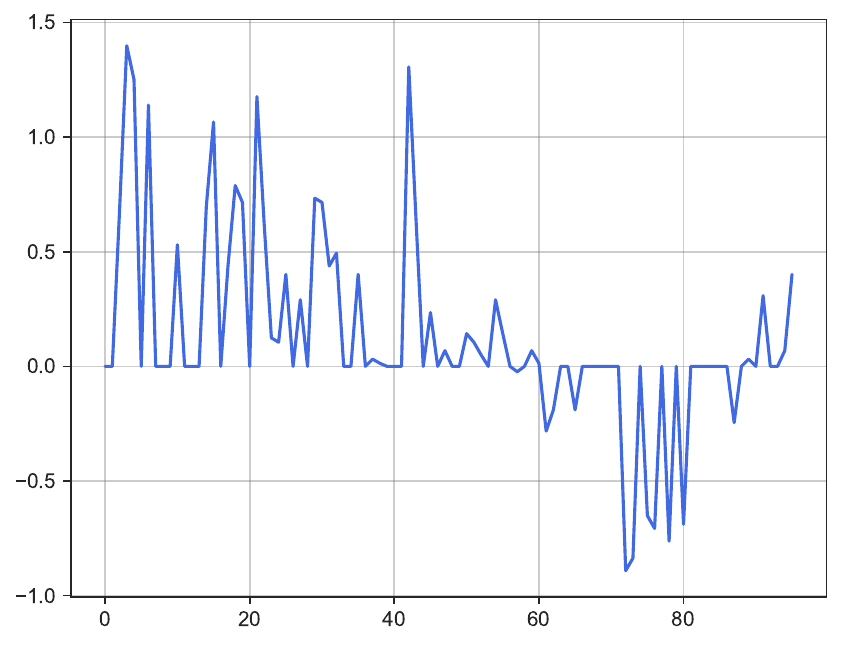}
    \caption{Data with 50\% missing values}
\end{subfigure}
\medskip
\begin{subfigure}{0.33\textwidth}
    \includegraphics[width=\linewidth]{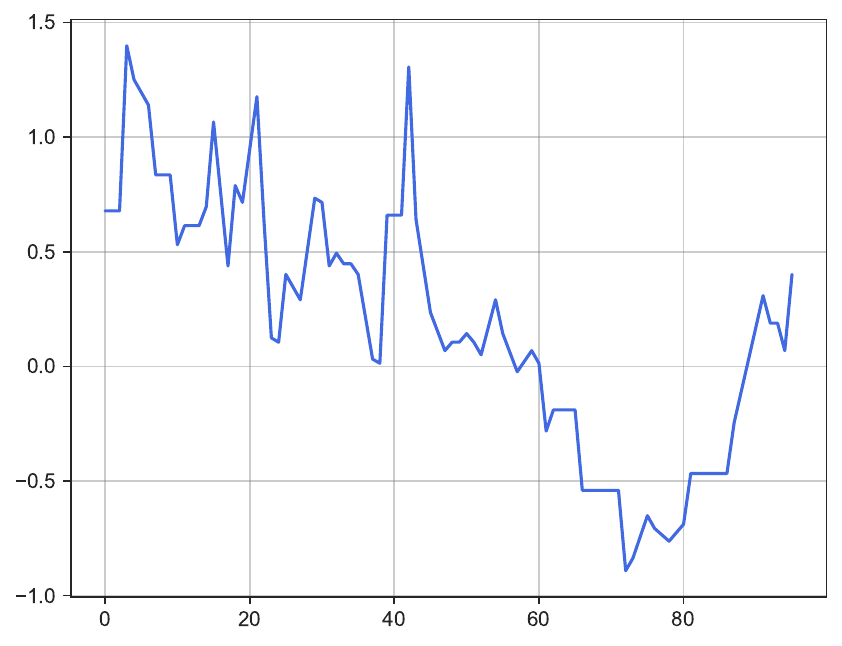}
    \caption{Per-interpolated data}
\end{subfigure}
\vskip -0.2in
\caption{Visualization of original data, data with 50\% missing values and pre-interpolated data of ECL dataset.}
\label{figure: per-interpolated}
\end{figure*}

\texttt{SymTime} adds masks randomly at the patch level during pre-training for time series reconstruction.  While in the imputation task, masks are added randomly at the data point level. Additionally, high masking rates may disrupt the original trend and periodic features of the time series. Therefore, we use Peri-midFormer's method to apply per-interpolation to the masked time series to restore the disrupted periodic features \cite{Peri-midFormer, TimesNet, TimeMixer, TimeMixer++, SIN, CauDiTS}. It is important to note that this method is general and independent of deep learning models. The use of this method aims to further enhance the potential of deep learning models. To further verify the effectiveness and impact of the per-interpolation method, we conduct experiments on the ECL time series imputation dataset, with results shown in Table \ref{table:pre-inter}. Taking the ECL time series dataset 0.5 mask ratio as an example, the effect of pre-interpolation is shown in Figure \ref{figure: per-interpolated}. We perform experiments with masking rate of $\{0.125, 0.25, 0.375, 0.50\}$ and compare models such as Peri-midFormer \cite{Peri-midFormer}, TimesNet \cite{TimesNet}, PatchTST \cite{PatchTST}, DLinear \cite{DLinear} and Pyraformer \cite{Pyraformer}. Per-interpolation represents the experimental results obtained using only linear interpolation. For a missing time series $x_t$ at time $t$, the method can be described as:
\begin{equation}
    x_{t} = \left\{\begin{matrix} 
\frac{x_{t - 1} + x_{t + 1}}{2}, & if \ (x_{t - 1}) \ne \mathrm{None} \ \& \ (x_{t + 1} \ne \mathrm{None} ) \\
x_{t + 1}, & if \ (x_{t - 1}) = \mathrm{None} \ \& \ (x_{t + 1} \ne \mathrm{None} ) \\
x_{t - 1}, & if \ (x_{t - 1}) \ne \mathrm{None} \ \& \ (x_{t + 1} = \mathrm{None} )
\end{matrix}\right. 
,
\label{eq: per-interpolated}
\end{equation}
where, $x_{t-1}$ and $x_{t+1}$ represent the values at the previous and next time points, respectively, while None indicates a missing value. The results in Table \ref{table:pre-inter} show that this method significantly improves the performance of all models in a model-independent manner.

\section{Related Work}
\label{sec:related work}

\subsection{Time Series Foundation Models}

In CV and NLP \cite{CLIP}, PTFMs have been demonstrated to adapt to a variety of downstream tasks after fine-tuning on specific datasets, exhibiting excellent generalization and scalability. Inspired by this, recent years have seen significant progress in PTFMs for TSA \cite{Transformer-in-TSA, KDD-Survey}, with the emergence of various pre-training methods. MOIRAI, through MTSM and reconstruction, has been pre-trained on large datasets (27B), yielding a universal forecasting model with significant zero-shot advantages \cite{MOIRAI}. Timer, after generative pre-training on large datasets (1B), has performed well in forecasting \cite{Timer}. TimeGPT trained a encoder-decoder Transformer with 100B data \cite{TimeGPT}. COMET, using multi-level contrastive learning on a large ECG dataset, has obtained a medical time series PTFMs with few-shot advantages \cite{COMET}. 

%Although these methods attempt to train large foundation models on large datasets, the existing time series datasets are still smaller compared to CV and NLP \cite{ALBEF}, failing to comprehensively cover all the representations of time series and facing issues of data distribution imbalance, which can lead to predictive biases in PTFMs \cite{ALBEF}. To address this issue, this paper generates 50 million pairs of series-symbol data, with a total sequence length of 60B, to cover all representations of time series as much as possible and to pre-train \texttt{SymTime} on this data to tackle the problem of data scarcity.

As discussed in Appendix Section \ref{sec:Analysis of Existing Dataset}, these baseline models still face challenges related to data scarcity and data imbalance. In the next section, we introduce the proposed data generation mechanism and the corresponding dual-modality foundation model designed to address these issues.

\subsection{Deep Learning and Symbolic Regression}

The central thesis of this paper is to regard time series as representations of complex dynamical systems \cite{AI-Feynman}. Traditionally, complex systems are modeled by observing time series utilizing ODE and PDE \cite{DE4ModelingCS}. With the advancement of machine learning, symbolic regression (SR) \cite{Exhaustive}, as a supervised learning method, can discover hidden mathematical expressions from numerical series. Although genetic algorithms (GAs) are the mainstream approach for SR \cite{GA4SR1, GA4SR2}, deep learning-based methods have also made significant progress. \cite{Symbolic} constructed an end-to-end SR model using Transformers, while SNIP built a large-scale pre-trained model through contrastive learning on symbolic expressions and numerical observations \cite{SNIP}. Both methods treat symbolic expressions as nature language and use deep neural networks to learn their features. Therefore, this paper employs a pre-trained LLM as a symbol encoder to learn the features of symbolic expressions and jointly trains a time series foundation model imbued with semantic information through contrastive learning \cite{CauDiTS}.

\subsection{Time Series Forecasting Models Based on Synthetic Data}

Unlike the representation pre-training conducted on the large synthetic S2 dataset in this paper, previous TSA models trained on synthetic data were mainly based on Prior-data Fitted Networks (PFN) \cite{TDBI, PFN}. This model learns prior distributions from synthetic data using Bayesian methods, enabling zero-shot inference. ForecastPFN generated a large number of synthetic time series by separately modeling the seasonal trend, global trend and noise based on given constraint expressions \cite{ForecastPFN}. Although PFN trained in this way offered certain zero-shot and few-shot advantages, this approach was limited to generating time series through sampling fixed expressions and performing linear combinations. In contrast, the S2 data generation mechanism proposed in this paper can sample an infinite variety of symbolic expressions \cite{SNIP, Symbolic, DL4Symbolic, CauDiTS}. TimePFN constructed synthetic datasets by filtering real time series with linear and periodic convolution kernels, training PFN for zero-shot inference \cite{TimePFN}. However, this method depends on real-world time series for filtering and linear transformations between channels. Compared to the S2 data generation mechanism, it can not create large-scale and fully representative synthetic datasets for model pre-training.

\section{Visualization}
\label{sec:visualization}

\begin{figure*}[!t]
\centering
\begin{subfigure}{0.33\textwidth}
    \includegraphics[width=\linewidth]{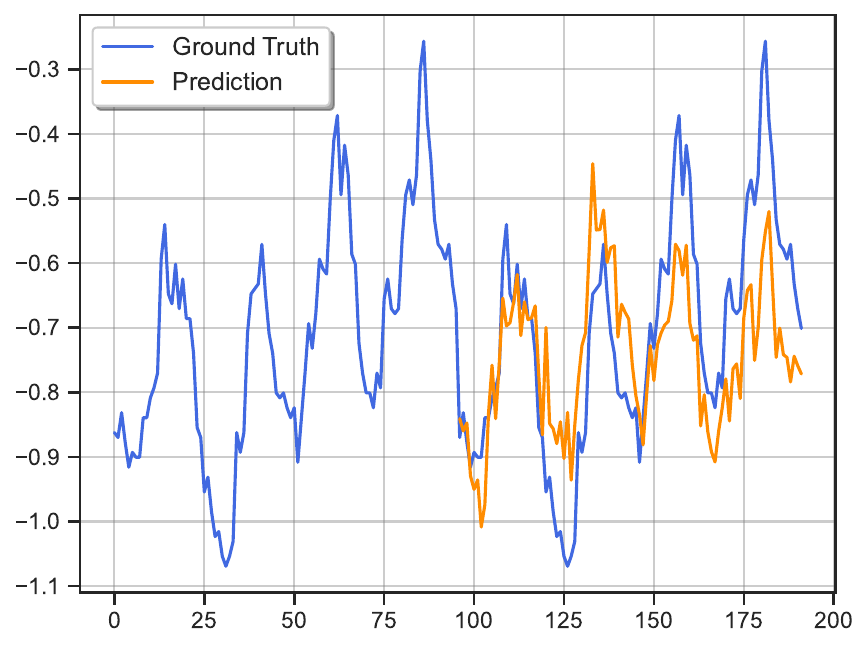}
    \caption{\textbf{SymTime}}
\end{subfigure}
\hfill
\begin{subfigure}{0.33\textwidth}
    \includegraphics[width=\linewidth]{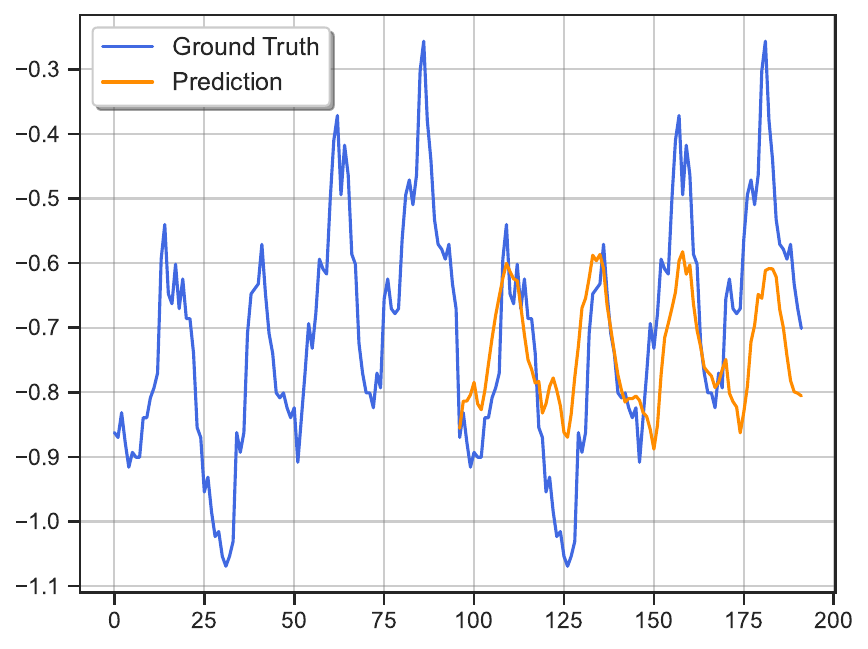}
    \caption{PatchTST}
\end{subfigure}
\medskip
\begin{subfigure}{0.33\textwidth}
    \includegraphics[width=\linewidth]{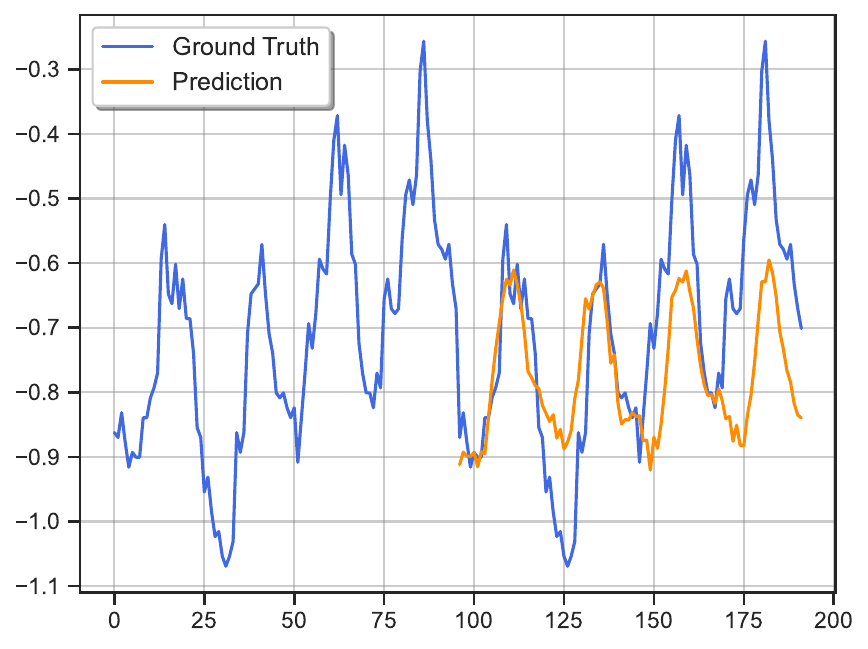}
    \caption{iTransformer}
\end{subfigure}
\hfill
\begin{subfigure}{0.33\textwidth}
    \includegraphics[width=\linewidth]{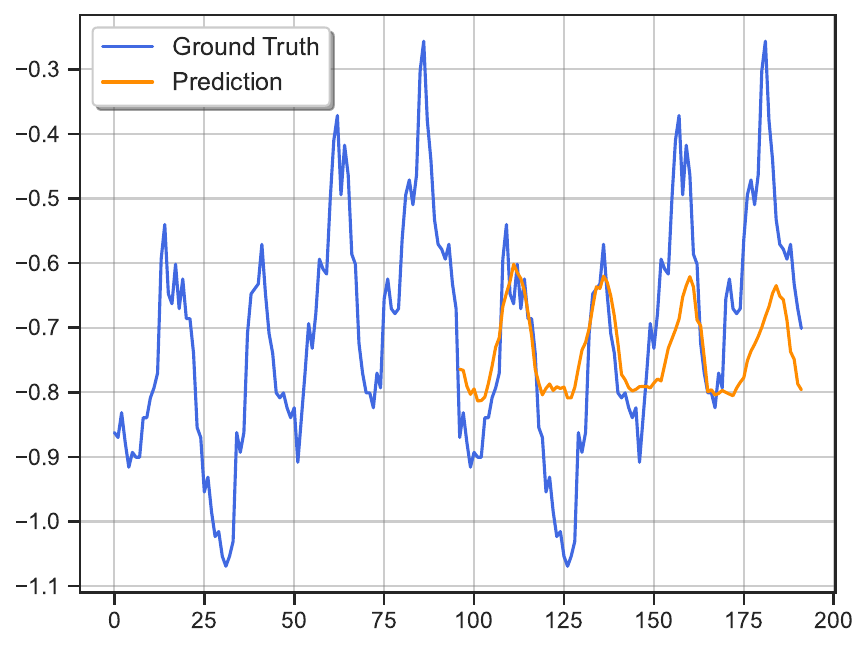}
    \caption{TimesNet}
\end{subfigure}
\medskip
\begin{subfigure}{0.33\textwidth}
    \includegraphics[width=\linewidth]{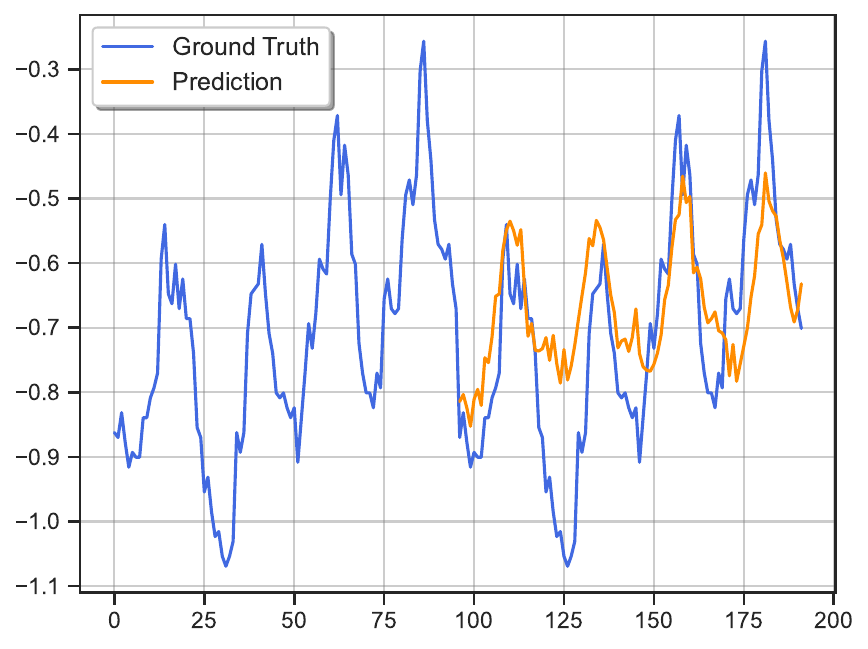}
    \caption{DLinear}
\end{subfigure}
\medskip
\begin{subfigure}{0.33\textwidth}
    \includegraphics[width=\linewidth]{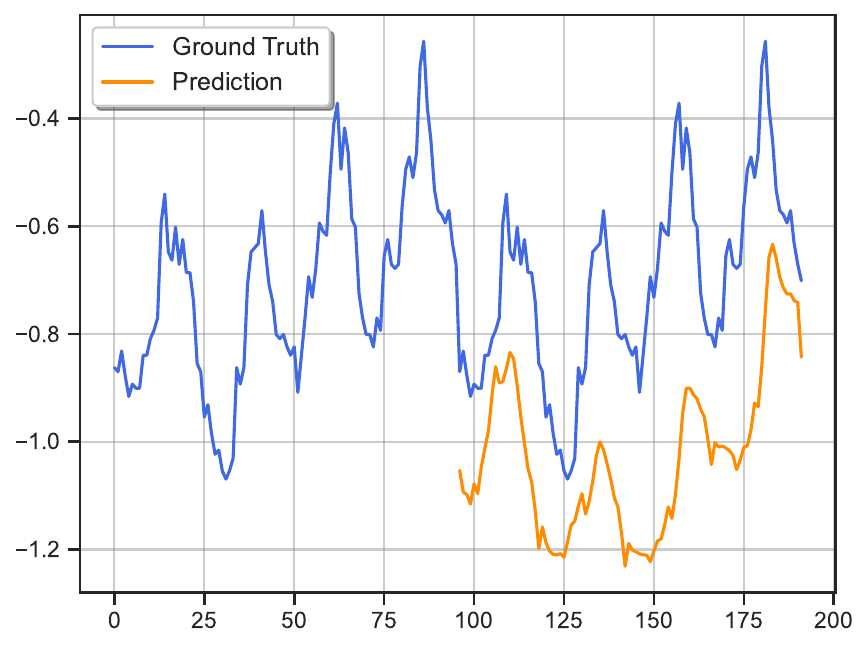}
    \caption{Autoformer}
\end{subfigure}
\vskip -0.2in
\caption{Visualization of long-term forecasting with 96 prediction length of ETTh1 dataset.}
\label{figure: long on ETTh1}
\end{figure*}

\begin{figure*}[!t]
\centering
\begin{subfigure}{0.33\textwidth}
    \includegraphics[width=\linewidth]{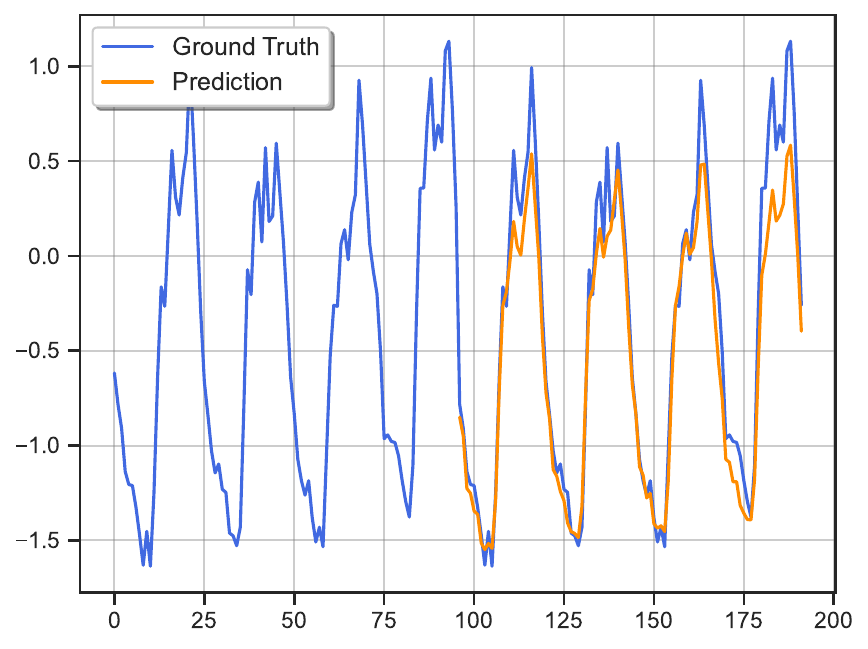}
    \caption{\textbf{SymTime}}
\end{subfigure}
\hfill
\begin{subfigure}{0.33\textwidth}
    \includegraphics[width=\linewidth]{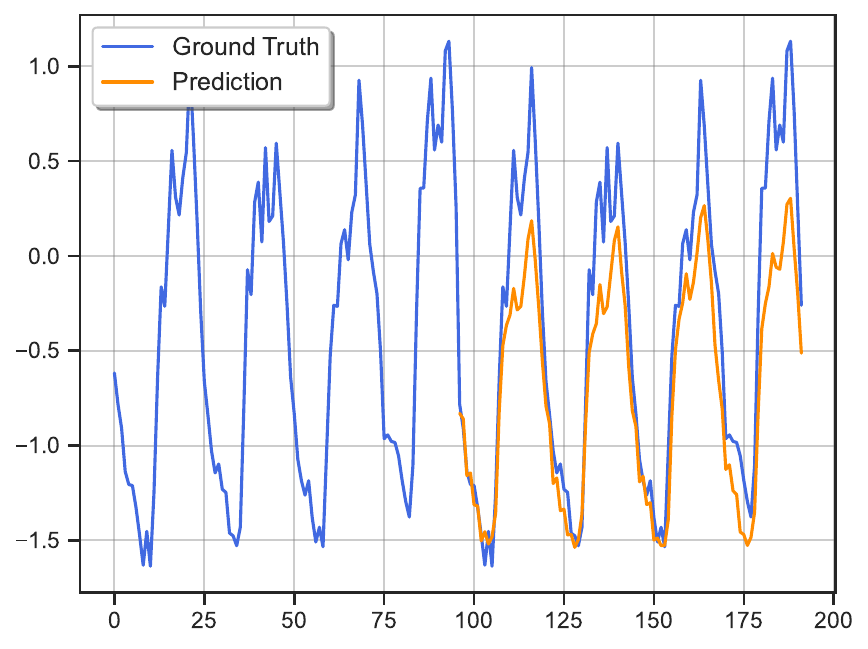}
    \caption{PatchTST}
\end{subfigure}
\medskip
\begin{subfigure}{0.33\textwidth}
    \includegraphics[width=\linewidth]{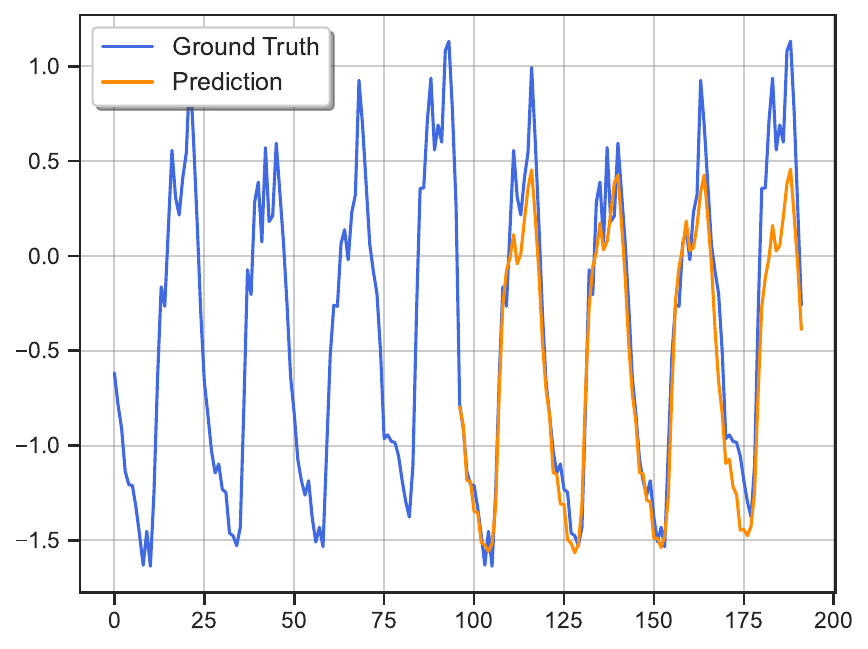}
    \caption{iTransformer}
\end{subfigure}
\hfill
\begin{subfigure}{0.33\textwidth}
    \includegraphics[width=\linewidth]{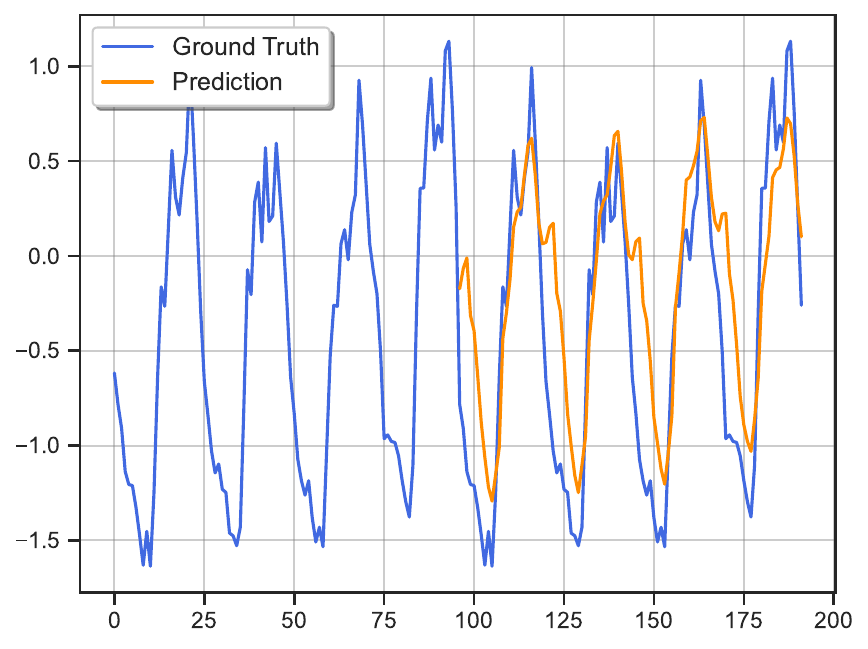}
    \caption{TimesNet}
\end{subfigure}
\medskip
\begin{subfigure}{0.33\textwidth}
    \includegraphics[width=\linewidth]{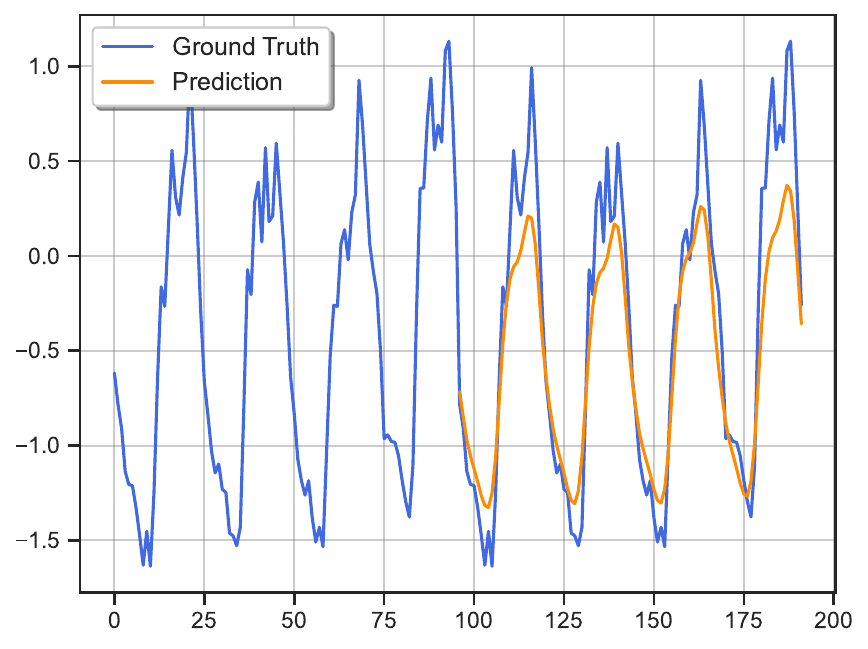}
    \caption{DLinear}
\end{subfigure}
\medskip
\begin{subfigure}{0.33\textwidth}
    \includegraphics[width=\linewidth]{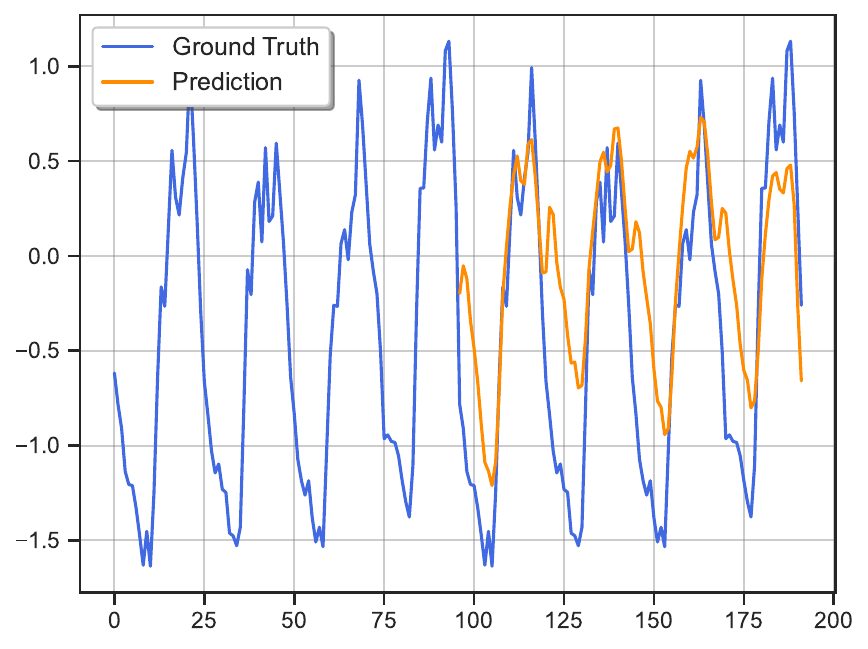}
    \caption{Autoformer}
\end{subfigure}
\vskip -0.2in
\caption{Visualization of long-term forecasting with 96 prediction length of Electricity dataset.}
\label{figure: long on ECL}
\end{figure*}

\subsection{Long-term Time Series Forecasting with 96 Prediction on ETTh1 (Figure \ref{figure: long on ETTh1}) and ECL (Figure \ref{figure: long on ECL})}

\begin{figure*}[ht]
\centering
\begin{subfigure}{0.33\textwidth}
    \includegraphics[width=\linewidth]{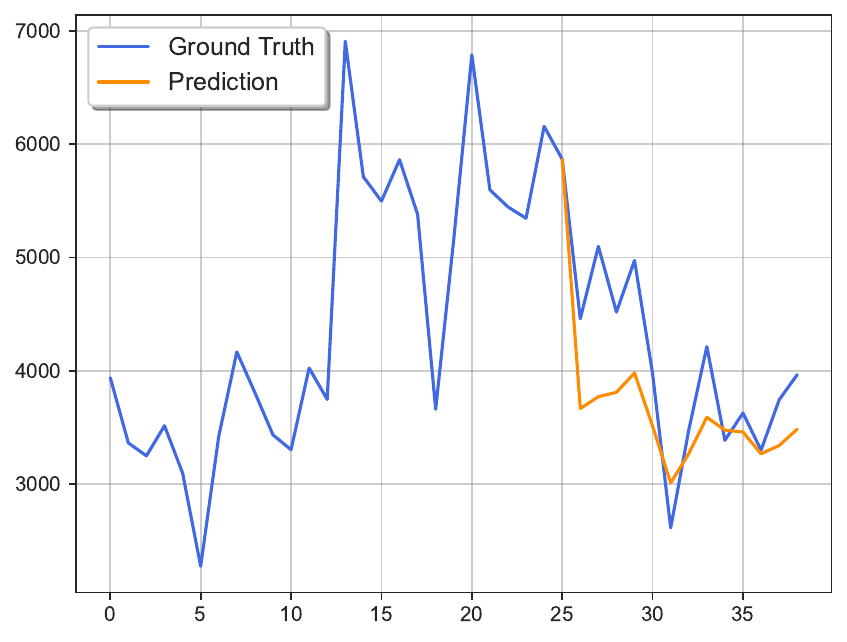}
    \caption{\textbf{SymTime}}
\end{subfigure}
\hfill
\begin{subfigure}{0.33\textwidth}
    \includegraphics[width=\linewidth]{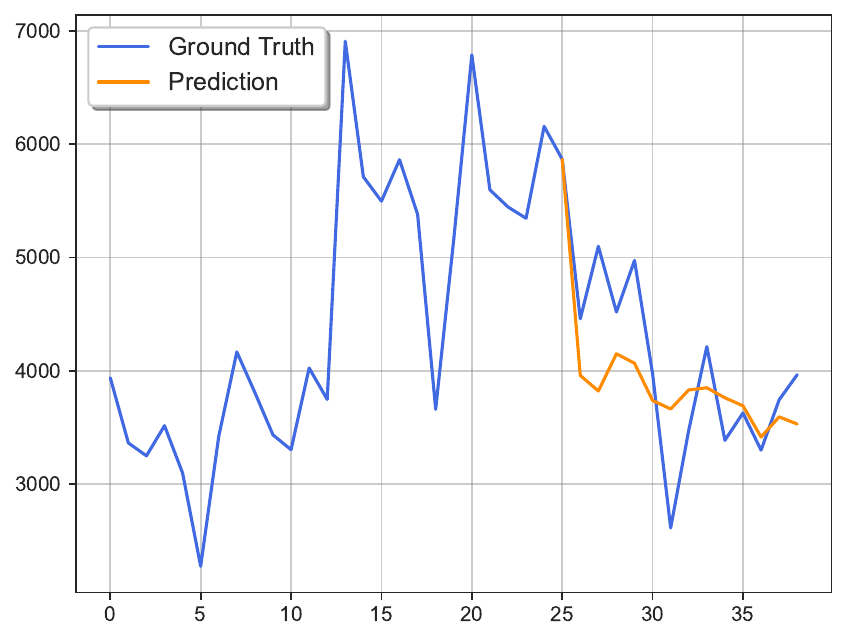}
    \caption{PatchTST}
\end{subfigure}
\medskip
\begin{subfigure}{0.33\textwidth}
    \includegraphics[width=\linewidth]{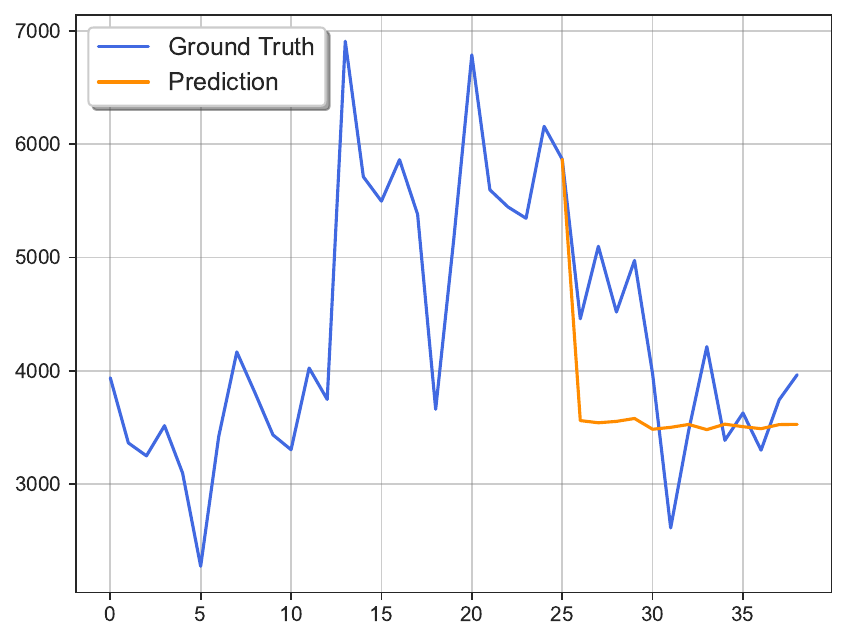}
    \caption{iTransformer}
\end{subfigure}
\hfill
\begin{subfigure}{0.33\textwidth}
    \includegraphics[width=\linewidth]{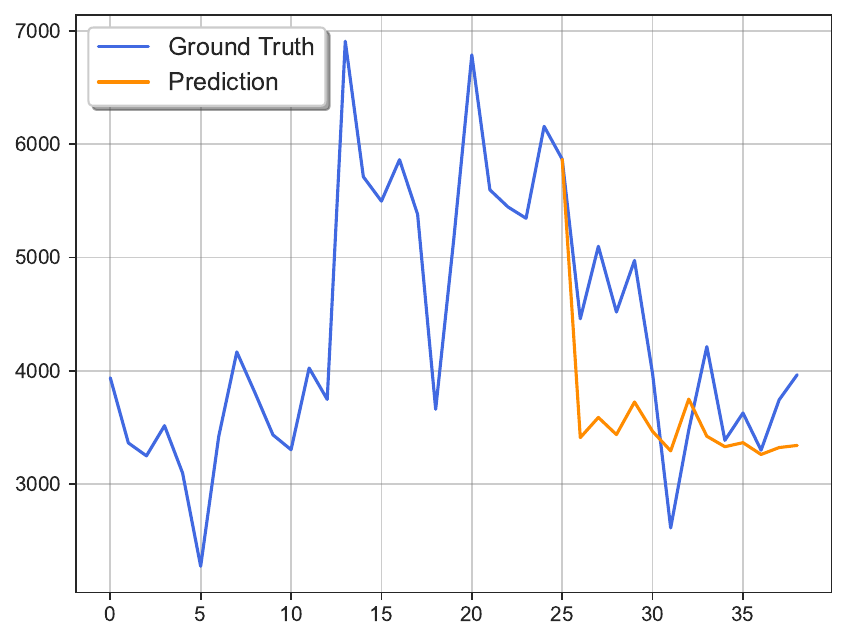}
    \caption{TimesNet}
\end{subfigure}
\medskip
\begin{subfigure}{0.33\textwidth}
    \includegraphics[width=\linewidth]{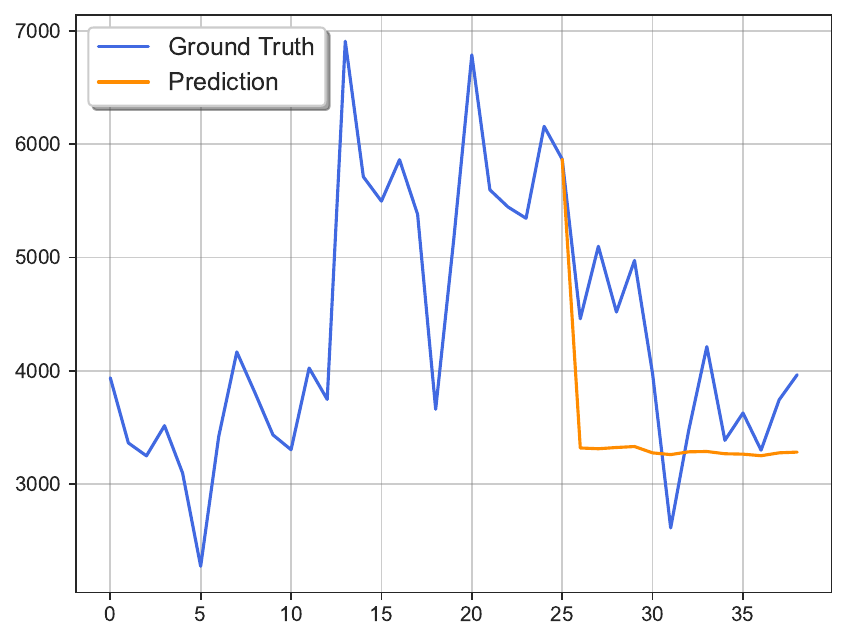}
    \caption{DLinear}
\end{subfigure}
\medskip
\begin{subfigure}{0.33\textwidth}
    \includegraphics[width=\linewidth]{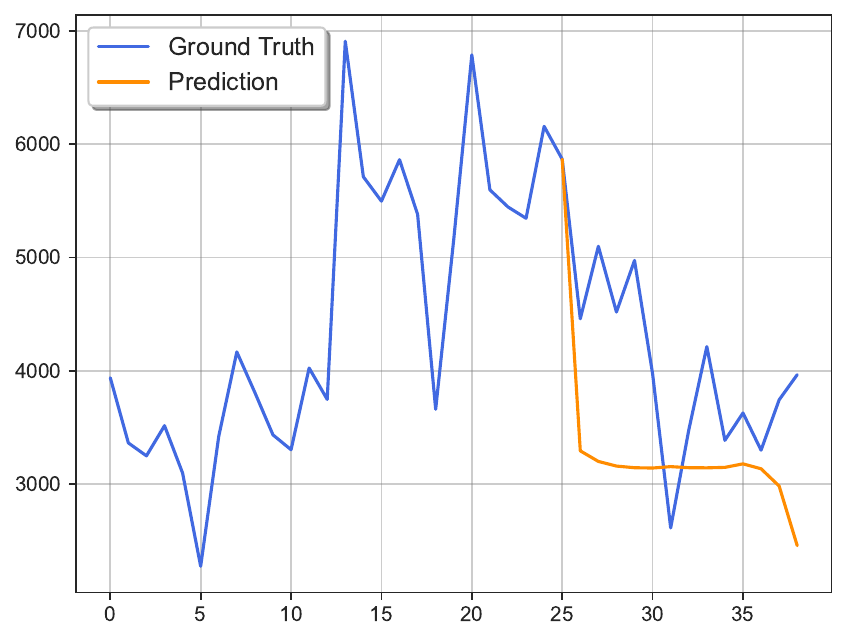}
    \caption{Autoformer}
\end{subfigure}
\caption{Visualization of time series short-term forecasting in M4 dataset Weekly.}
\label{figure: short on M4 Weekly}
\end{figure*}

\begin{figure*}[ht]
\centering
\begin{subfigure}{0.33\textwidth}
    \includegraphics[width=\linewidth]{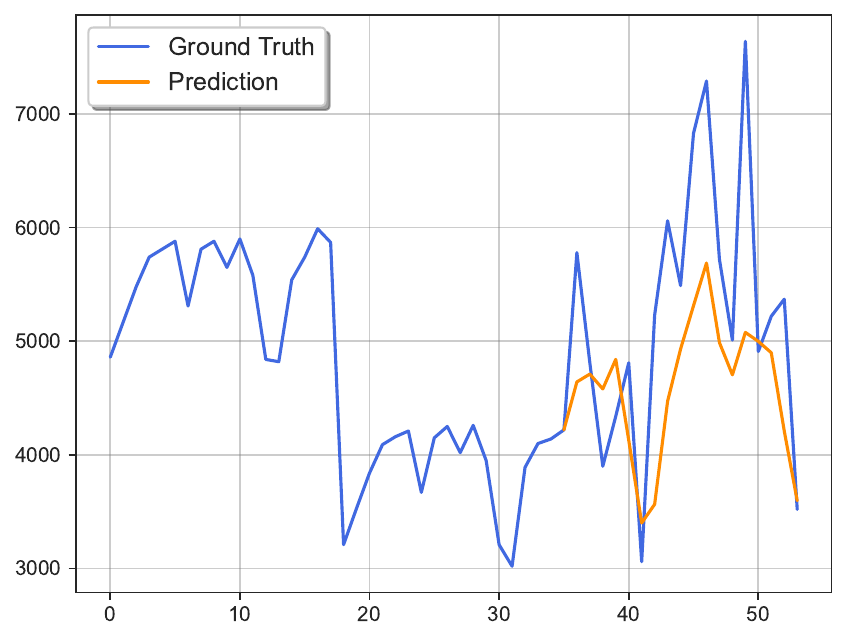}
    \caption{\textbf{SymTime}}
\end{subfigure}
\hfill
\begin{subfigure}{0.33\textwidth}
    \includegraphics[width=\linewidth]{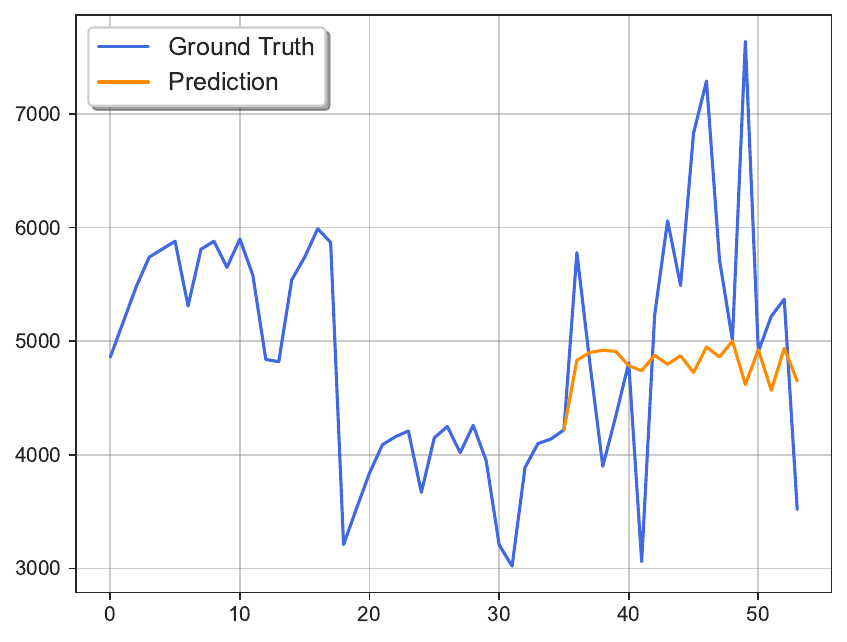}
    \caption{PatchTST}
\end{subfigure}
\medskip
\begin{subfigure}{0.33\textwidth}
    \includegraphics[width=\linewidth]{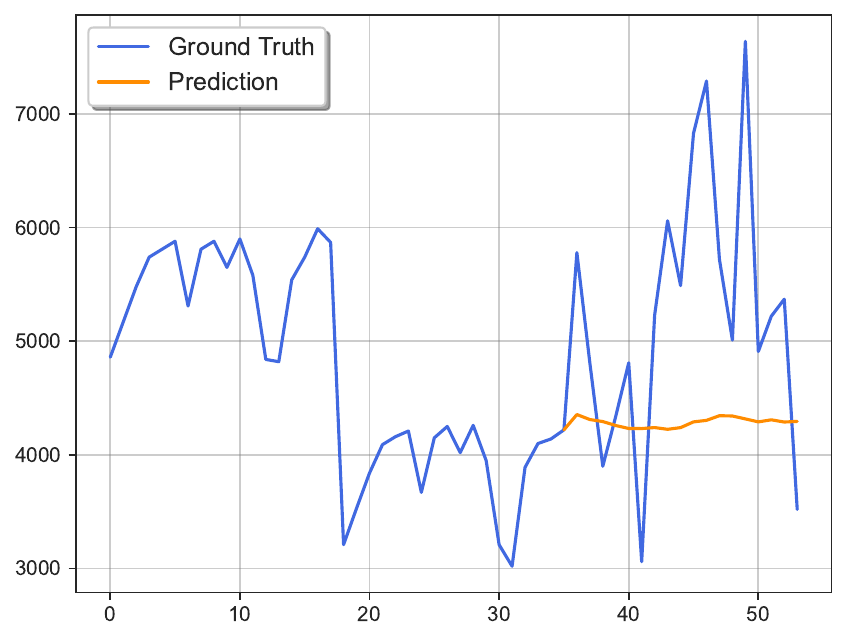}
    \caption{iTransformer}
\end{subfigure}
\hfill
\begin{subfigure}{0.33\textwidth}
    \includegraphics[width=\linewidth]{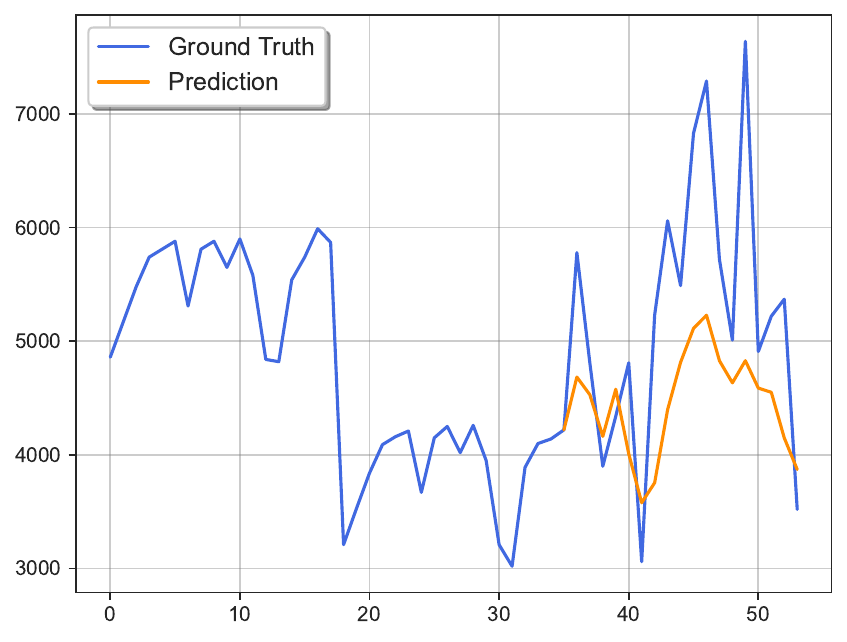}
    \caption{TimesNet}
\end{subfigure}
\medskip
\begin{subfigure}{0.33\textwidth}
    \includegraphics[width=\linewidth]{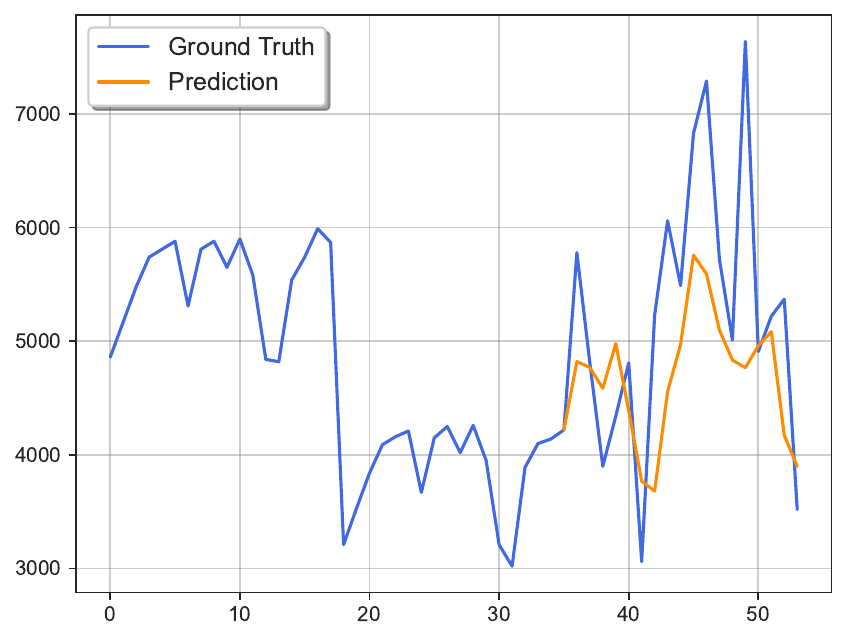}
    \caption{DLinear}
\end{subfigure}
\medskip
\begin{subfigure}{0.33\textwidth}
    \includegraphics[width=\linewidth]{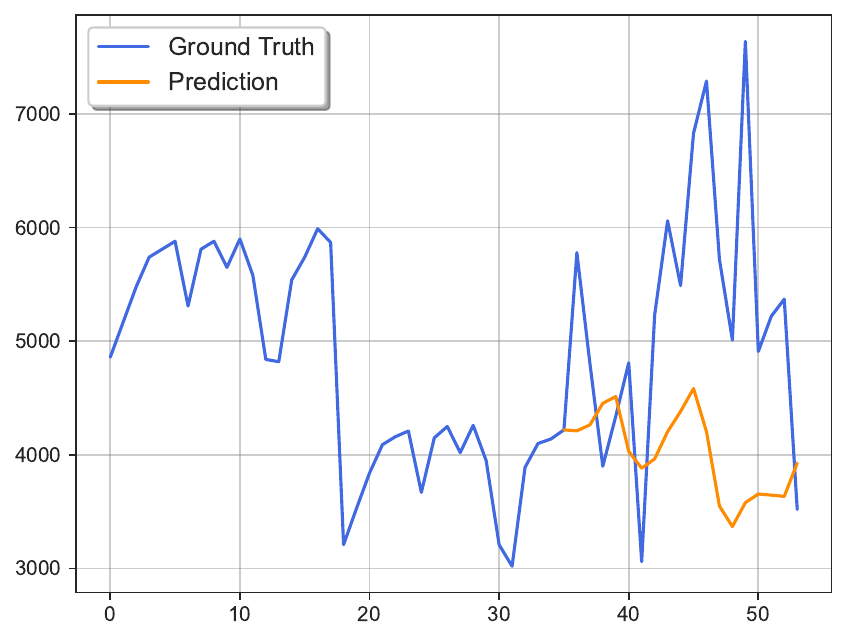}
    \caption{Autoformer}
\end{subfigure}
\caption{Visualization of time series short-term forecasting in M4 dataset Monthly.}
\label{figure: short on M4 Monthly}
\end{figure*}

\subsection{Short-term Time Series Forecasting on M4 Weekly (Figure \ref{figure: short on M4 Weekly}) and Monthly (Figure \ref{figure: short on M4 Monthly})}

\begin{figure*}[ht]
\centering
\begin{subfigure}{0.33\textwidth}
    \includegraphics[width=\linewidth]{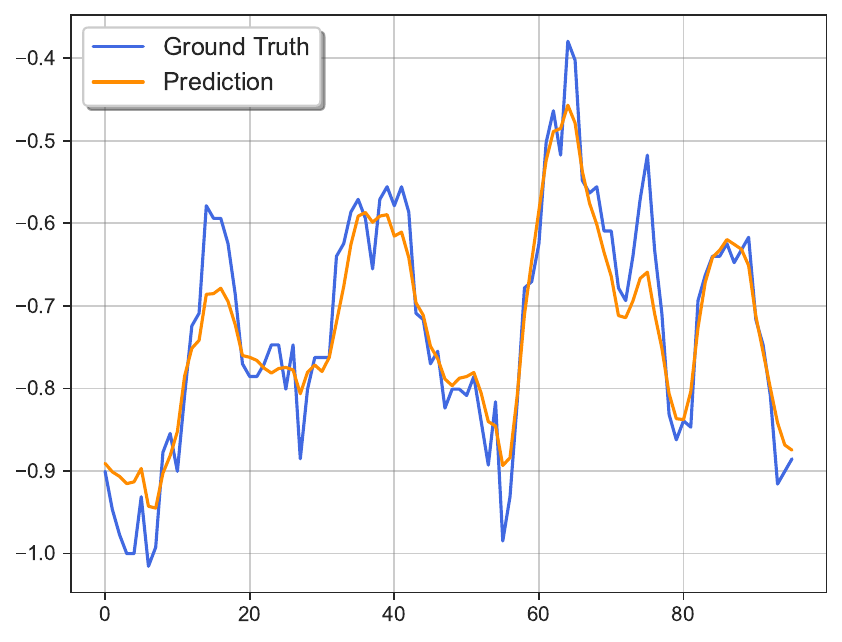}
    \caption{\textbf{SymTime}}
\end{subfigure}
\hfill
\begin{subfigure}{0.33\textwidth}
    \includegraphics[width=\linewidth]{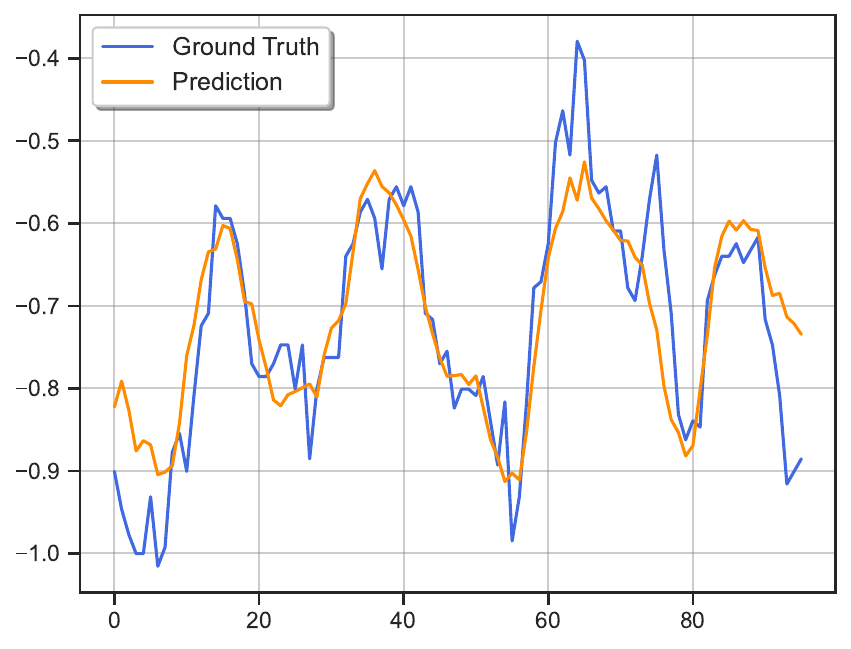}
    \caption{PatchTST}
\end{subfigure}
\medskip
\begin{subfigure}{0.33\textwidth}
    \includegraphics[width=\linewidth]{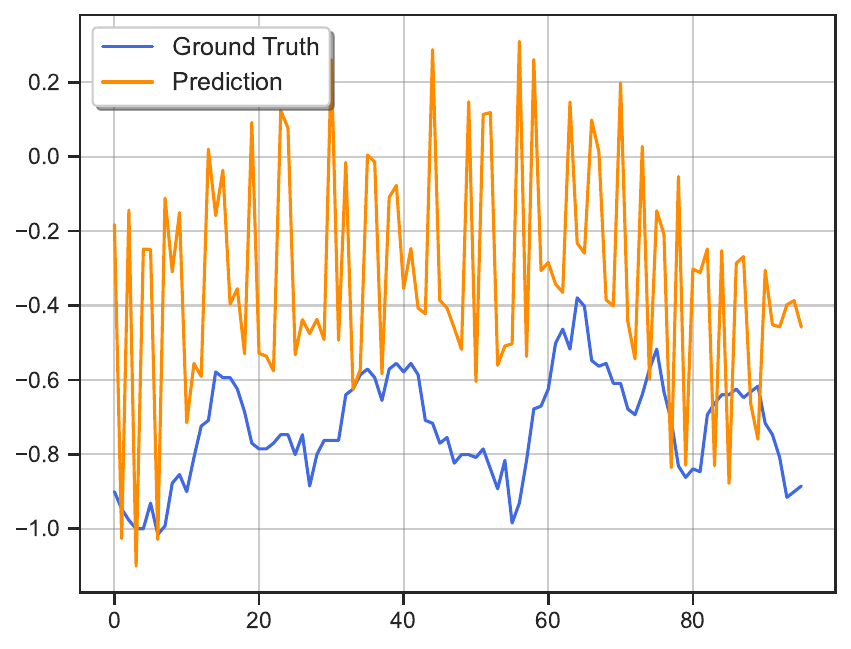}
    \caption{iTransformer}
\end{subfigure}
\hfill
\begin{subfigure}{0.33\textwidth}
    \includegraphics[width=\linewidth]{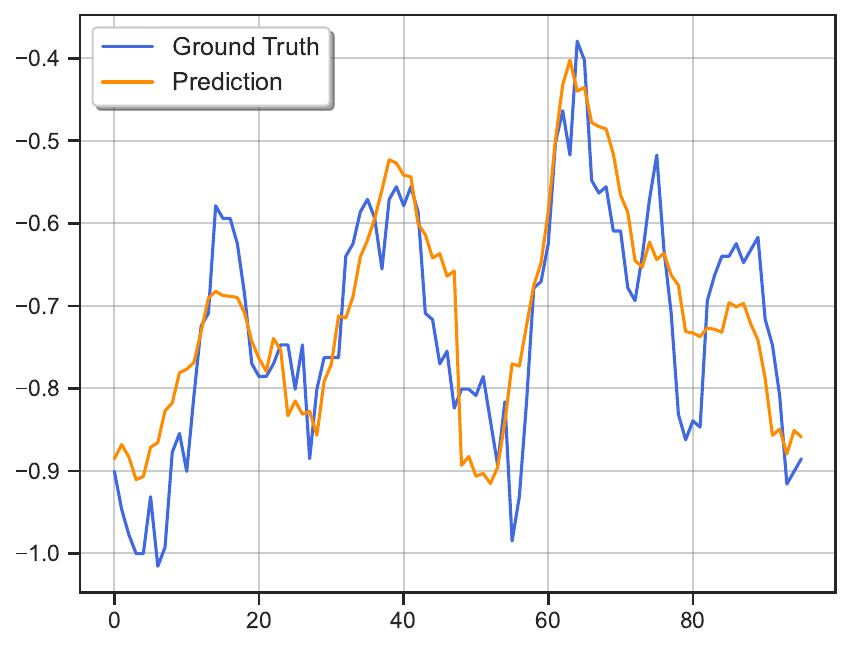}
    \caption{TimesNet}
\end{subfigure}
\medskip
\begin{subfigure}{0.33\textwidth}
    \includegraphics[width=\linewidth]{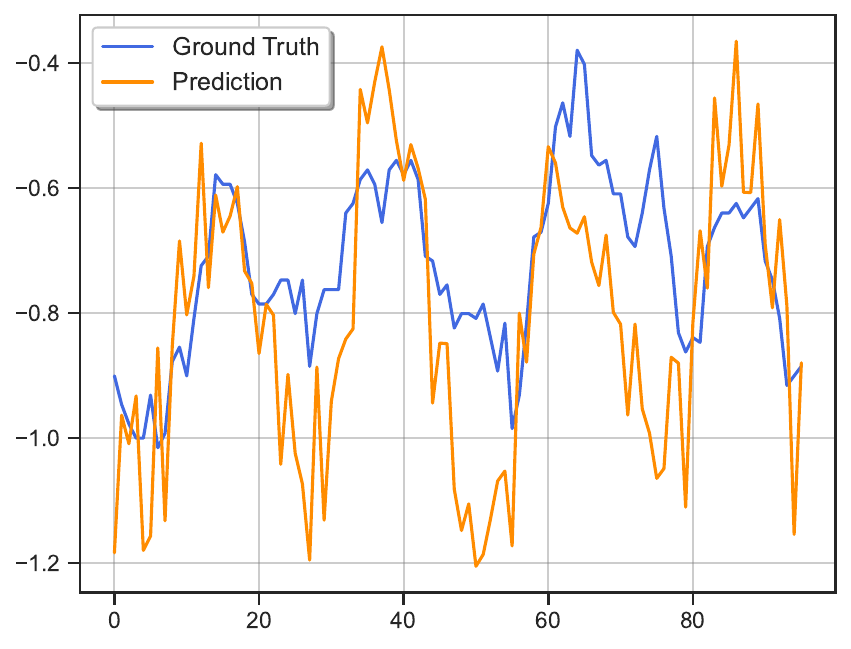}
    \caption{DLinear}
\end{subfigure}
\medskip
\begin{subfigure}{0.33\textwidth}
    \includegraphics[width=\linewidth]{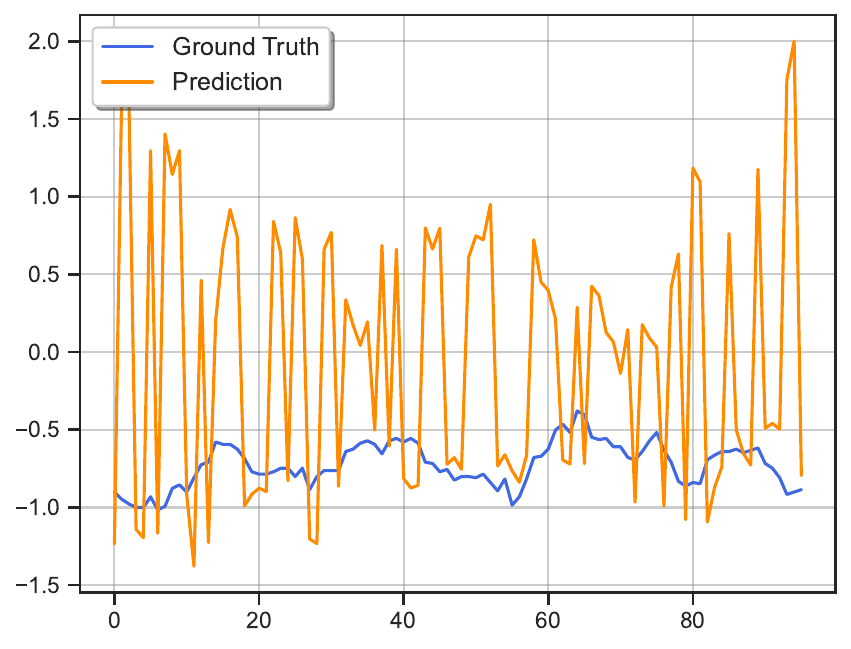}
    \caption{Autoformer}
\end{subfigure}
\caption{Visualization of time series imputation with 50\% mask rate of ETTh1 dataset.}
\label{figure: imputation on ETTh1}
\end{figure*}

\begin{figure*}[ht]
\centering
\begin{subfigure}{0.33\textwidth}
    \includegraphics[width=\linewidth]{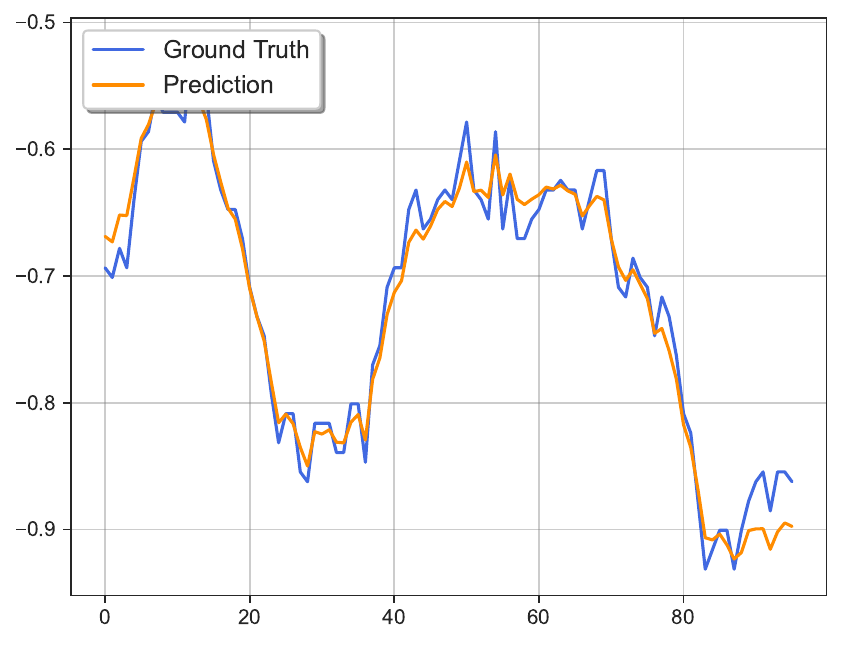}
    \caption{\textbf{SymTime}}
\end{subfigure}
\hfill
\begin{subfigure}{0.33\textwidth}
    \includegraphics[width=\linewidth]{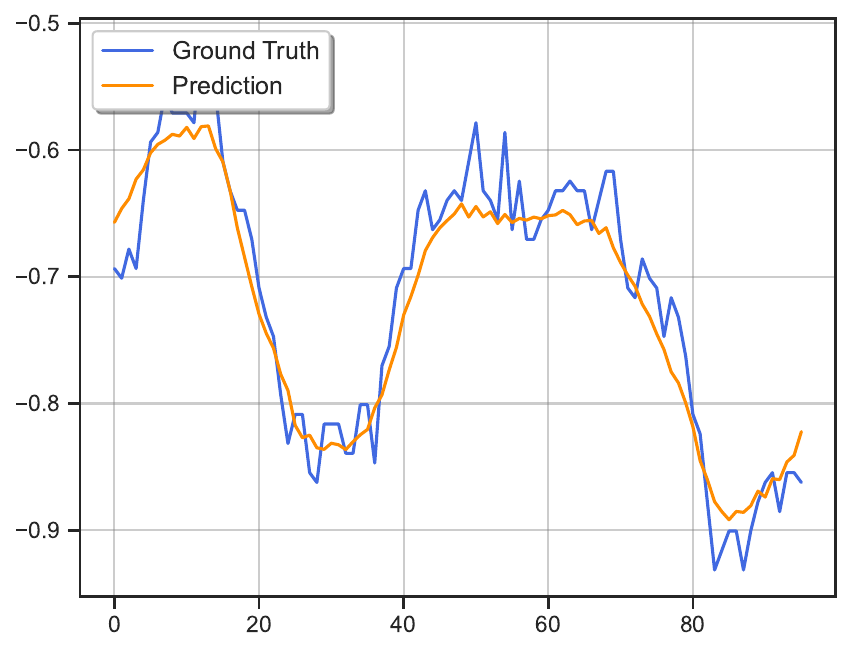}
    \caption{PatchTST}
\end{subfigure}
\medskip
\begin{subfigure}{0.33\textwidth}
    \includegraphics[width=\linewidth]{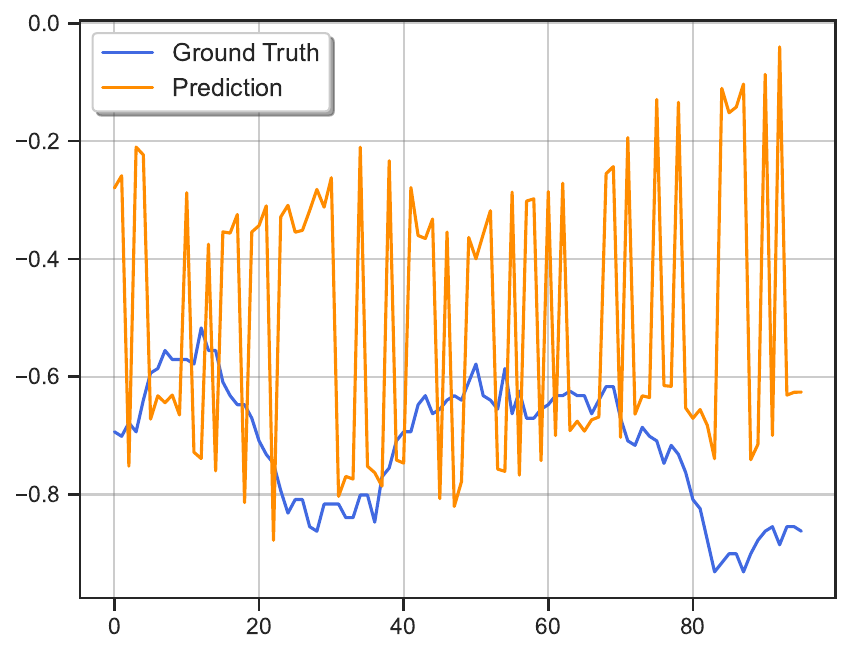}
    \caption{iTransformer}
\end{subfigure}
\hfill
\begin{subfigure}{0.33\textwidth}
    \includegraphics[width=\linewidth]{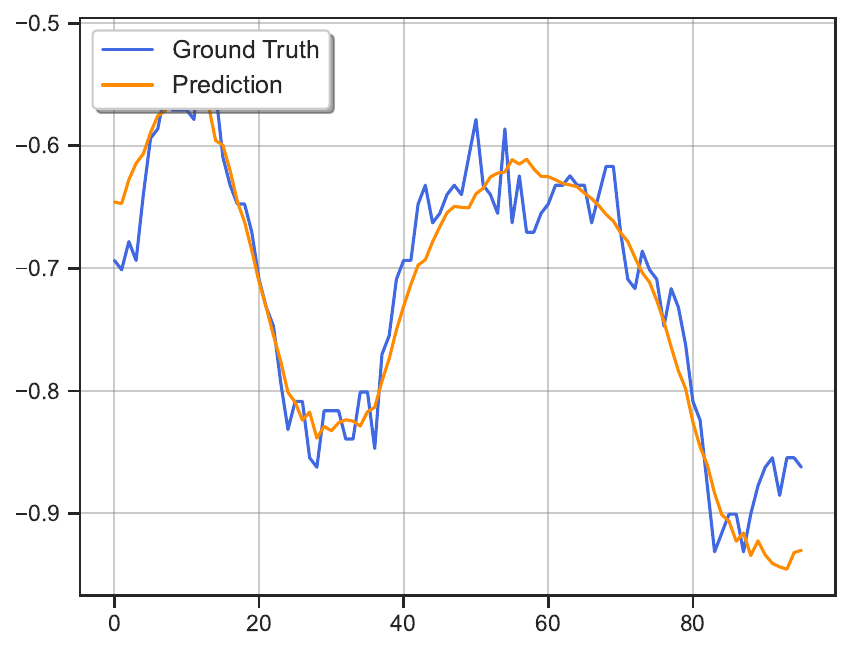}
    \caption{TimesNet}
\end{subfigure}
\medskip
\begin{subfigure}{0.33\textwidth}
    \includegraphics[width=\linewidth]{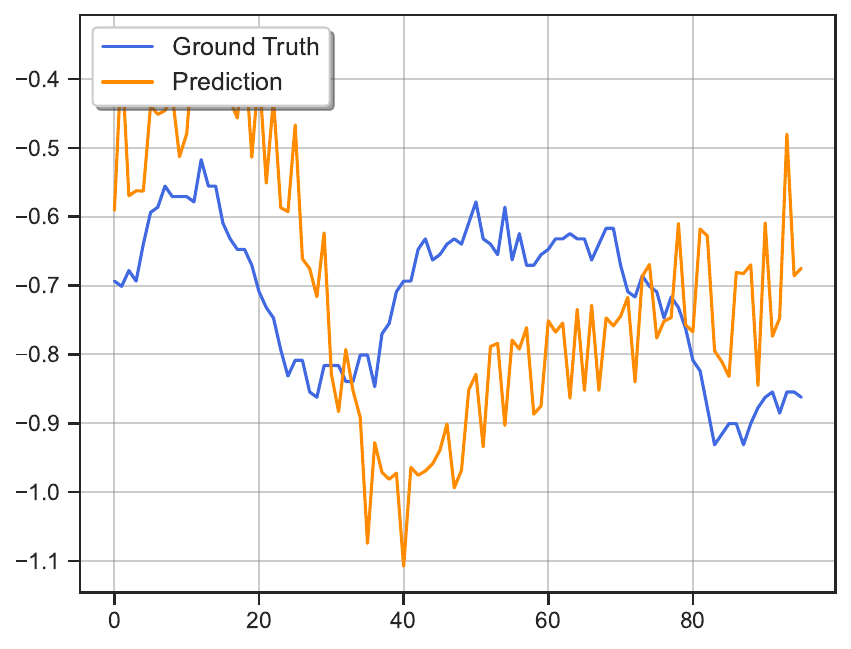}
    \caption{DLinear}
\end{subfigure}
\medskip
\begin{subfigure}{0.33\textwidth}
    \includegraphics[width=\linewidth]{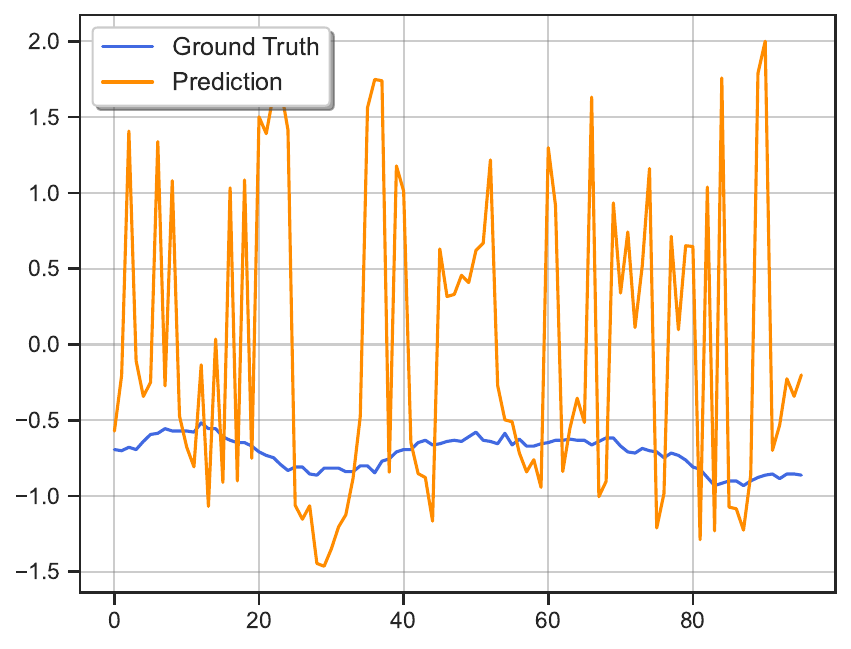}
    \caption{Autoformer}
\end{subfigure}
\caption{Visualization of time series imputation with 50\% mask rate of ETTm1 dataset.}
\label{figure: imputation on ETTm1}
\end{figure*}

\subsection{Time Series Imputation with 50\% mask rate on ETTh1 (Figure \ref{figure: imputation on ETTh1}) and ETTm1 (Figure \ref{figure: imputation on ETTm1})}

\section{Full Results}

For the five downstream TSA tasks results, we use (1) \href{https://github.com/WuQiangXDU/Peri-midFormer}{Peri-midFormer} for Peri-midFormer \cite{Peri-midFormer}, (2) \href{https://github.com/SalesforceAIResearch/uni2ts}{uni2ts} for Moirai \cite{MOIRAI}, (3) \href{https://github.com/thuml/Large-Time-Series-Model}{Large-Time-Series-Model} for Timer \cite{Timer}, (4) \href{https://github.com/KimMeen/Time-LLM}{Time-LLM} for Time-LLM \cite{Time-LLM}, (5) \href{https://github.com/emadeldeen24/TSLANet}{TSLANet} for TSLANet \cite{TSLANet}, (6) \href{https://github.com/panzijie825/S2IP-LLM}{S2IP-LLM} for $S^2$IP-LLM \cite{S2IP-LLM}, (7) \href{https://github.com/DAMO-DI-ML/NeurIPS2023-One-Fits-All}{NeurIPS2023-One-Fits-All} for GPT4TS \cite{GPT4TS}, (8) \href{https://github.com/mims-harvard/UniTS}{UniTS} for UniTS \cite{UniTS}, (9) \href{https://github.com/moment-timeseries-foundation-model/moment}{moment} for Moment \cite{Moment}, (10) \href{https://github.com/aikunyi/FilterNet}{FilterNet} for FilterNet \cite{FilterNet}, (11) \href{https://github.com/plumprc/RTSF}{RTSF} for RLinear \cite{RLinear}, and (12) \href{https://github.com/thuml/Time-Series-Library}{Time-Series-Library} for other models, such as TimesNet \cite{TimesNet}, PatchTST \cite{PatchTST}, TimeMixer \cite{TimeMixer}, iTransformer \cite{iTransformer}, DLinear \cite{DLinear}, Autoformer \cite{Autoformer} and Informer \cite{Informer}. To ensure a fair comparison, we use the original experimental configuration in the project scripts.

\subsection{Time Series Long-term Forecasting (Table \ref{table:long-term forecasting full results 1}, Table \ref{table:long-term forecasting full results 2} and Table \ref{table:long-term forecasting full results 3})}
\label{sec: long-term forecasting full results}

\subsection{Time Series Short-term Forecasting (Table \ref{table:short-term forecasting full results 1} and Table \ref{table:short-term forecasting full results 2})}
\label{sec: short-term forecasting full results}

\subsection{Time Series Classification (Table \ref{table:classification full results 1} and Table \ref{table:classification full results 2})}
\label{sec: classification full results}

\subsection{Time Series Imputation (Table \ref{table:imputation full results 1} and Table \ref{table:imputation full results 2})}
\label{sec: imputation full results}

\subsection{Time Series Anomaly Detection (Table \ref{table:anomaly detection full results})}
\label{sec: anomaly detection full results}

\newpage

% 时间序列长期预测表1
\begin{table}[ht]
\caption{Full results for the long-term forecasting task compared with Peri-midFormer \cite{Peri-midFormer}, Moirai \cite{MOIRAI}, Timer \cite{Timer}, Time-LLM \cite{Time-LLM}, TSLANet \cite{TSLANet}, $S^2$IP-LLM \cite{S2IP-LLM} and GPT4TS \cite{GPT4TS}. (* means former.) To ensure fairness in the comparison, we set the look-back window length of all models to \textbf{96}. Since the Timer and Moirai need to input a longer series to build a token, their windows are \textbf{672}. S2IP-LLM has a gradient explosion when the window is \textbf{96}, so its look-back window is \textbf{512}. \textbf{\textcolor{red}{Red}}: best, \textcolor{blue}{Blue}: second best.}
\centering
\vskip 0in
\begin{threeparttable}
\begin{small}
\setlength{\extrarowheight}{1.5pt}
\setlength{\tabcolsep}{3pt}
\begin{tabular}{c|c|cc|cc|cc|cc|cc|cc|cc|cc}
\toprule
\multicolumn{2}{c}{}  & \multicolumn{2}{c}{\textbf{SymTime}}   & \multicolumn{2}{c}{Peri-mid*} & \multicolumn{2}{c}{Moirai}   & \multicolumn{2}{c}{Timer}    & \multicolumn{2}{c}{Time-LLM}   & \multicolumn{2}{c}{TSLANet}   & \multicolumn{2}{c}{$S^2$IP-LLM}  & \multicolumn{2}{c}{GPT4TS}    \\

\multicolumn{2}{c}{\multirow{-2}{*}{Methods}} & \multicolumn{2}{c}{\textbf{Ours}}   & \multicolumn{2}{c}{\citeyearpar{Peri-midFormer}} & \multicolumn{2}{c}{\citeyearpar{MOIRAI}}   & \multicolumn{2}{c}{\citeyearpar{Timer}}    & \multicolumn{2}{c}{\citeyearpar{Time-LLM}}   & \multicolumn{2}{c}{\citeyearpar{TSLANet}}   & \multicolumn{2}{c}{\citeyearpar{S2IP-LLM}}  & \multicolumn{2}{c}{\citeyearpar{GPT4TS}}    \\
\cmidrule(lr){3-4} \cmidrule(lr){5-6} \cmidrule(lr){7-8} \cmidrule(lr){9-10} \cmidrule(lr){11-12} \cmidrule(lr){13-14} \cmidrule(lr){15-16} \cmidrule(lr){17-18}
\multicolumn{2}{c}{Metrics} & MSE   & MAE   & MSE & MAE & MSE   & MAE   & MSE   & MAE   & MSE   & MAE   & MSE   & MAE   & MSE   & MAE   & MSE   & MAE   \\

\midrule
   & 96  & 0.322    & 0.362    & 0.334   & 0.370   & 0.311    & 0.358    & 0.315 & 0.354    & 0.304    & 0.359    & 0.321    & 0.362    & 0.325    & 0.371 & 0.293    & 0.362 \\
   & 192 & 0.362    & 0.380    & 0.382   & 0.391   & 0.381    & 0.402    & 0.369 & 0.378    & 0.368    & 0.396    & 0.361    & 0.383    & 0.361    & 0.397 & 0.374    & 0.392 \\
   & 336 & 0.386    & 0.402    & 0.417   & 0.418   & 0.436    & 0.432    & 0.425 & 0.428    & 0.383    & 0.393    & 0.383    & 0.404    & 0.385    & 0.403 & 0.389    & 0.404 \\
   & 720 & 0.419    & 0.423    & 0.501   & 0.461   & 0.466    & 0.476    & 0.442 & 0.447    & 0.420    & 0.429    & 0.445    & 0.437    & 0.426    & 0.446 & 0.421    & 0.423 \\

\cmidrule(lr){2-18}
\multirow{-5}{*}{\rotatebox{90}{ETTm1}}  & Avg & 0.372    & {\color[HTML]{FF0000} \textbf{0.392}} & 0.409   & 0.410   & 0.398    & 0.417    & 0.388 & 0.402    & {\color[HTML]{FF0000} \textbf{0.369}} & {\color[HTML]{0000FF} 0.394}   & 0.377    & 0.397    & 0.374    & 0.404 & {\color[HTML]{0000FF} 0.369}   & 0.395 \\

\midrule
   & 96  & 0.176    & 0.260    & 0.174   & 0.255   & 0.179    & 0.267    & 0.168 & 0.254    & 0.177    & 0.269    & 0.179    & 0.261    & 0.174    & 0.263 & 0.171    & 0.265 \\
   & 192 & 0.244    & 0.306    & 0.249   & 0.305   & 0.244    & 0.311    & 0.429 & 0.425    & 0.239    & 0.305    & 0.243    & 0.303    & 0.232    & 0.306 & 0.226    & 0.304 \\
   & 336 & 0.306    & 0.343    & 0.319   & 0.349   & 0.335    & 0.371    & 0.476 & 0.457    & 0.301    & 0.340    & 0.308    & 0.345    & 0.300    & 0.344 & 0.288    & 0.345 \\
   & 720 & 0.405    & 0.401    & 0.418   & 0.405   & 0.425    & 0.444    & 0.545 & 0.497    & 0.382    & 0.381    & 0.403    & 0.401    & 0.359    & 0.386 & 0.372    & 0.398 \\

\cmidrule(lr){2-18}
\multirow{-5}{*}{\rotatebox{90}{ETTm2}}  & Avg & 0.283    & 0.328    & 0.290   & 0.328   & 0.296    & 0.348    & 0.405 & 0.408    & 0.275    & {\color[HTML]{FF0000} \textbf{0.324}} & 0.283    & 0.327    & {\color[HTML]{0000FF} 0.266}   & {\color[HTML]{0000FF} 0.325} & {\color[HTML]{FF0000} \textbf{0.264}} & 0.328 \\

\midrule
   & 96  & 0.376    & 0.400    & 0.382   & 0.403   & 0.369    & 0.408    & 0.374 & 0.404    & 0.386    & 0.395    & 0.387    & 0.405    & 0.380    & 0.403 & 0.388    & 0.399 \\
   & 192 & 0.428    & 0.431    & 0.436   & 0.435   & 0.441    & 0.450    & 0.430 & 0.438    & 0.421    & 0.424    & 0.448    & 0.436    & 0.410    & 0.427 & 0.425    & 0.429 \\
   & 336 & 0.463    & 0.456    & 0.492   & 0.455   & 0.469    & 0.469    & 0.458 & 0.453    & 0.438    & 0.450    & 0.451    & 0.437    & 0.426    & 0.442 & 0.444    & 0.455 \\
   & 720 & 0.450    & 0.458    & 0.508   & 0.490   & 0.486    & 0.490    & 0.475 & 0.480    & 0.506    & 0.510    & 0.505    & 0.485    & 0.610    & 0.543 & 0.479    & 0.477 \\

\cmidrule(lr){2-18}
\multirow{-5}{*}{\rotatebox{90}{ETTh1}}  & Avg & {\color[HTML]{FF0000} \textbf{0.430}} & {\color[HTML]{FF0000} \textbf{0.436}} & 0.455   & 0.446   & 0.441    & 0.454    & 0.434 & 0.444    & 0.438    & 0.445    & 0.448    & 0.441    & 0.456    & 0.454 & {\color[HTML]{0000FF} 0.434}   & {\color[HTML]{0000FF} 0.440} \\

\midrule
   & 96  & 0.293    & 0.347    & 0.312   & 0.358   & 0.288    & 0.350    & 0.315 & 0.360    & 0.307    & 0.369    & 0.289    & 0.345    & 0.292    & 0.353 & 0.292    & 0.351 \\
   & 192 & 0.376    & 0.397    & 0.388   & 0.403   & 0.390    & 0.426    & 0.411 & 0.423    & 0.349    & 0.384    & 0.362    & 0.391    & 0.355    & 0.388 & 0.351    & 0.394 \\
   & 336 & 0.410    & 0.431    & 0.443   & 0.443   & 0.441    & 0.435    & 0.465 & 0.467    & 0.394    & 0.420    & 0.350    & 0.389    & 0.368    & 0.417 & 0.380    & 0.421 \\
   & 720 & 0.423    & 0.445    & 0.455   & 0.459   & 0.487    & 0.435    & 0.521 & 0.515    & 0.426    & 0.454    & 0.418    & 0.439    & 0.434    & 0.460 & 0.424    & 0.446 \\

\cmidrule(lr){2-18}
\multirow{-5}{*}{\rotatebox{90}{ETTh2}}  & Avg & 0.375    & 0.405    & 0.400   & 0.416   & 0.402    & 0.411    & 0.428 & 0.441    & 0.369    & 0.407    & {\color[HTML]{FF0000} \textbf{0.355}} & {\color[HTML]{FF0000} \textbf{0.391}} & 0.362    & 0.405 & {\color[HTML]{0000FF} 0.359}   & {\color[HTML]{0000FF} 0.403} \\

\midrule
   & 96  & 0.166    & 0.213    & 0.157   & 0.201   & 0.156    & 0.206    & 0.289 & 0.331    & 0.172    & 0.221    & 0.177    & 0.216    & 0.162    & 0.213 & 0.184    & 0.224 \\
   & 192 & 0.212    & 0.254    & 0.244   & 0.273   & 0.229    & 0.274    & 0.314 & 0.349    & 0.194    & 0.241    & 0.226    & 0.258    & 0.197    & 0.246 & 0.230    & 0.263 \\
   & 336 & 0.267    & 0.294    & 0.283   & 0.303   & 0.282    & 0.316    & 0.339 & 0.363    & 0.286    & 0.282    & 0.279    & 0.588    & 0.281    & 0.299 & 0.285    & 0.302 \\
   & 720 & 0.342    & 0.344    & 0.364   & 0.355   & 0.395    & 0.401    & 0.375 & 0.388    & 0.337    & 0.332    & 0.355    & 0.346    & 0.333    & 0.339 & 0.362    & 0.352 \\

\cmidrule(lr){2-18}
\multirow{-5}{*}{\rotatebox{90}{Weather}}   & Avg & {\color[HTML]{0000FF} 0.247}   & 0.276    & 0.262   & 0.283   & 0.265    & 0.299    & 0.329 & 0.358    & 0.247    & {\color[HTML]{FF0000} \textbf{0.269}} & 0.259    & 0.352    & {\color[HTML]{FF0000} \textbf{0.243}} & {\color[HTML]{0000FF} 0.274} & 0.265    & 0.285 \\

\midrule
   & 96  & 0.162    & 0.253    & 0.151   & 0.245   & 0.137    & 0.221    & 0.150 & 0.244    & 0.149    & 0.242    & 0.176    & 0.261    & 0.149    & 0.251 & 0.186    & 0.272 \\
   & 192 & 0.173    & 0.264    & 0.168   & 0.259   & 0.158    & 0.243    & 0.159 & 0.252    & 0.167    & 0.261    & 0.182    & 0.268    & 0.171    & 0.269 & 0.190    & 0.277 \\
   & 336 & 0.194    & 0.285    & 0.184   & 0.268   & 0.167    & 0.255    & 0.190 & 0.271    & 0.188    & 0.270    & 0.199    & 0.285    & 0.199    & 0.291 & 0.205    & 0.292 \\
   & 720 & 0.220    & 0.304    & 0.207   & 0.297   & 0.207    & 0.290    & 0.210 & 0.300    & 0.214    & 0.301    & 0.240    & 0.317    & 0.244    & 0.319 & 0.245    & 0.323 \\

\cmidrule(lr){2-18}
\multirow{-5}{*}{\rotatebox{90}{ECL}} & Avg & 0.187    & 0.276    & 0.178   & 0.267   & {\color[HTML]{FF0000} \textbf{0.167}} & {\color[HTML]{FF0000} \textbf{0.252}} & {\color[HTML]{0000FF} 0.177} & {\color[HTML]{0000FF} 0.267}   & 0.180    & 0.269    & 0.199    & 0.283    & 0.191    & 0.283 & 0.206    & 0.291 \\

\midrule
   & 96  & 0.432    & 0.280    & 0.426   & 0.277   & 0.376    & 0.264    & 0.391 & 0.260    & 0.389    & 0.286    & 0.398    & 0.291    & 0.385    & 0.289 & 0.471    & 0.312 \\
   & 192 & 0.444    & 0.287    & 0.440   & 0.283   & 0.410    & 0.279    & 0.426 & 0.271    & 0.405    & 0.306    & 0.430    & 0.307    & 0.403    & 0.308 & 0.478    & 0.312 \\
   & 336 & 0.458    & 0.293    & 0.477   & 0.311   & 0.442    & 0.287    & 0.451 & 0.297    & 0.424    & 0.299    & 0.494    & 0.312    & 0.425    & 0.299 & 0.493    & 0.319 \\
   & 720 & 0.492    & 0.303    & 0.487   & 0.308   & 0.470    & 0.328    & 0.475 & 0.307    & 0.453    & 0.334    & 0.528    & 0.332    & 0.454    & 0.326 & 0.523    & 0.335 \\

\cmidrule(lr){2-18}
\multirow{-5}{*}{\rotatebox{90}{Traffic}}   & Avg & 0.457    & 0.291    & 0.458   & 0.295   & 0.424    & {\color[HTML]{0000FF} 0.289}   & 0.436 & {\color[HTML]{FF0000} \textbf{0.284}} & {\color[HTML]{0000FF} 0.418}   & 0.306    & 0.463    & 0.310    & {\color[HTML]{FF0000} \textbf{0.417}} & 0.306 & 0.491    & 0.320 \\

\midrule
   & 96  & 0.084    & 0.201    & 0.083   & 0.199   & 0.089    & 0.211    & 0.098 & 0.228    & 0.090    & 0.209    & 0.082    & 0.200    & 0.147    & 0.279 & 0.087    & 0.218 \\
   & 192 & 0.174    & 0.295    & 0.190   & 0.307   & 0.175    & 0.289    & 0.196 & 0.325    & 0.188    & 0.310    & 0.172    & 0.295    & 0.234    & 0.354 & 0.171    & 0.294 \\
   & 336 & 0.331    & 0.416    & 0.401   & 0.458   & 0.345    & 0.423    & 0.359 & 0.433    & 0.342    & 0.427    & 0.329    & 0.415    & 0.403    & 0.474 & 0.349    & 0.418 \\
   & 720 & 0.847    & 0.694    & 0.879   & 0.702   & 0.882    & 0.744    & 0.875 & 0.713    & 0.885    & 0.707    & 0.889    & 0.747    & 1.103    & 0.804 & 0.873    & 0.713 \\

\cmidrule(lr){2-18}
\multirow{-5}{*}{\rotatebox{90}{Exchange}}   & Avg & {\color[HTML]{FF0000} \textbf{0.359}} & {\color[HTML]{FF0000} \textbf{0.401}} & 0.388   & 0.417   & 0.373    & 0.417    & 0.382 & 0.425    & 0.376    & 0.414    & {\color[HTML]{0000FF} 0.368}   & 0.414    & 0.472    & 0.478 & 0.370    & {\color[HTML]{0000FF} 0.411} \\

\midrule
\multicolumn{2}{c}{Average} & {\color[HTML]{0000FF} 0.339}   & {\color[HTML]{FF0000} \textbf{0.351}} & 0.355   & 0.358   & 0.346    & 0.361    & 0.372 & 0.378    & {\color[HTML]{FF0000} \textbf{0.334}} & {\color[HTML]{0000FF} 0.353}   & 0.344    & 0.364    & 0.348    & 0.366 & 0.345    & 0.359 \\
\bottomrule
\end{tabular}
\end{small}
\end{threeparttable}
\label{table:long-term forecasting full results 1}
\end{table}

% 时间序列长期预测表2
\begin{table}[ht]
\caption{Full results for the long-term forecasting task compared with FilterNet \cite{FilterNet}, TimesNet \cite{TimesNet}, iTransformer \cite{iTransformer}, PatchTST \cite{PatchTST}, RLinear \cite{RLinear}, DLinear \cite{DLinear} and TimeMixer \cite{TimeMixer}. To ensure fairness in the comparison, we set the look-back window length of all models to \textbf{96}. \textbf{\textcolor{red}{Red}}: best, \textcolor{blue}{Blue}: second best.}
\centering
\vskip 0.10in
\begin{threeparttable}
\begin{small}
\setlength{\extrarowheight}{1.5pt}
\setlength{\tabcolsep}{3pt}
\begin{tabular}{c|c|cc|cc|cc|cc|cc|cc|cc|cc}
\toprule
\multicolumn{2}{c}{}  & \multicolumn{2}{c}{\textbf{SymTime}}    & \multicolumn{2}{c}{FilterNet}   & \multicolumn{2}{c}{TimesNet} & \multicolumn{2}{c}{iTransformer}   & \multicolumn{2}{c}{PatchTST}   & \multicolumn{2}{c}{RLinear} & \multicolumn{2}{c}{DLinear}   & \multicolumn{2}{c}{TimeMixer}   \\

\multicolumn{2}{c}{\multirow{-2}{*}{Methods}} & \multicolumn{2}{c}{\textbf{(Ours)}}    & \multicolumn{2}{c}{\citeyearpar{FilterNet}}   & \multicolumn{2}{c}{\citeyearpar{TimesNet}} & \multicolumn{2}{c}{\citeyearpar{iTransformer}}   & \multicolumn{2}{c}{\citeyearpar{PatchTST}}   & \multicolumn{2}{c}{\citeyearpar{RLinear}} & \multicolumn{2}{c}{\citeyearpar{DLinear}}   & \multicolumn{2}{c}{\citeyearpar{TimeMixer}}   \\
\cmidrule(lr){3-4} \cmidrule(lr){5-6} \cmidrule(lr){7-8} \cmidrule(lr){9-10} \cmidrule(lr){11-12} \cmidrule(lr){13-14} \cmidrule(lr){15-16} \cmidrule(lr){17-18}
\multicolumn{2}{c}{Metrics} & MSE   & MAE   & MSE   & MAE   & MSE  & MAE   & MSE   & MAE   & MSE  & MAE  & MSE  & MAE   & MSE  & MAE  & MSE   & MAE   \\

\midrule
    & 96  & 0.322    & 0.362    & 0.321    & 0.361    & 0.331   & 0.372  & 0.343    & 0.377    & 0.324 & 0.365 & 0.355 & 0.376    & 0.345 & 0.372 & 0.323    & 0.361    \\
    & 192   & 0.362    & 0.380    & 0.367    & 0.387    & 0.397   & 0.402  & 0.381    & 0.395    & 0.367 & 0.389 & 0.387 & 0.392    & 0.382 & 0.391 & 0.362    & 0.383    \\
    & 336   & 0.386    & 0.402    & 0.401    & 0.409    & 0.427   & 0.427  & 0.419    & 0.418    & 0.400 & 0.409 & 0.424 & 0.415    & 0.414 & 0.414 & 0.388    & 0.403    \\
    & 720   & 0.419    & 0.423    & 0.477    & 0.448    & 0.493   & 0.463  & 0.487    & 0.457    & 0.460 & 0.445 & 0.487 & 0.450    & 0.473 & 0.450 & 0.454    & 0.442    \\

\cmidrule(lr){2-18}
\multirow{-5}{*}{\rotatebox{90}{ETTm1}}   & Avg  & {\color[HTML]{FF0000} \textbf{0.372}} & {\color[HTML]{FF0000} \textbf{0.392}} & 0.392    & 0.401    & 0.412   & 0.416  & 0.407    & 0.412    & 0.388 & 0.402 & 0.413 & 0.408    & 0.403 & 0.407 & {\color[HTML]{0000FF} 0.382}   & {\color[HTML]{0000FF} 0.397}   \\

\midrule
    & 96  & 0.176    & 0.260    & 0.175    & 0.258    & 0.185   & 0.265  & 0.185    & 0.271    & 0.182 & 0.266 & 0.182 & 0.265    & 0.194 & 0.293 & 0.177    & 0.259    \\
    & 192   & 0.244    & 0.306    & 0.240    & 0.301    & 0.256   & 0.310  & 0.254    & 0.314    & 0.250 & 0.311 & 0.246 & 0.304    & 0.283 & 0.360 & 0.245    & 0.306    \\
    & 336   & 0.306    & 0.343    & 0.311    & 0.347    & 0.314   & 0.345  & 0.315    & 0.352    & 0.313 & 0.350 & 0.307 & 0.342    & 0.376 & 0.423 & 0.298    & 0.338    \\
    & 720   & 0.405    & 0.401    & 0.414    & 0.405    & 0.424   & 0.412  & 0.413    & 0.407    & 0.417 & 0.412 & 0.407 & 0.398    & 0.529 & 0.509 & 0.395    & 0.396    \\

\cmidrule(lr){2-18}
\multirow{-5}{*}{\rotatebox{90}{ETTm2}}   & Avg  & {\color[HTML]{0000FF} 0.283}   & 0.328    & 0.285    & 0.328    & 0.295   & 0.333  & 0.292    & 0.336    & 0.291 & 0.335 & 0.286 & {\color[HTML]{0000FF} 0.327}   & 0.346 & 0.396 & {\color[HTML]{FF0000} \textbf{0.279}} & {\color[HTML]{FF0000} \textbf{0.325}} \\

\midrule
    & 96  & 0.376    & 0.400    & 0.382    & 0.402    & 0.409   & 0.425  & 0.394    & 0.409    & 0.381 & 0.400 & 0.386 & 0.395    & 0.396 & 0.411 & 0.385    & 0.400    \\
    & 192   & 0.428    & 0.431    & 0.430    & 0.429    & 0.469   & 0.460  & 0.447    & 0.440    & 0.429 & 0.433 & 0.437 & 0.424    & 0.446 & 0.441 & 0.441    & 0.431    \\
    & 336   & 0.463    & 0.456    & 0.472    & 0.451    & 0.507   & 0.478  & 0.491    & 0.464    & 0.475 & 0.460 & 0.479 & 0.446    & 0.490 & 0.468 & 0.482    & 0.450    \\
    & 720   & 0.450    & 0.458    & 0.481    & 0.473    & 0.521   & 0.497  & 0.517    & 0.501    & 0.517 & 0.502 & 0.481 & 0.470    & 0.514 & 0.511 & 0.504    & 0.482    \\

\cmidrule(lr){2-18}
\multirow{-5}{*}{\rotatebox{90}{ETTh1}}   & Avg  & {\color[HTML]{FF0000} \textbf{0.430}} & {\color[HTML]{0000FF} 0.436}   & {\color[HTML]{0000FF} 0.441}   & 0.439    & 0.476   & 0.465  & 0.462    & 0.454    & 0.451 & 0.449 & 0.446 & {\color[HTML]{FF0000} \textbf{0.434}} & 0.461 & 0.458 & 0.453    & 0.441    \\

\midrule
    & 96  & 0.293    & 0.347    & 0.293    & 0.343    & 0.331   & 0.372  & 0.300    & 0.350    & 0.301 & 0.351 & 0.318 & 0.363    & 0.348 & 0.401 & 0.293    & 0.343    \\
    & 192   & 0.376    & 0.397    & 0.374    & 0.396    & 0.429   & 0.423  & 0.380    & 0.399    & 0.374 & 0.398 & 0.401 & 0.412    & 0.473 & 0.474 & 0.376    & 0.396    \\
    & 336   & 0.410    & 0.431    & 0.417    & 0.430    & 0.450   & 0.451  & 0.422    & 0.432    & 0.429 & 0.439 & 0.436 & 0.442    & 0.588 & 0.539 & 0.425    & 0.432    \\
    & 720   & 0.423    & 0.445    & 0.449    & 0.460    & 0.459   & 0.466  & 0.429    & 0.447    & 0.443 & 0.461 & 0.442 & 0.454    & 0.829 & 0.656 & 0.457    & 0.459    \\

\cmidrule(lr){2-18}
\multirow{-5}{*}{\rotatebox{90}{ETTh2}}   & Avg  & {\color[HTML]{FF0000} \textbf{0.375}} & {\color[HTML]{FF0000} \textbf{0.405}} & 0.383    & 0.407    & 0.417   & 0.428  & {\color[HTML]{0000FF} 0.383}   & {\color[HTML]{0000FF} 0.407}   & 0.387 & 0.412 & 0.399 & 0.418    & 0.559 & 0.518 & 0.388    & 0.408    \\

\midrule
    & 96  & 0.166    & 0.213    & 0.162    & 0.207    & 0.171   & 0.222  & 0.176    & 0.215    & 0.177 & 0.219 & 0.192 & 0.232    & 0.197 & 0.258 & 0.172    & 0.220    \\
    & 192   & 0.212    & 0.254    & 0.210    & 0.250    & 0.234   & 0.273  & 0.226    & 0.258    & 0.222 & 0.258 & 0.240 & 0.271    & 0.237 & 0.296 & 0.227    & 0.259    \\
    & 336   & 0.267    & 0.294    & 0.265    & 0.290    & 0.284   & 0.306  & 0.281    & 0.299    & 0.281 & 0.299 & 0.292 & 0.307    & 0.282 & 0.332 & 0.266    & 0.294    \\
    & 720   & 0.342    & 0.344    & 0.342    & 0.340    & 0.358   & 0.352  & 0.359    & 0.350    & 0.356 & 0.348 & 0.364 & 0.353    & 0.347 & 0.385 & 0.346    & 0.347    \\

\cmidrule(lr){2-18}
\multirow{-5}{*}{\rotatebox{90}{Weather}}  & Avg  & {\color[HTML]{0000FF} 0.247}   & {\color[HTML]{0000FF} 0.276}   & {\color[HTML]{FF0000} \textbf{0.245}} & {\color[HTML]{FF0000} \textbf{0.272}} & 0.262   & 0.288  & 0.260    & 0.281    & 0.259 & 0.281 & 0.272 & 0.291    & 0.266 & 0.318 & 0.253    & 0.280    \\

\midrule
    & 96  & 0.162    & 0.253    & 0.147    & 0.245    & 0.167   & 0.271  & 0.148    & 0.240    & 0.180 & 0.272 & 0.201 & 0.281    & 0.210 & 0.302 & 0.157    & 0.249    \\
    & 192   & 0.173    & 0.264    & 0.160    & 0.250    & 0.186   & 0.288  & 0.165    & 0.256    & 0.188 & 0.279 & 0.201 & 0.283    & 0.210 & 0.305 & 0.170    & 0.261    \\
    & 336   & 0.194    & 0.285    & 0.173    & 0.267    & 0.203   & 0.304  & 0.179    & 0.271    & 0.204 & 0.296 & 0.215 & 0.298    & 0.223 & 0.319 & 0.186    & 0.276    \\
    & 720   & 0.220    & 0.304    & 0.210    & 0.309    & 0.227   & 0.322  & 0.209    & 0.298    & 0.246 & 0.328 & 0.257 & 0.331    & 0.258 & 0.350 & 0.227    & 0.311    \\

\cmidrule(lr){2-18}
\multirow{-5}{*}{\rotatebox{90}{ECL}}  & Avg  & 0.187    & 0.276    & {\color[HTML]{FF0000} \textbf{0.173}} & {\color[HTML]{0000FF} 0.268}   & 0.196   & 0.296  & {\color[HTML]{0000FF} 0.175}   & {\color[HTML]{FF0000} \textbf{0.267}} & 0.204 & 0.294 & 0.219 & 0.298    & 0.225 & 0.319 & 0.185    & 0.274    \\

\midrule
    & 96  & 0.432    & 0.280    & 0.430    & 0.294    & 0.589   & 0.316  & 0.393    & 0.268    & 0.461 & 0.298 & 0.649 & 0.389    & 0.696 & 0.429 & 0.479    & 0.299    \\
    & 192   & 0.444    & 0.287    & 0.452    & 0.307    & 0.616   & 0.328  & 0.413    & 0.277    & 0.467 & 0.301 & 0.601 & 0.366    & 0.647 & 0.407 & 0.490    & 0.303    \\
    & 336   & 0.458    & 0.293    & 0.470    & 0.316    & 0.628   & 0.333  & 0.424    & 0.283    & 0.483 & 0.308 & 0.609 & 0.369    & 0.653 & 0.410 & 0.493    & 0.304    \\
    & 720   & 0.492    & 0.303    & 0.498    & 0.323    & 0.667   & 0.352  & 0.458    & 0.300    & 0.517 & 0.325 & 0.647 & 0.387    & 0.695 & 0.429 & 0.534    & 0.319    \\

\cmidrule(lr){2-18}
\multirow{-5}{*}{\rotatebox{90}{Traffic}}  & Avg  & {\color[HTML]{0000FF} 0.457}   & {\color[HTML]{0000FF} 0.291}   & 0.463    & 0.310    & 0.625   & 0.332  & {\color[HTML]{FF0000} \textbf{0.422}} & {\color[HTML]{FF0000} \textbf{0.282}} & 0.482 & 0.308 & 0.627 & 0.378    & 0.673 & 0.419 & 0.499    & 0.306    \\

\midrule
    & 96  & 0.084    & 0.201    & 0.091    & 0.211    & 0.115   & 0.246  & 0.094    & 0.216    & 0.088 & 0.205 & 0.093 & 0.217    & 0.093 & 0.226 & 0.091    & 0.210    \\
    & 192   & 0.174    & 0.295    & 0.186    & 0.305    & 0.213   & 0.335  & 0.185    & 0.307    & 0.189 & 0.309 & 0.184 & 0.307    & 0.184 & 0.324 & 0.185    & 0.304    \\
    & 336   & 0.331    & 0.416    & 0.380    & 0.449    & 0.367   & 0.440  & 0.336    & 0.422    & 0.327 & 0.415 & 0.351 & 0.432    & 0.328 & 0.436 & 0.361    & 0.435    \\
    & 720   & 0.847    & 0.694    & 0.896    & 0.712    & 0.978   & 0.753  & 0.893    & 0.716    & 0.886 & 0.706 & 0.886 & 0.714    & 0.880 & 0.705 & 0.974    & 0.741    \\

\cmidrule(lr){2-18}
\multirow{-5}{*}{\rotatebox{90}{Exchange}}  & Avg  & {\color[HTML]{FF0000} \textbf{0.359}} & {\color[HTML]{FF0000} \textbf{0.401}} & 0.388    & 0.419    & 0.418   & 0.443  & 0.377    & 0.415    & 0.373 & {\color[HTML]{0000FF} 0.409} & 0.379 & 0.418    & {\color[HTML]{0000FF} 0.371} & 0.423 & 0.403    & 0.423    \\

\midrule
\multicolumn{2}{c}{Average} & {\color[HTML]{FF0000} \textbf{0.339}} & {\color[HTML]{FF0000} \textbf{0.351}} & {\color[HTML]{0000FF} 0.346}   & {\color[HTML]{0000FF} 0.356}   & 0.388   & 0.375  & 0.347    & 0.357    & 0.354 & 0.361 & 0.380 & 0.371    & 0.413 & 0.407 & 0.355    & 0.357 \\
\bottomrule
\end{tabular}
\end{small}
\end{threeparttable}
\label{table:long-term forecasting full results 2}
\end{table}

% 时间序列长期预测表3
\begin{table}[ht]
\caption{Full results for the long-term forecasting task compared with Autoformer \cite{Autoformer}, Crossformer \cite{Crossformer}, FEDformer \cite{FEDformer}, ETSforemr \cite{ETSformer}, Stationary \cite{Non-stationary-transformers}, LightTS \cite{LightTS}, Informer \cite{Informer}. (Stationary means Nonstationary Transformer.) To ensure fairness in the comparison, we set the look-back window length of all models to \textbf{96}. \textbf{\textcolor{red}{Red}}: best, \textcolor{blue}{Blue}: second best.}
\centering
\vskip 0.10in
\begin{threeparttable}
\begin{small}
\setlength{\extrarowheight}{1.5pt}
\setlength{\tabcolsep}{3pt}
\begin{tabular}{c|c|cc|cc|cc|cc|cc|cc|cc|cc}
\toprule
\multicolumn{2}{c}{}  & \multicolumn{2}{c}{\textbf{SymTime}}    & \multicolumn{2}{c}{Autoformer} & \multicolumn{2}{c}{Crossformer}  & \multicolumn{2}{c}{FEDformer}   & \multicolumn{2}{c}{ETSformer}   & \multicolumn{2}{c}{Stationary} & \multicolumn{2}{c}{LightTS}   & \multicolumn{2}{c}{Informer} \\
\multicolumn{2}{c}{\multirow{-2}{*}{Methods}} & \multicolumn{2}{c}{\textbf{(Ours)}}    & \multicolumn{2}{c}{\citeyearpar{Autoformer}} & \multicolumn{2}{c}{\citeyearpar{Crossformer}}  & \multicolumn{2}{c}{\citeyearpar{FEDformer}}   & \multicolumn{2}{c}{\citeyearpar{ETSformer}}   & \multicolumn{2}{c}{\citeyearpar{Non-stationary-transformers}} & \multicolumn{2}{c}{\citeyearpar{LightTS}}   & \multicolumn{2}{c}{\citeyearpar{Informer}} \\
\cmidrule(lr){3-4} \cmidrule(lr){5-6} \cmidrule(lr){7-8} \cmidrule(lr){9-10} \cmidrule(lr){11-12} \cmidrule(lr){13-14} \cmidrule(lr){15-16} \cmidrule(lr){17-18}

\multicolumn{2}{c}{Metrics} & MSE   & MAE   & MSE  & MAE  & MSE  & MAE  & MSE  & MAE  & MSE  & MAE  & MSE  & MAE  & MSE  & MAE  & MSE  & MAE   \\

\midrule
   & 96  & 0.322    & 0.362    & 0.501   & 0.479   & 0.360 & 0.399 & 0.378 & 0.418 & 0.375 & 0.398 & 0.418   & 0.415   & 0.390 & 0.411 & 0.619   & 0.549  \\
   & 192 & 0.362    & 0.380    & 0.578   & 0.510   & 0.422 & 0.449 & 0.438 & 0.449 & 0.408 & 0.410 & 0.506   & 0.454   & 0.425 & 0.436 & 0.760   & 0.645  \\
   & 336 & 0.386    & 0.402    & 0.668   & 0.552   & 0.589 & 0.557 & 0.456 & 0.462 & 0.435 & 0.428 & 0.530   & 0.482   & 0.463 & 0.464 & 1.093   & 0.812  \\
   & 720 & 0.419    & 0.423    & 0.602   & 0.524   & 0.838 & 0.706 & 0.530 & 0.498 & 0.499 & 0.462 & 0.610   & 0.525   & 0.547 & 0.520 & 1.114   & 0.806  \\

\cmidrule(lr){2-18}
\multirow{-5}{*}{\rotatebox{90}{ETTm1}}  & Avg & {\color[HTML]{FF0000} \textbf{0.372}} & {\color[HTML]{FF0000} \textbf{0.392}} & 0.587   & 0.516   & 0.552 & 0.528 & 0.450 & 0.457 & {\color[HTML]{0000FF} 0.429} & {\color[HTML]{0000FF} 0.425} & 0.516   & 0.469   & 0.456 & 0.458 & 0.896   & 0.703  \\

\midrule
   & 96  & 0.176    & 0.260    & 0.245   & 0.323   & 0.274 & 0.268 & 0.196 & 0.284 & 0.189 & 0.280 & 0.240   & 0.308   & 0.226 & 0.323 & 0.467   & 0.533  \\
   & 192 & 0.244    & 0.306    & 0.289   & 0.345   & 0.366 & 0.380 & 0.264 & 0.325 & 0.275 & 0.319 & 0.428   & 0.402   & 0.361 & 0.421 & 0.742   & 0.664  \\
   & 336 & 0.306    & 0.343    & 0.342   & 0.378   & 0.437 & 0.453 & 0.324 & 0.363 & 0.314 & 0.357 & 0.521   & 0.449   & 0.474 & 0.488 & 1.184   & 0.825  \\
   & 720 & 0.405    & 0.401    & 0.441   & 0.429   & 0.506 & 0.623 & 0.434 & 0.428 & 0.414 & 0.413 & 0.602   & 0.501   & 0.760 & 0.631 & 4.039   & 1.530  \\

\cmidrule(lr){2-18}
\multirow{-5}{*}{\rotatebox{90}{ETTm2}}  & Avg & {\color[HTML]{FF0000} \textbf{0.283}} & {\color[HTML]{FF0000} \textbf{0.328}} & 0.329   & 0.369   & 0.396 & 0.431 & 0.305 & 0.350 & {\color[HTML]{0000FF} 0.298} & {\color[HTML]{0000FF} 0.342} & 0.448   & 0.415   & 0.455 & 0.466 & 1.608   & 0.888  \\

\midrule
   & 96  & 0.376    & 0.400    & 0.453   & 0.459   & 0.462 & 0.473 & 0.376 & 0.417 & 0.494 & 0.479 & 0.550   & 0.503   & 0.448 & 0.450 & 0.926   & 0.741  \\
   & 192 & 0.428    & 0.431    & 0.481   & 0.470   & 0.495 & 0.484 & 0.431 & 0.454 & 0.538 & 0.504 & 0.655   & 0.569   & 0.503 & 0.483 & 0.968   & 0.757  \\
   & 336 & 0.463    & 0.456    & 0.519   & 0.495   & 0.693 & 0.626 & 0.461 & 0.469 & 0.574 & 0.521 & 0.791   & 0.639   & 0.554 & 0.513 & 1.144   & 0.849  \\
   & 720 & 0.450    & 0.458    & 0.510   & 0.508   & 0.668 & 0.599 & 0.502 & 0.499 & 0.562 & 0.535 & 0.797   & 0.652   & 0.627 & 0.578 & 1.214   & 0.880  \\

\cmidrule(lr){2-18}
\multirow{-5}{*}{\rotatebox{90}{ETTh1}}  & Avg & {\color[HTML]{FF0000} \textbf{0.430}} & {\color[HTML]{FF0000} \textbf{0.436}} & 0.491   & 0.483   & 0.580 & 0.545 & {\color[HTML]{0000FF} 0.442} & {\color[HTML]{0000FF} 0.460} & 0.542 & 0.510 & 0.698   & 0.591   & 0.533 & 0.506 & 1.063   & 0.807  \\

\midrule
   & 96  & 0.293    & 0.347    & 0.383   & 0.416   & 0.367 & 0.347 & 0.346 & 0.390 & 0.340 & 0.391 & 0.417   & 0.432   & 0.417 & 0.448 & 3.132   & 1.425  \\
   & 192 & 0.376    & 0.397    & 0.479   & 0.467   & 0.450 & 0.459 & 0.428 & 0.439 & 0.430 & 0.439 & 0.529   & 0.486   & 0.546 & 0.520 & 5.552   & 1.957  \\
   & 336 & 0.410    & 0.431    & 0.476   & 0.481   & 0.532 & 0.521 & 0.469 & 0.474 & 0.485 & 0.479 & 0.591   & 0.517   & 0.619 & 0.554 & 4.926   & 1.873  \\
   & 720 & 0.423    & 0.445    & 0.494   & 0.503   & 0.614 & 0.633 & 0.473 & 0.486 & 0.500 & 0.497 & 0.601   & 0.531   & 0.972 & 0.704 & 4.201   & 1.741  \\

\cmidrule(lr){2-18}
\multirow{-5}{*}{\rotatebox{90}{ETTh2}}  & Avg & {\color[HTML]{FF0000} \textbf{0.375}} & {\color[HTML]{FF0000} \textbf{0.405}} & 0.458   & 0.467   & 0.491 & 0.490 & {\color[HTML]{0000FF} 0.429} & {\color[HTML]{0000FF} 0.447} & 0.439 & 0.452 & 0.534   & 0.491   & 0.639 & 0.556 & 4.453   & 1.749  \\

\midrule
   & 96  & 0.166    & 0.213    & 0.276   & 0.343   & 0.174 & 0.243 & 0.218 & 0.299 & 0.197 & 0.281 & 0.184   & 0.233   & 0.174 & 0.235 & 0.357   & 0.415  \\
   & 192 & 0.212    & 0.254    & 0.305   & 0.361   & 0.235 & 0.307 & 0.281 & 0.344 & 0.237 & 0.312 & 0.248   & 0.286   & 0.218 & 0.276 & 0.458   & 0.456  \\
   & 336 & 0.267    & 0.294    & 0.372   & 0.405   & 0.277 & 0.342 & 0.337 & 0.375 & 0.298 & 0.353 & 0.337   & 0.349   & 0.267 & 0.316 & 0.520   & 0.501  \\
   & 720 & 0.342    & 0.344    & 0.430   & 0.437   & 0.369 & 0.407 & 0.423 & 0.429 & 0.352 & 0.288 & 0.399   & 0.385   & 0.353 & 0.366 & 0.926   & 0.705  \\

\cmidrule(lr){2-18}
\multirow{-5}{*}{\rotatebox{90}{Weather}}   & Avg & {\color[HTML]{FF0000} \textbf{0.247}} & {\color[HTML]{FF0000} \textbf{0.276}} & 0.346   & 0.387   & 0.264 & 0.325 & 0.315 & 0.362 & {\color[HTML]{0000FF} 0.271} & 0.309 & 0.292   & 0.313   & 0.253 & {\color[HTML]{0000FF} 0.298} & 0.565   & 0.519  \\

\midrule
   & 96  & 0.162    & 0.253    & 0.198   & 0.313   & 0.146 & 0.249 & 0.202 & 0.314 & 0.187 & 0.304 & 0.167   & 0.270   & 0.211 & 0.313 & 0.342   & 0.423  \\
   & 192 & 0.173    & 0.264    & 0.218   & 0.329   & 0.163 & 0.262 & 0.211 & 0.323 & 0.199 & 0.315 & 0.183   & 0.284   & 0.223 & 0.326 & 0.360   & 0.442  \\
   & 336 & 0.194    & 0.285    & 0.253   & 0.352   & 0.198 & 0.296 & 0.222 & 0.335 & 0.212 & 0.329 & 0.194   & 0.295   & 0.243 & 0.346 & 0.365   & 0.445  \\
   & 720 & 0.220    & 0.304    & 0.265   & 0.367   & 0.245 & 0.346 & 0.272 & 0.373 & 0.233 & 0.345 & 0.224   & 0.321   & 0.277 & 0.371 & 0.412   & 0.469  \\

\cmidrule(lr){2-18}
\multirow{-5}{*}{\rotatebox{90}{ECL}} & Avg & {\color[HTML]{FF0000} \textbf{0.187}} & {\color[HTML]{FF0000} \textbf{0.276}} & 0.233   & 0.340   & {\color[HTML]{0000FF} 0.188} & {\color[HTML]{0000FF} 0.288} & 0.227 & 0.337 & 0.208 & 0.323 & 0.192   & 0.292   & 0.239 & 0.339 & 0.370   & 0.445  \\

\midrule
   & 96  & 0.432    & 0.280    & 0.608   & 0.383   & 0.516 & 0.268 & 0.592 & 0.372 & 0.607 & 0.392 & 0.621   & 0.347   & 0.667 & 0.419 & 0.720   & 0.407  \\
   & 192 & 0.444    & 0.287    & 0.630   & 0.397   & 0.541 & 0.283 & 0.598 & 0.371 & 0.621 & 0.399 & 0.643   & 0.355   & 0.662 & 0.425 & 0.738   & 0.414  \\
   & 336 & 0.458    & 0.293    & 0.622   & 0.387   & 0.566 & 0.351 & 0.636 & 0.397 & 0.622 & 0.396 & 0.650   & 0.360   & 0.683 & 0.436 & 0.833   & 0.470  \\
   & 720 & 0.492    & 0.303    & 0.689   & 0.396   & 0.610 & 0.403 & 0.639 & 0.395 & 0.632 & 0.396 & 0.670   & 0.365   & 0.700 & 0.455 & 0.854   & 0.491  \\

\cmidrule(lr){2-18}
\multirow{-5}{*}{\rotatebox{90}{Traffic}}   & Avg & {\color[HTML]{FF0000} \textbf{0.457}} & {\color[HTML]{FF0000} \textbf{0.291}} & 0.637   & 0.391   & {\color[HTML]{0000FF} 0.558} & {\color[HTML]{0000FF} 0.326} & 0.616 & 0.384 & 0.621 & 0.396 & 0.646   & 0.357   & 0.678 & 0.434 & 0.786   & 0.445  \\

\midrule
   & 96  & 0.084    & 0.201    & 0.191   & 0.318   & 0.276 & 0.383 & 0.162 & 0.291 & 0.085 & 0.204 & 0.132   & 0.254   & 0.128 & 0.266 & 0.896   & 0.761  \\
   & 192 & 0.174    & 0.295    & 0.315   & 0.407   & 0.540 & 0.552 & 0.276 & 0.382 & 0.182 & 0.303 & 0.251   & 0.361   & 0.292 & 0.402 & 1.146   & 0.861  \\
   & 336 & 0.331    & 0.416    & 0.480   & 0.519   & 1.229 & 0.873 & 0.442 & 0.488 & 0.348 & 0.428 & 0.467   & 0.507   & 0.500 & 0.536 & 1.628   & 1.017  \\
   & 720 & 0.847    & 0.694    & 1.255   & 0.868   & 1.721 & 1.055 & 1.175 & 0.833 & 1.025 & 0.774 & 1.304   & 0.837   & 1.002 & 0.763 & 2.552   & 1.299  \\

\cmidrule(lr){2-18}
\multirow{-5}{*}{\rotatebox{90}{Exchange}}   & Avg & {\color[HTML]{FF0000} \textbf{0.359}} & {\color[HTML]{FF0000} \textbf{0.401}} & 0.560   & 0.528   & 0.942 & 0.716 & 0.514 & 0.498 & {\color[HTML]{0000FF} 0.410} & {\color[HTML]{0000FF} 0.427} & 0.538   & 0.490   & 0.480 & 0.492 & 1.555   & 0.984  \\

\midrule
\multicolumn{2}{c}{Average} & {\color[HTML]{FF0000} \textbf{0.339}} & {\color[HTML]{FF0000} \textbf{0.351}} & 0.455   & 0.435   & 0.496 & 0.456 & 0.456 & 0.412 & {\color[HTML]{0000FF} 0.402} & {\color[HTML]{0000FF} 0.398} & 0.483   & 0.427   & 0.467 & 0.444 & 1.412   & 0.818 \\ 
\bottomrule
\end{tabular}
\end{small}
\end{threeparttable}
\label{table:long-term forecasting full results 3}
\end{table}

% 时间序列短期预测表1
\begin{table}[ht]
\caption{Full results for the short-term forecasting task in the M4 dataset compared with Peri-midFormer \cite{Peri-midFormer}, $S^2$IP-LLM \cite{S2IP-LLM}, Time-LLM \cite{Time-LLM}, GPT4TS \cite{GPT4TS}, TimeMixer \cite{TimeMixer}, PatchTST \cite{PatchTST}, iTransformer \cite{iTransformer}, TimesNet \cite{TimesNet}, DLinear \cite{DLinear}, Informer \cite{Informer}. (* means former.) \textbf{\textcolor{red}{Red}}: best, \textcolor{blue}{Blue}: second best.}
\centering
\vskip 0.10in
\begin{threeparttable}
\begin{small}
\setlength{\extrarowheight}{2pt}
\setlength{\tabcolsep}{3pt}
\begin{tabular}{c|c|ccccccccccc}
\toprule
\multicolumn{2}{c}{Methods} & \textbf{SymTime} & Peri-mid* & $S^2$IP-LLM & Time-LLM & GPT4TS & TimeMixer & PatchTST & iTrans* & TimesNet & DLinear & In* \\
\multicolumn{2}{c}{Metric} & \textbf{(Ours)} & \citeyearpar{Peri-midFormer} & \citeyearpar{S2IP-LLM} & \citeyearpar{Time-LLM} & \citeyearpar{GPT4TS} & \citeyearpar{TimeMixer} & \citeyearpar{PatchTST} & \citeyearpar{iTransformer} & \citeyearpar{TimesNet} & \citeyearpar{DLinear} & \citeyearpar{Informer} \\
\midrule
 & SMAPE & {\color[HTML]{FF0000} \textbf{13.355}} & 13.483 & 14.931 & 13.450 & 14.847 & {\color[HTML]{0000FF} 13.369} & 13.677 & 13.724 & 13.463 & 14.340  & 14.698 \\
 & MASE  & {\color[HTML]{FF0000} \textbf{2.997}}  & 3.080 & 3.345 & 3.184 & 3.628 & {\color[HTML]{0000FF} 3.009} & 3.049 & 3.157 & 3.058 & 3.112 & 3.293 \\
\multirow{-3}{*}{\rotatebox{90}{Yearly}}   & OWA   & {\color[HTML]{FF0000} \textbf{0.786}}  & 0.800 & 0.878    & 0.819 & 0.911 & {\color[HTML]{0000FF} 0.787}  & 0.802 & 0.817 & 0.797 & 0.830 & 0.864 \\
\midrule
 & SMAPE & {\color[HTML]{0000FF} 10.060} & {\color[HTML]{FF0000} \textbf{10.037}} & 10.655 & 10.671 & 10.389 & 10.131 & 10.922 & 13.473 & 10.069 & 10.510  & 16.172 \\
 & MASE & 1.183 & {\color[HTML]{FF0000} \textbf{1.170}}  & 1.249 & 1.276 & 1.228 & 1.186 & 1.326 & 1.722        & {\color[HTML]{0000FF} 1.175} & 1.241 & 2.136 \\
\multirow{-3}{*}{\rotatebox{90}{Quarterly}}  & OWA   & {\color[HTML]{FF0000} \textbf{0.872}}  & {\color[HTML]{0000FF} 0.882} & 0.939 & 0.950 & 0.919 & 0.893 & 0.979 & 1.240 & 0.886 & 0.930 & 1.513 \\
\midrule
 & SMAPE & {\color[HTML]{FF0000} \textbf{12.608}} & 12.795 & 13.012 & 13.416   & 12.907 & 12.762 & 14.200   & 13.674 & {\color[HTML]{0000FF} 12.760} & 13.382  & 15.446 \\
 & MASE & {\color[HTML]{FF0000} \textbf{0.925}}  & 0.948 & 0.973 & 1.045 & 0.954 & {\color[HTML]{0000FF} 0.940}  & 1.111 & 1.068 & 0.947 & 1.007 & 1.247 \\
\multirow{-3}{*}{\rotatebox{90}{Monthly}} & OWA & {\color[HTML]{FF0000} \textbf{0.872}} & 0.889 & 0.909    & 0.957 & 0.896 & {\color[HTML]{0000FF} 0.884}  & 1.015 & 0.976 & 0.887 & 0.937 & 1.122 \\
\midrule
 & SMAPE & {\color[HTML]{0000FF} 4.941} & {\color[HTML]{FF0000} \textbf{4.912}} & 5.540 & 4.973 & 5.266 & 5.085 & 5.658 & 5.598 & 4.995 & 5.122 & 6.839 \\
 & MASE  & {\color[HTML]{0000FF} 3.327} & {\color[HTML]{FF0000} \textbf{3.260}}  & 8.426 & 3.412 & 3.595 & 3.403 & 3.626 & 3.957 & 3.346 & 3.608 & 4.536 \\
\multirow{-3}{*}{\rotatebox{90}{Others}} & OWA   & {\color[HTML]{0000FF} 1.045} & {\color[HTML]{FF0000} \textbf{1.031}} & 3.792 & 1.059 & 1.121 & 1.072 & 1.167 & 1.213 & 1.053 & 1.108 & 1.435 \\
\midrule
 & SMAPE & {\color[HTML]{FF0000} \textbf{11.785}} & 11.897 & 12.514 & 12.584 & 12.367 & {\color[HTML]{0000FF} 11.885} & 12.866 & 13.233 & 11.888 & 12.500  & 15.018 \\
 & MASE & {\color[HTML]{FF0000} \textbf{1.584}}  & 1.607 & 1.726 & 1.763 & 1.767 & {\color[HTML]{0000FF} 1.598}  & 1.734 & 1.850 & 1.607 & 1.678 & 2.096 \\
\multirow{-3}{*}{\rotatebox{90}{Average}} & OWA & {\color[HTML]{FF0000} \textbf{0.849}} & 0.859 & 0.913    & 0.915 & 0.918 & {\color[HTML]{0000FF} 0.856} & 0.928 & 0.972 & 0.858 & 0.899 & 1.102 \\
\bottomrule
\end{tabular}
\end{small}
\end{threeparttable}
\label{table:short-term forecasting full results 1}
\end{table}

% 时间序列短期预测表2
\begin{table}[ht]
\caption{Full results for the short-term forecasting task in the M4 dataset compared with LightTS \cite{LightTS}, Autoformer \cite{Autoformer}, Crossformer \cite{Crossformer}, FEDformer \cite{FEDformer}, ETSformer \cite{ETSformer}, Nonstationary Transformer (Stationary) \cite{Non-stationary-transformers}, FiLM \cite{FiLM}, MICN \cite{MICN}, Reformer \cite{Reformer}, Pyraformer \cite{Pyraformer}. (* means former.) \textbf{\textcolor{red}{Red}}: best, \textcolor{blue}{Blue}: second best.}
\centering
\vskip 0.10in
\begin{threeparttable}
\begin{small}
\setlength{\extrarowheight}{2pt}
\setlength{\tabcolsep}{4.2pt}
\begin{tabular}{c|c|ccccccccccc}
\toprule
\multicolumn{2}{c}{Methods} & \textbf{SymTime} & LightTS & Auto* & Cross* & FED* & ETS* & Stationary & FiLM & MICN & Re* & Pyra* \\
\multicolumn{2}{c}{Metric} & \textbf{(Ours)} & \citeyearpar{LightTS} & \citeyearpar{Autoformer} & \citeyearpar{Crossformer} & \citeyearpar{FEDformer} & \citeyearpar{ETSformer} & \citeyearpar{Non-stationary-transformers} & \citeyearpar{FiLM} & \citeyearpar{MICN} & \citeyearpar{Reformer} & \citeyearpar{Pyraformer} \\
\midrule
 & SMAPE & {\color[HTML]{FF0000} \textbf{13.355}} & {\color[HTML]{0000FF} 13.444} & 17.764 & 79.308 & 13.508 & 18.009 & 13.717 & 14.076 & 14.557 & 13.752 & 14.594 \\
 & MASE  & {\color[HTML]{FF0000} \textbf{2.997}}  & 3.022 & 3.919 & 18.692 & 3.051 & 4.487 & 3.078 & {\color[HTML]{0000FF} 3.017} & 3.380 & 3.088 & 3.269 \\
\multirow{-3}{*}{\rotatebox{90}{Yearly}} & OWA & {\color[HTML]{FF0000} \textbf{0.786}}  & {\color[HTML]{0000FF} 0.792}  & 1.037 & 4.778 & 0.797 & 1.115 & 0.807 & 0.810 & 0.871 & 0.809 & 0.858 \\
\midrule
 & SMAPE & {\color[HTML]{FF0000} \textbf{10.060}} & {\color[HTML]{0000FF} 10.252} & 13.968 & 74.943 & 10.706 & 13.376 & 10.958 & 10.711 & 11.408 & 10.900 & 11.654 \\
 & MASE & {\color[HTML]{FF0000} \textbf{1.183}}  & {\color[HTML]{0000FF} 1.183} & 1.754 & 13.133 & 1.263 & 1.906 & 1.325 & 1.292 & 1.384 & 1.316 & 1.392 \\
\multirow{-3}{*}{\rotatebox{90}{Quarterly}} & OWA & {\color[HTML]{FF0000} \textbf{0.872}} & {\color[HTML]{0000FF} 0.897} & 1.274 & 8.191 & 0.947 & 1.302 & 0.981 & 0.957 & 1.022 & 0.975 & 1.037 \\
\midrule
 & SMAPE & {\color[HTML]{FF0000} \textbf{12.608}} & {\color[HTML]{0000FF} 12.798} & 18.200 & 68.892 & 13.925 & 14.588 & 13.917 & 13.362 & 13.803 & 13.949 & 14.963 \\
 & MASE  & {\color[HTML]{FF0000} \textbf{0.925}} & {\color[HTML]{0000FF} 0.957} & 1.574 & 11.199 & 1.062 & 1.368 & 1.097 & 1.016 & 1.078 & 1.096 & 1.165 \\
\multirow{-3}{*}{\rotatebox{90}{Monthly}} & OWA & {\color[HTML]{FF0000} \textbf{0.872}} & {\color[HTML]{0000FF} 0.894}  & {\color[HTML]{0000FF} 1.371} & {\color[HTML]{0000FF} 7.654} & {\color[HTML]{0000FF} 0.982} & {\color[HTML]{0000FF} 1.149} & {\color[HTML]{0000FF} 0.998} & {\color[HTML]{0000FF} 0.941} & {\color[HTML]{0000FF} 0.985} & {\color[HTML]{0000FF} 0.999} & {\color[HTML]{0000FF} 1.066} \\
\midrule
 & SMAPE & {\color[HTML]{0000FF} 4.941} & 5.324 & 6.738 & 176.164 & {\color[HTML]{FF0000} \textbf{4.888}} & 7.267 & 6.302 & 5.387 & 6.090 & 6.611 & 5.605 \\
 & MASE  & {\color[HTML]{0000FF} 3.327} & 3.410 & 4.853 & 116.723 & {\color[HTML]{FF0000} \textbf{3.244}} & 5.240 & 4.064 & 3.670 & 4.203 & 4.492 & 3.966 \\
\multirow{-3}{*}{\rotatebox{90}{Others}} & OWA & {\color[HTML]{0000FF} 1.045} & 1.098 & 1.474 & 36.941 & {\color[HTML]{FF0000} \textbf{1.026}} & 1.591 & 1.304 & 1.146 & 1.304 & 1.404 & 1.215 \\
\midrule
 & SMAPE & {\color[HTML]{FF0000} \textbf{11.785}} & {\color[HTML]{0000FF} 11.962} & 16.511 & 78.103 & 12.605 & 14.718 & 12.780 & 12.491 & 13.016 & 12.805 & 13.616 \\
 & MASE  & {\color[HTML]{FF0000} \textbf{1.584}}  & {\color[HTML]{0000FF} 1.609} & 2.321 & 18.663 & 1.677 & 2.408 & 1.756 & 1.675 & 1.837 & 1.777 & 1.843 \\
\multirow{-3}{*}{\rotatebox{90}{Average}} & OWA & {\color[HTML]{FF0000} \textbf{0.849}}  & {\color[HTML]{0000FF} 0.862}  & 1.215 & 7.759 & 0.903 & 1.172 & 0.930 & 0.899 & 0.960 & 0.937 & 0.984 \\
\bottomrule
\end{tabular}
\end{small}
\end{threeparttable}
\label{table:short-term forecasting full results 2}
\end{table}

\begin{table}[ht]
\caption{Full results for time series classification task compared with (1) classical methods: DTW \cite{DTW}, XGBoost \cite{XGBoost}, Rocket \cite{Rocket}; (2) RNN-based methods: LSTM \cite{LSTM}, LSTNet \cite{LSTNet}, LSSL \cite{LSSL}; (3) CNN-based methods: InceptionTime (InTime) \cite{InceptionTime}, TCN \cite{TCN}, TimesNet \cite{TimesNet}, TSLANet \cite{TSLANet}. We report the classification accuracy (\%) as the result. \textbf{\textcolor{red}{Red}}: best, \textcolor{blue}{Blue}: second best.}
\centering
\vskip 0.10in
\begin{threeparttable}
\begin{small}
\renewcommand{\multirowsetup}{\centering}
\setlength{\extrarowheight}{3pt}
\setlength{\tabcolsep}{3pt}
\begin{tabular}{c|ccccccccccc}
\toprule
& \multicolumn{3}{c}{Classical Methods} & \multicolumn{3}{c}{RNN-based} & \multicolumn{4}{c}{CNN-based} & \\
\cmidrule(lr){2-4} \cmidrule(lr){5-7} \cmidrule(lr){8-11}
& DTW  & XGBoost & Rocket & LSTM & LSTNet & LSSL & InTime & TCN & TimesNet & TSLANet & \textbf{SymTime} \\ 
\multirow{-3}{*}{Datasets / Methods} 
& \citeyearpar{DTW}  & \citeyearpar{XGBoost} & \citeyearpar{Rocket} & \citeyearpar{LSTM} & \citeyearpar{LSTNet} & \citeyearpar{LSSL} & \citeyearpar{InceptionTime} & \citeyearpar{TCN} & \citeyearpar{TimesNet} & \citeyearpar{TSLANet} & (\textbf{Ours}) \\ \midrule
EthanolConcentration & 32.3 & {\color[HTML]{0000FF} 43.7} & {\color[HTML]{FF0000} \textbf{45.2}} & 32.3 & 39.9 & 31.1 & 39.1 & 28.9 & 35.7 & 30.4 & 37.3 \\
FaceDetection & 52.9 & 63.3 & 64.7 & 57.7 & 65.7 & 66.7 & 65.4 & 52.8 & {\color[HTML]{0000FF} 68.6} & 66.7 & {\color[HTML]{FF0000} \textbf{69.2}} \\
Handwriting & 28.6 & 15.8 & {\color[HTML]{FF0000} \textbf{58.8}} & 15.2 & 25.8 & 24.6 & 46.9 & 53.3 & 32.1 & {\color[HTML]{0000FF} 57.9} & 36.7 \\
Heartbeat & 71.7 & 73.2 & 75.6 & 72.2 & 77.1 & 72.7 & 75.2 & 75.6 & {\color[HTML]{FF0000} \textbf{78.0}} & {\color[HTML]{0000FF} 77.5} & 74.1 \\
JapaneseVowels & 94.9 & 86.5 & 96.2 & 79.7 & 98.1 & {\color[HTML]{0000FF} 98.4} & 95.1 & {\color[HTML]{FF0000} \textbf{98.9}} & {\color[HTML]{0000FF} 98.4} & 95.1 & 98.1 \\
PEMS-SF & 71.1 & {\color[HTML]{FF0000} \textbf{98.3}} & 75.1 & 39.9 & 86.7 & 86.1 & 79.6 & 68.8 & 89.6 & 83.8 & {\color[HTML]{0000FF} 92.5} \\
SelfRegulationSCP1 & 77.7 & 84.6 & {\color[HTML]{0000FF} 90.8} & 68.9 & 84.0 & {\color[HTML]{0000FF} 90.8} & 87.2 & 84.6 & {\color[HTML]{FF0000} \textbf{91.8}} & {\color[HTML]{FF0000} \textbf{91.8}} & 89.8 \\
SelfRegulationSCP2 & 53.9 & 48.9 & 53.3 & 46.6 & 52.8 & 52.2 & 53.6 & 55.6 & {\color[HTML]{0000FF} 57.2} & 53.3 & {\color[HTML]{FF0000} \textbf{58.9}} \\
SpokenArabicDigits & 96.3 & 69.6 & 71.2 & 31.9 & {\color[HTML]{FF0000} \textbf{100.0}} & {\color[HTML]{FF0000} \textbf{100.0}} & 96.3 & 95.6 & {\color[HTML]{0000FF} 99.0} & 98.0 & 98.9 \\
UWaveGestureLibrary & 90.3 & 75.9 & {\color[HTML]{FF0000} \textbf{94.4}} & 41.2 & 87.8 & 85.9 & {\color[HTML]{0000FF} 92.4} & 88.4 & 85.3 & 89.4 & 89.4 \\ \midrule
Average Accuracy & 67.0 & 66.0 & 72.5 & 48.6 & 71.8 & 70.9 & 73.1 & 70.3 & 73.6 & {\color[HTML]{0000FF} 74.4} & {\color[HTML]{FF0000} \textbf{74.5}} \\ \bottomrule
\end{tabular}
\end{small}
\end{threeparttable}
\label{table:classification full results 1}
\end{table}

\begin{table}[ht]
\caption{Full reuslts for time series classification task compared with (1) Transformer-based methods: Autoformer \cite{Autoformer}, FEDformer \cite{FEDformer}, ETSformer \cite{ETSformer}, Informer \cite{Informer}, iTransformer \cite{iTransformer}, PatchTST (Patch) \cite{PatchTST}, GPT4TS (GPT) \cite{GPT4TS}, UniTS \cite{UniTS}, Peri-midformer \cite{Peri-midFormer} and (2) MLP-based methods: DLinear \cite{DLinear}, LightTS \cite{LightTS}. We report the classification accuracy (\%) as the results. * means former. \textbf{\textcolor{red}{Red}}: best, \textcolor{blue}{Blue}: second best.}
\centering
\vskip 0.10in
\begin{threeparttable}
\begin{small}
\renewcommand{\multirowsetup}{\centering}
\setlength{\extrarowheight}{3pt}
\setlength{\tabcolsep}{3pt}
\begin{tabular}{c|cccccccccccc}
\toprule
& \multicolumn{9}{c}{Transformer-based} & \multicolumn{2}{c}{MLP-based} & \\
\cmidrule(lr){2-10} \cmidrule(lr){11-12}
& Auto* & FED* & ETS* & In* & iTrans* & Patch & GPT & UniTS & Peri-mid* & DLinear & LightTS & \textbf{SymTime} \\
\multirow{-3}{*}{Datasets / Methods} 
& \citeyearpar{Autoformer} & \citeyearpar{FEDformer} & \citeyearpar{ETSformer} & \citeyearpar{Informer} & \citeyearpar{iTransformer} & \citeyearpar{PatchTST} & \citeyearpar{GPT4TS} & \citeyearpar{UniTS} & \citeyearpar{Peri-midFormer} & \citeyearpar{DLinear} & \citeyearpar{LightTS} & (\textbf{Ours}) \\
\midrule
EthanolConcentration & 31.6 & 31.2 & 28.1 & 31.6 & 27.0 & 29.6 & {\color[HTML]{FF0000} \textbf{34.2}} & {\color[HTML]{0000FF} 37.3} & {\color[HTML]{FF0000} \textbf{47.3}} & 32.6 & 29.7 & {\color[HTML]{0000FF} 37.3} \\
FaceDetection & {\color[HTML]{FF0000} \textbf{68.4}}  & 66.0 & 66.3 & {\color[HTML]{0000FF} 67.0} & 67.0 & 67.8 & {\color[HTML]{FF0000} \textbf{69.2}} & 67.5 & {\color[HTML]{0000FF} 68.7} & 68.0 & 67.5 & {\color[HTML]{FF0000} \textbf{69.2}} \\
Handwriting & {\color[HTML]{FF0000} \textbf{36.7}}  & 28.0 & 32.5 & {\color[HTML]{0000FF} 32.8} & 27.2 & 23.2 & 32.7 & {\color[HTML]{0000FF} 27.0} & {\color[HTML]{FF0000} \textbf{31.5}} & 27.0 & 26.1 & {\color[HTML]{FF0000} \textbf{36.7}} \\
Heartbeat & 74.6 & 73.7 & 71.2 & {\color[HTML]{0000FF} 80.5}  & 75.6 & 75.7 & {\color[HTML]{0000FF} 77.2} & {\color[HTML]{0000FF} 80.5} & {\color[HTML]{FF0000} \textbf{86.3}} & 75.1 & 75.1 & 74.1 \\
JapaneseVowels & 96.2 & 98.4 & 95.9 & {\color[HTML]{FF0000} \textbf{98.9}}  & 97.6 & 94.0 & {\color[HTML]{0000FF} 98.6} & {\color[HTML]{FF0000} \textbf{97.8}} & 96.8 & 96.2    & 96.2 & {\color[HTML]{0000FF} 98.1} \\
PEMS-SF & 82.7 & 80.9 & 86.0 & {\color[HTML]{0000FF} 81.5} & {\color[HTML]{FF0000} \textbf{85.5}} & 80.9     & 87.9 & {\color[HTML]{FF0000} \textbf{93.1}} & 88.2 & 75.1    & 88.4 & {\color[HTML]{0000FF} 92.5} \\
SelfRegulationSCP1 & 84.0 & 88.7 & 89.6 & {\color[HTML]{0000FF} 90.1} & {\color[HTML]{FF0000} \textbf{92.2}} & 82.2 & {\color[HTML]{0000FF} 87.2} & {\color[HTML]{FF0000} \textbf{89.6}} & 87.4 & 87.3    & 89.8 & 89.8 \\
SelfRegulationSCP2 & {\color[HTML]{FF0000} \textbf{50.6}}  & {\color[HTML]{FF0000} \textbf{54.4}}  & {\color[HTML]{FF0000} \textbf{55.0}}  & {\color[HTML]{FF0000} \textbf{53.3}}  & 54.4 & 53.6 & {\color[HTML]{0000FF} 59.4} & {\color[HTML]{FF0000} \textbf{61.1}} & 55.4 & 50.5 & {\color[HTML]{FF0000} \textbf{51.1}}  & {\color[HTML]{0000FF} 58.9} \\
SpokenArabicDigits & {\color[HTML]{FF0000} \textbf{100.0}} & {\color[HTML]{FF0000} \textbf{100.0}} & {\color[HTML]{FF0000} \textbf{100.0}} & {\color[HTML]{FF0000} \textbf{100.0}} & 98.0 & 98.0 & 95.2 & {\color[HTML]{0000FF} 98.9} & 98.0 & 81.4    & {\color[HTML]{FF0000} \textbf{100.0}} & {\color[HTML]{0000FF} 98.9} \\
UWaveGestureLibrary & 85.9 & 85.3 & 85.0 & 85.6 & 85.9 & 81.7 & 85.1 & {\color[HTML]{0000FF} 87.8} & {\color[HTML]{0000FF} 84.3} & 82.1 & 80.3 & {\color[HTML]{FF0000} \textbf{89.4}} \\
\midrule
Average Accuracy & 71.1 & 70.7 & 71.0 & 72.1 & 71.1 & 68.7     & 72.7 & 74.1 & {\color[HTML]{0000FF} 74.4} & 67.5    & 70.4 & {\color[HTML]{FF0000} \textbf{74.5}} \\ 
\bottomrule
\end{tabular}
\end{small}
\end{threeparttable}
\label{table:classification full results 2}
\end{table}

\begin{table}[ht]
\caption{Full results for time series imputation task, where we randomly mask \{12.5\%, 25\%, 37.5\%, 50\%\} time points of length-96 time series to compare the model performance under different missing degrees. We compare with GPT4TS \cite{GPT4TS}, TimesNet \cite{TimesNet}, Peri-midFormer \cite{Peri-midFormer}, Moment \cite{Moment}, iTransformer \cite{iTransformer}, PatchTST \cite{PatchTST}, DLinear \cite{DLinear} in this table. (* means former.) The standard deviation is within 0.5\%. \textbf{\textcolor{red}{Red}}: best, \textcolor{blue}{Blue}: second best.}
\centering
\vskip 0.10in
\begin{threeparttable}
\begin{small}
\renewcommand{\multirowsetup}{\centering}
\setlength{\extrarowheight}{3pt}
\setlength{\tabcolsep}{3pt}
\begin{tabular}{c|c|cc|cc|cc|cc|cc|cc|cc|cc}
\toprule
\multicolumn{2}{c}{}  & \multicolumn{2}{c}{\textbf{SymTime}}    & \multicolumn{2}{c}{GPT4TS}    & \multicolumn{2}{c}{TimesNet}   & \multicolumn{2}{c}{Peri-mid*}  & \multicolumn{2}{c}{Moment} & \multicolumn{2}{c}{iTrans*} & \multicolumn{2}{c}{PatchTST} & \multicolumn{2}{c}{DLinear} \\

\multicolumn{2}{c}{\multirow{-2}{*}{Models}} & \multicolumn{2}{c}{\textbf{(Ours)}}  & \multicolumn{2}{c}{\citeyearpar{GPT4TS}}    & \multicolumn{2}{c}{\citeyearpar{TimesNet}}   & \multicolumn{2}{c}{\citeyearpar{Peri-midFormer}}  & \multicolumn{2}{c}{\citeyearpar{Moment}} & \multicolumn{2}{c}{\citeyearpar{iTransformer}} & \multicolumn{2}{c}{\citeyearpar{PatchTST}} & \multicolumn{2}{c}{\citeyearpar{DLinear}} \\
\cmidrule(lr){3-4} \cmidrule(lr){5-6} \cmidrule(lr){7-8} \cmidrule(lr){9-10} \cmidrule(lr){11-12} \cmidrule(lr){13-14} \cmidrule(lr){15-16} \cmidrule(lr){17-18}
\multicolumn{2}{c}{Mask Ratio} & MSE   & MAE   & MSE   & MAE   & MSE   & MAE   & MSE   & MAE   & MSE   & MAE   & MSE   & MAE  & MSE  & MAE   & MSE   & MAE   \\
\midrule

   & 12.5\% & 0.032    & 0.110    & \textbf{0.018} & \textbf{0.090} & 0.019    & 0.091    & 0.032    & 0.109    & 0.069  & 0.170  & 0.046  & 0.147   & 0.045   & 0.137  & 0.056  & 0.162  \\
   & 25\%  & 0.034    & 0.113    & \textbf{0.024} & 0.102    & \textbf{0.024} & \textbf{0.101} & 0.034    & 0.112    & 0.071  & 0.169  & 0.060  & 0.171   & 0.046   & 0.139  & 0.077  & 0.191  \\
   & 37.5\% & 0.037    & 0.118    & \textbf{0.029} & \textbf{0.111} & \textbf{0.029} & 0.112    & 0.037    & 0.117    & 0.069  & 0.163  & 0.077  & 0.195   & 0.049   & 0.143  & 0.100  & 0.218  \\
   & 50\%  & 0.041    & 0.126    & 0.042    & 0.132    & \textbf{0.036} & \textbf{0.124} & 0.042    & 0.126    & 0.086  & 0.169  & 0.104  & 0.228   & 0.055   & 0.152  & 0.129  & 0.247  \\
\cmidrule(lr){2-18}
\multirow{-5}{*}{\rotatebox{90}{ETTm1}}   & Avg   & 0.036    & 0.116    & {\color[HTML]{0000FF} 0.028}   & {\color[HTML]{0000FF} 0.109}   & {\color[HTML]{FF0000} \textbf{0.027}} & {\color[HTML]{FF0000} \textbf{0.107}} & 0.036    & 0.116    & 0.074  & 0.168  & 0.072  & 0.185   & 0.049   & 0.143  & 0.090  & 0.204  \\
\midrule

   & 12.5\% & 0.024    & 0.084    & \textbf{0.018} & 0.081    & 0.019    & 0.081    & 0.023    & 0.081    & 0.032  & 0.108  & 0.052  & 0.151   & 0.026   & 0.094  & 0.067  & 0.171  \\
   & 25\%  & 0.024    & 0.086    & 0.021    & 0.082    & 0.021    & 0.086    & 0.024    & 0.084    & 0.029  & 0.105  & 0.070  & 0.179   & 0.028   & 0.099  & 0.089  & 0.200  \\
   & 37.5\% & 0.027    & 0.089    & 0.023    & 0.090    & 0.023    & 0.091    & 0.026    & 0.089    & 0.032  & 0.109  & 0.091  & 0.204   & 0.031   & 0.104  & 0.112  & 0.226  \\
   & 50\%  & 0.030    & 0.093    & 0.027    & 0.098    & 0.026    & 0.098    & 0.030    & 0.095    & 0.031  & 0.110  & 0.117  & 0.232   & 0.034   & 0.109  & 0.140  & 0.253  \\
\cmidrule(lr){2-18}
\multirow{-5}{*}{\rotatebox{90}{ETTm2}}   & Avg   & {\color[HTML]{0000FF} 0.026}   & 0.088    & {\color[HTML]{FF0000} \textbf{0.022}} & {\color[HTML]{FF0000} \textbf{0.088}} & {\color[HTML]{FF0000} \textbf{0.022}} & {\color[HTML]{0000FF} 0.089}   & {\color[HTML]{0000FF} 0.026}   & 0.087    & 0.031  & 0.108  & 0.082  & 0.191   & 0.030   & 0.101  & 0.102  & 0.212  \\
\midrule

   & 12.5\% & 0.074    & 0.179    & 0.063    & 0.171    & 0.062    & 0.169    & 0.069    & 0.173    & 0.160  & 0.239  & 0.098  & 0.220   & 0.097   & 0.203  & 0.111  & 0.232  \\
   & 25\%  & 0.082    & 0.190    & 0.080    & 0.190    & 0.081    & 0.191    & 0.079    & 0.185    & 0.142  & 0.238  & 0.125  & 0.249   & 0.115   & 0.221  & 0.149  & 0.269  \\
   & 37.5\% & 0.100    & 0.205    & 0.107    & 0.218    & 0.098    & 0.210    & 0.096    & 0.202    & 0.121  & 0.228  & 0.156  & 0.278   & 0.134   & 0.239  & 0.187  & 0.301  \\
   & 50\%  & 0.123    & 0.230    & 0.121    & 0.221    & 0.116    & 0.227    & 0.122    & 0.226    & 0.132  & 0.231  & 0.213  & 0.327   & 0.160   & 0.260  & 0.229  & 0.332  \\
\cmidrule(lr){2-18}
\multirow{-5}{*}{\rotatebox{90}{ETTh1}}   & Avg   & 0.095    & 0.201    & 0.093    & 0.200    & {\color[HTML]{FF0000} \textbf{0.089}} & {\color[HTML]{0000FF} 0.199}   & {\color[HTML]{0000FF} 0.091}   & {\color[HTML]{FF0000} \textbf{0.196}} & 0.139  & 0.234  & 0.148  & 0.269   & 0.126   & 0.231  & 0.169  & 0.283  \\
\midrule

   & 12.5\% & 0.051    & 0.138    & 0.041    & 0.129    & 0.040    & 0.132    & 0.051    & 0.139    & 0.051  & 0.150  & 0.095  & 0.210   & 0.058   & 0.153  & 0.109  & 0.223  \\
   & 25\%  & 0.055    & 0.146    & 0.046    & 0.138    & 0.047    & 0.144    & 0.054    & 0.142    & 0.079  & 0.177  & 0.120  & 0.239   & 0.063   & 0.160  & 0.146  & 0.260  \\
   & 37.5\% & 0.059    & 0.152    & 0.060    & 0.160    & 0.054    & 0.154    & 0.058    & 0.148    & 0.056  & 0.155  & 0.149  & 0.266   & 0.068   & 0.167  & 0.180  & 0.290  \\
   & 50\%  & 0.064    & 0.157    & 0.061    & 0.160    & 0.061    & 0.164    & 0.064    & 0.159    & 0.056  & 0.154  & 0.192  & 0.302   & 0.074   & 0.175  & 0.217  & 0.319  \\
\cmidrule(lr){2-18}
\multirow{-5}{*}{\rotatebox{90}{ETTh2}}   & Avg   & 0.058    & {\color[HTML]{0000FF} 0.148}   & {\color[HTML]{0000FF} 0.052}   & {\color[HTML]{FF0000} \textbf{0.147}} & {\color[HTML]{FF0000} \textbf{0.050}} & {\color[HTML]{0000FF} 0.148}   & 0.057    & {\color[HTML]{FF0000} \textbf{0.147}} & 0.061  & 0.159  & 0.139  & 0.254   & 0.066   & 0.164  & 0.163  & 0.273  \\
\midrule

   & 12.5\% & 0.037    & 0.122    & 0.080    & 0.195    & 0.088    & 0.203    & 0.047    & 0.140    & 0.095  & 0.211  & 0.073  & 0.190   & 0.061   & 0.170  & 0.084  & 0.206  \\
   & 25\%  & 0.046    & 0.139    & 0.089    & 0.205    & 0.092    & 0.208    & 0.053    & 0.162    & 0.093  & 0.211  & 0.090  & 0.214   & 0.072   & 0.185  & 0.113  & 0.243  \\
   & 37.5\% & 0.060    & 0.160    & 0.094    & 0.217    & 0.096    & 0.214    & 0.067    & 0.179    & 0.094  & 0.211  & 0.107  & 0.235   & 0.082   & 0.198  & 0.141  & 0.273  \\
   & 50\%  & 0.075    & 0.181    & 0.108    & 0.231    & 0.102    & 0.221    & 0.085    & 0.195    & 0.092  & 0.210  & 0.127  & 0.257   & 0.097   & 0.216  & 0.173  & 0.303  \\
\cmidrule(lr){2-18}
\multirow{-5}{*}{\rotatebox{90}{ECL}}  & Avg   & {\color[HTML]{FF0000} \textbf{0.054}} & {\color[HTML]{FF0000} \textbf{0.151}} & 0.093    & 0.212    & 0.094    & 0.211    & {\color[HTML]{0000FF} 0.063}   & {\color[HTML]{0000FF} 0.169}   & 0.094  & 0.211  & 0.099  & 0.224   & 0.078   & 0.192  & 0.128  & 0.256  \\
\midrule

   & 12.5\% & 0.025    & 0.035    & 0.026    & 0.047    & 0.026    & 0.049    & 0.025    & 0.037    & 0.033  & 0.073  & 0.038  & 0.087   & 0.028   & 0.049  & 0.039  & 0.091  \\
   & 25\%  & 0.027    & 0.037    & 0.030    & 0.055    & 0.030    & 0.056    & 0.026    & 0.037    & 0.036  & 0.078  & 0.046  & 0.106   & 0.032   & 0.055  & 0.049  & 0.112  \\
   & 37.5\% & 0.029    & 0.039    & 0.033    & 0.061    & 0.032    & 0.058    & 0.029    & 0.041    & 0.034  & 0.075  & 0.055  & 0.122   & 0.035   & 0.059  & 0.057  & 0.125  \\
   & 50\%  & 0.032    & 0.042    & 0.039    & 0.070    & 0.034    & 0.062    & 0.034    & 0.048    & 0.035  & 0.075  & 0.068  & 0.142   & 0.039   & 0.064  & 0.067  & 0.139  \\
\cmidrule(lr){2-18}
\multirow{-5}{*}{\rotatebox{90}{Weather}}  & Avg   & {\color[HTML]{FF0000} \textbf{0.028}} & {\color[HTML]{FF0000} \textbf{0.038}} & 0.032    & 0.058    & 0.030    & 0.056    & {\color[HTML]{0000FF} 0.029}   & {\color[HTML]{0000FF} 0.041}   & 0.035  & 0.075  & 0.052  & 0.114   & 0.033   & 0.057  & 0.053  & 0.116  \\
\midrule
\multicolumn{2}{c}{Average} & {\color[HTML]{FF0000} \textbf{0.050}} & {\color[HTML]{FF0000} \textbf{0.125}} & 0.053    & 0.136    & 0.052    & 0.135    & {\color[HTML]{FF0000} \textbf{0.050}} & {\color[HTML]{0000FF} 0.126}   & 0.072  & 0.159  & 0.099  & 0.206   & 0.064   & 0.148  & 0.118  & 0.224 \\
\bottomrule
\end{tabular}
\end{small}
\end{threeparttable}
\label{table:imputation full results 1}
\end{table}

\begin{table}[ht]
\caption{Full results for time series imputation task, where we randomly mask \{12.5\%, 25\%, 37.5\%, 50\%\} time points of length-96 time series to compare the model performance under different missing degrees. We compare with Stationary \cite{Non-stationary-transformers}, LightTS \cite{LightTS}, ETSformer \cite{ETSformer}, FEDformer \cite{FEDformer}, Informer \cite{Informer}, Reformer \cite{Reformer} and Pyraformer \cite{Pyraformer} in this table. (Stationary means Nonstationary Transformer.) The standard deviation is within 0.5\%. \textbf{\textcolor{red}{Red}}: best, \textcolor{blue}{Blue}: second best.}
\centering
\vskip 0.10in
\begin{threeparttable}
\begin{small}
\renewcommand{\multirowsetup}{\centering}
\setlength{\extrarowheight}{3pt}
\setlength{\tabcolsep}{3pt}
\begin{tabular}{c|c|cc|cc|cc|cc|cc|cc|cc|cc}
\toprule
\multicolumn{2}{c}{}  & \multicolumn{2}{c}{\textbf{SymTime}}    & \multicolumn{2}{c}{Stationary}  & \multicolumn{2}{c}{LightTS}   & \multicolumn{2}{c}{ETSformer} & \multicolumn{2}{c}{FEDformer} & \multicolumn{2}{c}{Informer}   & \multicolumn{2}{c}{Reformer}   & \multicolumn{2}{c}{Pyraformer} \\
\multicolumn{2}{c}{\multirow{-2}{*}{Methods}} & \multicolumn{2}{c}{\textbf{(Ours)}}  & \multicolumn{2}{c}{Stationary}  & \multicolumn{2}{c}{LightTS}   & \multicolumn{2}{c}{ETSformer} & \multicolumn{2}{c}{FEDformer} & \multicolumn{2}{c}{Informer}   & \multicolumn{2}{c}{Reformer}   & \multicolumn{2}{c}{Pyraformer} \\
\cmidrule(lr){3-4} \cmidrule(lr){5-6} \cmidrule(lr){7-8} \cmidrule(lr){9-10} \cmidrule(lr){11-12} \cmidrule(lr){13-14} \cmidrule(lr){15-16} \cmidrule(lr){17-18}
\multicolumn{2}{c}{Mask Ratio} & MSE   & MAE   & MSE   & MAE   & MSE  & MAE  & MSE  & MAE  & MSE  & MAE  & MSE  & MAE  & MSE  & MAE  & MSE  & MAE  \\
\midrule

   & 12.5\% & 0.032    & 0.110    & 0.026    & 0.107    & 0.054 & 0.158 & 0.034   & 0.130   & 0.068   & 0.188   & 0.027 & 0.115 & 0.032 & 0.126 & 0.670   & 0.541   \\
   & 25\%  & 0.034    & 0.113    & 0.032    & 0.119    & 0.061 & 0.173 & 0.053   & 0.162   & 0.097   & 0.230   & 0.040 & 0.140 & 0.042 & 0.146 & 0.689   & 0.553   \\
   & 37.5\% & 0.037    & 0.118    & 0.039    & 0.131    & 0.073 & 0.189 & 0.082   & 0.201   & 0.134   & 0.287   & 0.071 & 0.189 & 0.063 & 0.182 & 0.737   & 0.581   \\
   & 50\%  & 0.041    & 0.126    & 0.047    & 0.145    & 0.086 & 0.207 & 0.130   & 0.257   & 0.188   & 0.323   & 0.091 & 0.208 & 0.082 & 0.208 & 0.770   & 0.605   \\
\cmidrule(lr){2-18}
\multirow{-5}{*}{\rotatebox{90}{ETTm1}}   & Avg   & {\color[HTML]{FF0000} \textbf{0.036}} & {\color[HTML]{FF0000} \textbf{0.116}} & {\color[HTML]{FF0000} \textbf{0.036}} & {\color[HTML]{0000FF} 0.126}   & 0.068 & 0.182 & 0.075   & 0.187   & 0.121   & 0.257   & 0.057 & 0.163 & {\color[HTML]{0000FF} 0.055} & 0.166 & 0.717   & 0.570   \\
\midrule

   & 12.5\% & 0.024    & 0.084    & 0.021    & 0.088    & 0.051 & 0.150 & 0.061   & 0.169   & 0.109   & 0.239   & 0.196 & 0.326 & 0.108 & 0.228 & 0.394   & 0.470   \\
   & 25\%  & 0.024    & 0.086    & 0.024    & 0.096    & 0.069 & 0.176 & 0.093   & 0.214   & 0.166   & 0.295   & 0.295 & 0.414 & 0.136 & 0.262 & 0.421   & 0.482   \\
   & 37.5\% & 0.027    & 0.089    & 0.027    & 0.103    & 0.074 & 0.185 & 0.137   & 0.253   & 0.237   & 0.356   & 0.155 & 0.293 & 0.175 & 0.300 & 0.478   & 0.521   \\
   & 50\%  & 0.030    & 0.093    & 0.030    & 0.108    & 0.078 & 0.192 & 0.237   & 0.332   & 0.323   & 0.412   & 0.214 & 0.325 & 0.211 & 0.329 & 0.568   & 0.560   \\
\cmidrule(lr){2-18}
\multirow{-5}{*}{\rotatebox{90}{ETTm2}}   & Avg   & {\color[HTML]{FF0000} \textbf{0.026}} & {\color[HTML]{FF0000} \textbf{0.088}} & {\color[HTML]{FF0000} \textbf{0.026}} & {\color[HTML]{0000FF} 0.099}   & {\color[HTML]{0000FF} 0.068} & 0.176 & 0.132   & 0.242   & 0.209   & 0.326   & 0.215 & 0.340 & 0.157 & 0.280 & 0.465   & 0.508   \\
\midrule

   & 12.5\% & 0.074    & 0.179    & 0.060    & 0.165    & 0.119 & 0.239 & 0.073   & 0.195   & 0.126   & 0.265   & 0.068 & 0.187 & 0.074 & 0.194 & 0.857   & 0.609   \\
   & 25\%  & 0.082    & 0.190    & 0.080    & 0.189    & 0.144 & 0.266 & 0.105   & 0.234   & 0.169   & 0.305   & 0.096 & 0.220 & 0.102 & 0.227 & 0.829   & 0.672   \\
   & 37.5\% & 0.100    & 0.205    & 0.102    & 0.212    & 0.171 & 0.292 & 0.144   & 0.276   & 0.220   & 0.348   & 0.128 & 0.253 & 0.135 & 0.261 & 0.830   & 0.675   \\
   & 50\%  & 0.123    & 0.230    & 0.133    & 0.240    & 0.201 & 0.317 & 0.200   & 0.327   & 0.298   & 0.403   & 0.166 & 0.287 & 0.179 & 0.298 & 0.854   & 0.691   \\
\cmidrule(lr){2-18}
\multirow{-5}{*}{\rotatebox{90}{ETTh1}}   & Avg   & {\color[HTML]{0000FF} 0.095}   & {\color[HTML]{FF0000} \textbf{0.201}} & {\color[HTML]{FF0000} \textbf{0.094}} & {\color[HTML]{FF0000} \textbf{0.201}} & 0.159 & 0.278 & 0.130   & 0.258   & 0.204   & 0.330   & 0.115 & {\color[HTML]{0000FF} 0.237} & 0.122 & 0.245 & 0.842   & 0.682   \\
\midrule

   & 12.5\% & 0.051    & 0.138    & 0.042    & 0.133    & 0.094 & 0.208 & 0.134   & 0.251   & 0.187   & 0.319   & 0.271 & 0.384 & 0.163 & 0.289 & 0.976   & 0.754   \\
   & 25\%  & 0.055    & 0.146    & 0.049    & 0.147    & 0.140 & 0.255 & 0.180   & 0.294   & 0.279   & 0.396   & 0.362 & 0.450 & 0.206 & 0.331 & 1.037   & 0.774   \\
   & 37.5\% & 0.059    & 0.152    & 0.056    & 0.158    & 0.159 & 0.274 & 0.243   & 0.341   & 0.402   & 0.465   & 0.401 & 0.469 & 0.252 & 0.370 & 1.107   & 0.800   \\
   & 50\%  & 0.064    & 0.157    & 0.065    & 0.170    & 0.180 & 0.293 & 0.353   & 0.408   & 0.604   & 0.504   & 0.437 & 0.487 & 0.316 & 0.419 & 1.193   & 0.838   \\
\cmidrule(lr){2-18}
\multirow{-5}{*}{\rotatebox{90}{ETTh2}}   & Avg   & {\color[HTML]{0000FF} 0.058}   & {\color[HTML]{FF0000} \textbf{0.148}} & {\color[HTML]{FF0000} \textbf{0.053}} & {\color[HTML]{0000FF} 0.152}   & 0.143 & 0.258 & 0.228   & 0.324   & 0.368   & 0.421   & 0.368 & 0.448 & 0.234 & 0.352 & 1.079   & 0.792   \\
\midrule

   & 12.5\% & 0.037    & 0.122    & 0.093    & 0.210    & 0.077 & 0.198 & 0.185   & 0.323   & 0.197   & 0.324   & 0.152 & 0.279 & 0.190 & 0.308 & 0.297   & 0.383   \\
   & 25\%  & 0.046    & 0.139    & 0.097    & 0.214    & 0.099 & 0.228 & 0.207   & 0.340   & 0.208   & 0.345   & 0.166 & 0.290 & 0.197 & 0.312 & 0.294   & 0.380   \\
   & 37.5\% & 0.060    & 0.160    & 0.102    & 0.220    & 0.120 & 0.252 & 0.226   & 0.355   & 0.219   & 0.337   & 0.178 & 0.297 & 0.203 & 0.315 & 0.296   & 0.381   \\
   & 50\%  & 0.075    & 0.181    & 0.108    & 0.228    & 0.138 & 0.272 & 0.251   & 0.372   & 0.235   & 0.357   & 0.189 & 0.305 & 0.210 & 0.319 & 0.299   & 0.383   \\
\cmidrule(lr){2-18}
\multirow{-5}{*}{\rotatebox{90}{ECL}}  & Avg   & {\color[HTML]{FF0000} \textbf{0.049}} & {\color[HTML]{FF0000} \textbf{0.151}} & {\color[HTML]{0000FF} 0.100}   & {\color[HTML]{0000FF} 0.218}   & 0.108 & 0.238 & 0.217   & 0.347   & 0.215   & 0.341   & 0.171 & 0.293 & 0.200 & 0.313 & 0.297   & 0.382   \\
\midrule

   & 12.5\% & 0.025    & 0.035    & 0.027    & 0.051    & 0.039 & 0.092 & 0.042   & 0.103   & 0.057   & 0.141   & 0.040 & 0.108 & 0.031 & 0.076 & 0.140   & 0.220   \\
   & 25\%  & 0.027    & 0.037    & 0.029    & 0.056    & 0.045 & 0.105 & 0.056   & 0.131   & 0.066   & 0.155   & 0.045 & 0.130 & 0.035 & 0.082 & 0.147   & 0.229   \\
   & 37.5\% & 0.029    & 0.039    & 0.033    & 0.062    & 0.049 & 0.110 & 0.081   & 0.180   & 0.083   & 0.180   & 0.049 & 0.101 & 0.040 & 0.091 & 0.156   & 0.240   \\
   & 50\%  & 0.032    & 0.042    & 0.037    & 0.068    & 0.054 & 0.117 & 0.102   & 0.207   & 0.103   & 0.207   & 0.054 & 0.114 & 0.046 & 0.099 & 0.164   & 0.249   \\
\cmidrule(lr){2-18}
\multirow{-5}{*}{\rotatebox{90}{Weather}}  & Avg   & {\color[HTML]{FF0000} \textbf{0.028}} & {\color[HTML]{FF0000} \textbf{0.038}} & {\color[HTML]{0000FF} 0.032}   & {\color[HTML]{0000FF} 0.059}   & 0.047 & 0.106 & 0.071   & 0.155   & 0.077   & 0.171   & 0.047 & 0.113 & 0.038 & 0.087 & 0.152   & 0.235   \\
\midrule

\multicolumn{2}{c}{Average} & {\color[HTML]{FF0000} \textbf{0.049}} & {\color[HTML]{FF0000} \textbf{0.124}} & {\color[HTML]{0000FF} 0.057}   & {\color[HTML]{0000FF} 0.143}   & 0.099 & 0.206 & 0.142   & 0.252   & 0.199   & 0.308   & 0.162 & 0.265 & 0.134 & 0.241 & 0.592   & 0.528  \\
\bottomrule
\end{tabular}
\end{small}
\end{threeparttable}
\label{table:imputation full results 2}
\end{table}

\begin{table}[ht]
\caption{Full reuslts for time series anomaly detection task, where P, R and F1 represent the precision, recall and F1-score (\%) respectively. F1-score is the harmonic mean of precision and recall. A higher value of P, R and F1 indicates a better performance. We compare with: Transformer \cite{Transformer}, Reformer \cite{Reformer}, Informer \cite{Informer}, Autoformer \cite{Autoformer}, Crossformer \cite{Crossformer}, iTransformer \cite{iTransformer}, Anomaly \cite{Anomaly-Transformer}, Stationary \cite{Non-stationary-transformers}, DLinear \cite{DLinear}, LightTS \cite{LightTS}, ETSformer \cite{ETSformer}, FEDformer \cite{FEDformer}, PatchTST \cite{PatchTST}, TimesNet \cite{TimesNet}, GPT4TS \cite{GPT4TS}, Peri-midFormer \cite{Peri-midFormer}, UniTS \cite{UniTS}, where Anomaly means the Anomaly Transformer and Stationary means the Non-stationary Transformer. The standard deviation is within 1\%. \textbf{\textcolor{red}{Red}}: best, \textcolor{blue}{Blue}: second best.}
\centering
\vskip 0.10in
\begin{threeparttable}
\begin{small}
\renewcommand{\multirowsetup}{\centering}
\setlength{\extrarowheight}{4.5pt}
\setlength{\tabcolsep}{2pt}
\begin{tabular}{c|ccc|ccc|ccc|ccc|ccc|c}
\toprule
Datasets & \multicolumn{3}{c}{SMD} & \multicolumn{3}{c}{MSL} & \multicolumn{3}{c}{SMAP} & \multicolumn{3}{c}{SWaT} & \multicolumn{3}{c}{PSM} & Avg F1 \\
\cmidrule(lr){2-4} \cmidrule(lr){5-7} \cmidrule(lr){8-10} \cmidrule(lr){11-13} \cmidrule(lr){14-16} \cmidrule(lr){17-17}
Metircs & P & R & F1 & P & R & F1 & P & R & F1 & P & R & F1 & P & R & F1 & (\%) \\
\midrule
Transformer \citeyearpar{Transformer}  & 78.44 & 65.26 & 71.24 & 89.85 & 73.71 & 80.99 & 90.77 & 61.76 & 73.50 & 96.82 & 66.41 & 79.76 & 99.31 & 83.18 & 90.53 & 79.20 \\

Reformer \citeyearpar{Reformer} & 72.50 & {\color[HTML]{0000FF} 84.19} & 77.90 & 90.24 & 73.78 & 81.18 & 90.63 & {\color[HTML]{0000FF} 62.48} & {\color[HTML]{0000FF} 73.97} & {\color[HTML]{0000FF} 99.94} & 66.75 & 80.04 & 99.73 & 83.03 & 90.62 & 80.74 \\

Informer \citeyearpar{Informer} & 72.51 & 84.13 & 77.88 & 90.10 & 73.68 & 81.07 & 90.57 & 61.51 & 73.26 & 99.83 & 67.24 & 80.35 & 99.03 & 83.21 & 90.43 & 80.60 \\

Autoformer \citeyearpar{Autoformer} & 78.46 & 65.11 & 71.17 & {\color[HTML]{0000FF} 90.59} & 75.26 & 82.22 & 90.84 & 62.39 & {\color[HTML]{0000FF} 73.97} & {\color[HTML]{FF0000} \textbf{99.95}} & 65.57 & 79.19 & {\color[HTML]{FF0000} \textbf{99.99}} & 78.96 & 88.24 & 78.96 \\

Crossformer \citeyearpar{Crossformer} & 71.89 & 83.41 & 77.22 & 90.32 & 72.74 & 80.59 & 89.68 & 53.63 & 67.12 & 98.00 & 83.59 & 90.22 & 97.49 & 88.02 & 92.52 & 81.53 \\

iTransformer \citeyearpar{iTransformer} & 76.13 & 84.70 & 80.19 & 86.15 & 62.54 & 72.47 & 90.68 & 52.78 & 66.72 & 92.23 & 93.05 & 92.64 & 97.92 & 92.03 & 94.88 & 81.38 \\

Anomaly \citeyearpar{Anomaly-Transformer} & {\color[HTML]{FF0000} \textbf{88.91}} & 82.23 & {\color[HTML]{FF0000} \textbf{85.49}} & 79.61 & {\color[HTML]{FF0000} \textbf{87.37}} & {\color[HTML]{0000FF} 83.31} & 91.85 & 58.11 & 71.18 & 72.51 & {\color[HTML]{FF0000} \textbf{97.32}} & 83.10 & 68.35 & 94.72 & 79.40 & 80.50 \\

Stationary \citeyearpar{Non-stationary-transformers} & 78.51 & {\color[HTML]{FF0000} \textbf{87.98}} & 82.97 & 86.86 & 68.63 & 76.68 & 90.62 & 55.74 & 69.02 & 89.26 & {\color[HTML]{0000FF} 95.42} & 92.24 & 98.17 & {\color[HTML]{FF0000} \textbf{96.30}} & {\color[HTML]{0000FF} 97.23} & 83.63 \\

DLinear \citeyearpar{DLinear} & 75.91 & 84.02 & 79.76 & 89.68 & 75.31 & 81.87 & 89.87 & 53.79 & 67.30 & 92.26 & 93.05 & 92.66 & 98.65 & 94.70 & 96.64 & 83.64 \\

LightTS \citeyearpar{LightTS} & 87.10 & 78.42 & 82.53 & 82.40 & 75.78 & 78.95 & {\color[HTML]{FF0000} \textbf{92.58}} & 55.27 & 69.21 & 91.98 & 94.72 & {\color[HTML]{0000FF} 93.33} & 98.37 & 95.97 & 97.15 & 84.23 \\

ETSformer \citeyearpar{ETSformer} & 87.44 & 79.23 & 83.13 & 85.13 & {\color[HTML]{0000FF} 84.93} & {\color[HTML]{FF0000} \textbf{85.03}} & {\color[HTML]{0000FF} 92.25}          & 55.75 & 69.50 & 90.02 & 80.36 & 84.91 & 99.31 & 85.28 & 91.76 & 82.87 \\

FEDformer \citeyearpar{FEDformer} & 72.82 & 81.68 & 76.99 & {\color[HTML]{FF0000} \textbf{90.72}} & 75.41 & 82.36 & 90.47 & 58.10 & 70.76 & 99.95 & 65.55 & 79.18 & {\color[HTML]{0000FF} 99.98} & 81.92 & 90.05 & 79.46 \\

PatchTST \citeyearpar{PatchTST} & 87.26 & 82.14 & 84.62 & 88.34 & 70.96 & 78.70 & 90.64 & 55.46 & 68.82 & 91.10 & 80.94 & 85.72 & 98.84 & 93.47 & 96.08 & 82.79 \\

TimesNet \citeyearpar{TimesNet} & {\color[HTML]{0000FF} 88.07} & 80.97 & 84.37 & 88.83 & 74.68 & 81.14 & 89.98 & 56.02 & {\color[HTML]{0000FF} 69.05} & 91.99 & 93.24 & 92.61 & 98.46 & 95.70 & 97.06 & 84.85 \\

GPT4TS \citeyearpar{GPT4TS} & 87.68 & 81.52 & 84.49 & 82.09 & 81.97 & 82.03 & 90.12 & 55.70 & 68.85 & 92.12 & 93.09 & 92.60 & 98.36 & 95.85 & 97.09 & 85.01 \\

Peri-midFormer \citeyearpar{Peri-midFormer} & 86.97 & 81.37 & 84.08 & 88.66 & 74.02 & 80.68 & 90.02 & 54.03 & 67.53 & 90.74 & 92.55 & 91.64 & 98.46 & 94.06 & 96.21 & 84.03 \\

UniTS \citeyearpar{UniTS} & 82.42 & 84.99 & 83.69 & 91.32 & 73.04 & 81.16 & 90.58 & {\color[HTML]{FF0000} \textbf{62.55}} & {\color[HTML]{FF0000} \textbf{74.00}} & 92.60 & 92.42 & 92.51 & 98.45 & {\color[HTML]{0000FF} 96.19} & {\color[HTML]{FF0000} \textbf{97.31}} & {\color[HTML]{FF0000} \textbf{85.73}} \\

\textbf{\texttt{SymTime} (Ours)} & 87.93 & 81.56 & {\color[HTML]{0000FF} 84.62} & 89.46 & 75.31 & 81.77 & 90.34 & 56.96 & 69.87 & 95.94 & 91.39 & {\color[HTML]{FF0000} \textbf{93.61}} & 98.89 & 95.32 & 97.07 & {\color[HTML]{0000FF} 85.39} \\

\bottomrule
\end{tabular}
\end{small}
\end{threeparttable}
\label{table:anomaly detection full results}
\end{table}

\end{document}